\documentclass{article}

\PassOptionsToPackage{numbers,sort&compress}{natbib}

\usepackage[preprint]{neurips_2026}

\usepackage[utf8]{inputenc}
\usepackage[T1]{fontenc}
\usepackage{amsmath,amssymb,amsthm}
\usepackage{mathtools}          
\usepackage{thmtools}
\usepackage[shortlabels]{enumitem}
\usepackage{booktabs}
\usepackage{graphicx}
\usepackage{xcolor}
\usepackage{algorithm}
\usepackage{algpseudocode}
\usepackage{nicefrac}
\usepackage{microtype}
\usepackage{url}
\usepackage{appendix}
\usepackage{wrapfig}
\usepackage{placeins}
\usepackage{titletoc}
\usepackage[colorlinks,citecolor=blue,urlcolor=blue,linkcolor=red,hypertexnames=false]{hyperref}
\definecolor{mydarkblue}{rgb}{0,0.08,0.45}

\titlecontents{lsection}
  [0em]
  {\addvspace{0.45em}\bfseries}
  {\contentslabel{1.8em}}
  {}
  {\titlerule*[0.5pc]{.}\contentspage}
\titlecontents{lsubsection}
  [1.8em]
  {\small}
  {\contentslabel{2.4em}}
  {}
  {\titlerule*[0.5pc]{.}\contentspage}
\makeatletter
\AtBeginDocument{%
  \def\ttl@gobblecontents#1#2#3#4{\ignorespaces}%
}
\makeatother

\newtheorem{theorem}{Theorem}
\newtheorem{lemma}[theorem]{Lemma}
\newtheorem{proposition}[theorem]{Proposition}
\newtheorem{corollary}[theorem]{Corollary}
\newtheorem{definition}[theorem]{Definition}
\newtheorem{assumption}{Assumption}

\newtheorem{setting}[theorem]{Setting}

\newtheorem{conjecture}[theorem]{Conjecture}

\theoremstyle{remark}
\newtheorem{remark}[theorem]{Remark}

\newcommand{\AV}{\mathcal{A}_v}
\newcommand{\AVw}{\mathcal{A}_{v,\mathrm{within}}}
\newcommand{\AVb}{\mathcal{A}_{v,\mathrm{between}}}

\newcommand{\safeincludegraphics}[2][]{%
  \IfFileExists{#2}{\includegraphics[#1]{#2}}{%
    \fbox{\parbox[c][0.24\textheight][c]{0.92\linewidth}{\centering Missing figure: \texttt{\detokenize{#2}}}}}
}

\title{Velocity Ambiguity Profiles:
       Time-Resolved Bayes-Risk Diagnostics
       for Flow Matching}

\author{
  Yunpeng Mei\textsuperscript{1,2}\thanks{Equal contribution.} \quad Xiaowen Zhu\textsuperscript{1}\textsuperscript{*} \quad Chenyu Wang\textsuperscript{1}\textsuperscript{*} \quad Chenbo Xin\textsuperscript{1}\textsuperscript{*} \\
  \textbf{Hongjie Cao\textsuperscript{1} \quad Jiamin Wang\textsuperscript{1} \quad Jie Chen\textsuperscript{1} \quad
  Gang Wang\textsuperscript{1}\thanks{Corresponding author.}} 
   \\[0.5em]
  \normalfont \textsuperscript{1}Beijing Institute of Technology \quad \textsuperscript{2}ByteDance
}

\begin{document}
\maketitle
\begin{abstract}
Flow matching trains a continuum of velocity regression problems but usually
reports a loss averaged over interpolation time, obscuring which regions are
limited by intrinsic ambiguity and which remain data-limited. We introduce the
Velocity Ambiguity Profile (VAP), $\AV(t)$, the time-local Bayes-risk floor of
velocity prediction, and decompose time-resolved loss as
$L_N(t)=\AV(t)+R(t;N)$, where only the excess term $R(t;N)$ is reducible by more
data or a different estimator. For $N$ training samples and a $K$-component mixture in the structured
example, we prove two statistical benchmarks. In an intrinsic local-regression benchmark, $\AV(t)$ enters the time-dependent noise
factor of the standard nonparametric rate. In separated isotropic Gaussian
mixtures, the Bayes velocity becomes locally affine, and a label-oracle
residual-regression benchmark has parametric excess-risk scale $K\AV(t)/N$,
matched within the same separated residual Gaussian-location subproblem.
Controlled Gaussian-mixture experiments validate this diagnostic through a
dense transition sweep, calibrated oracle-to-blind and empirical-proxy audits,
and a local controlled time-sampler stress test in which a frozen
reducible-error proxy reduces reducible loss while cross-cell sensitivity
delimits aggressive sampling. On CIFAR-100 DINOv2 feature flow, guided time
sampling reduces uniformly evaluated velocity loss by $6.78\%$ and global
feature Fr\'echet distance (FD) by $23.6\%$, while class-conditional kernel inception
distance (KID) shows no improvement. VAP turns an averaged flow-matching objective into
a time-resolved diagnostic for locating ambiguity-limited and data-limited
regions.
\end{abstract}


\section{Introduction}\label{sec:intro}

Flow matching reduces distribution learning to a continuum of velocity
regression problems indexed by interpolation time, but training and evaluation
usually report a path-averaged loss. This hides where additional data can help
and where the target remains intrinsically ambiguous.

This paper asks a simple diagnostic question: \emph{which interpolation times
remain limited by reducible estimation error, and which are already limited by
intrinsic velocity ambiguity?} A high loss at time~$t$ is actionable only when
it exceeds the Bayes-risk floor. It may reflect an under-trained estimator, or
it may reflect multiple data realizations remaining plausible under the same
noisy interpolant. Finite-data error can be reduced; posterior ambiguity
cannot. The conditional-mean orthogonality behind this decomposition is
classical. Our contribution is to make the induced posterior ambiguity
profile path-specific, expose its mode and information-set structure, and use
it for time-local diagnostics.

\paragraph{Velocity ambiguity as time-resolved Bayes risk.}
We propose the \emph{Velocity Ambiguity Profile} (VAP), $\AV(t)$, as that
diagnostic. Let $a$ be an expert action, $\varepsilon\sim\mathcal N(0,I)$ be
Gaussian noise, $c$ be the observation-language context, and let
$x_t=(1-t)a+t\varepsilon$, $u_t=\varepsilon-a$, with the convention that $t=0$
is the data endpoint and $t=1$ is the noise endpoint. VAP is the conditional
velocity variance averaged over the induced interpolant distribution
\begin{equation}\label{eq:vap-intro}
  \AV(t)
  \;\coloneqq\;
  \mathbb{E}_{x_t,c}\!\left[
    \operatorname{tr}\operatorname{Cov}(u_t\mid x_t,c,t)
  \right].
\end{equation}
For $t>0$, $\AV(t)$ is equivalently the posterior action variance rescaled by
$1/t^2$. VAP specializes conditional Bayes risk to the velocity regression
problems induced by the flow path and makes it diagnostic along flow-matching
time. For any trained velocity estimator, the time-resolved loss decomposes as
$L_N(t)=\AV(t)+R(t;N)$, where $\AV(t)$ is the irreducible Bayes-risk floor and
$R(t;N)$ is the reducible excess risk. Only $R(t;N)$ can be lowered by
collecting more data or changing the estimator.

\begin{wrapfigure}[30]{r}{0.5\linewidth}
  \centering
\safeincludegraphics[width=\linewidth]{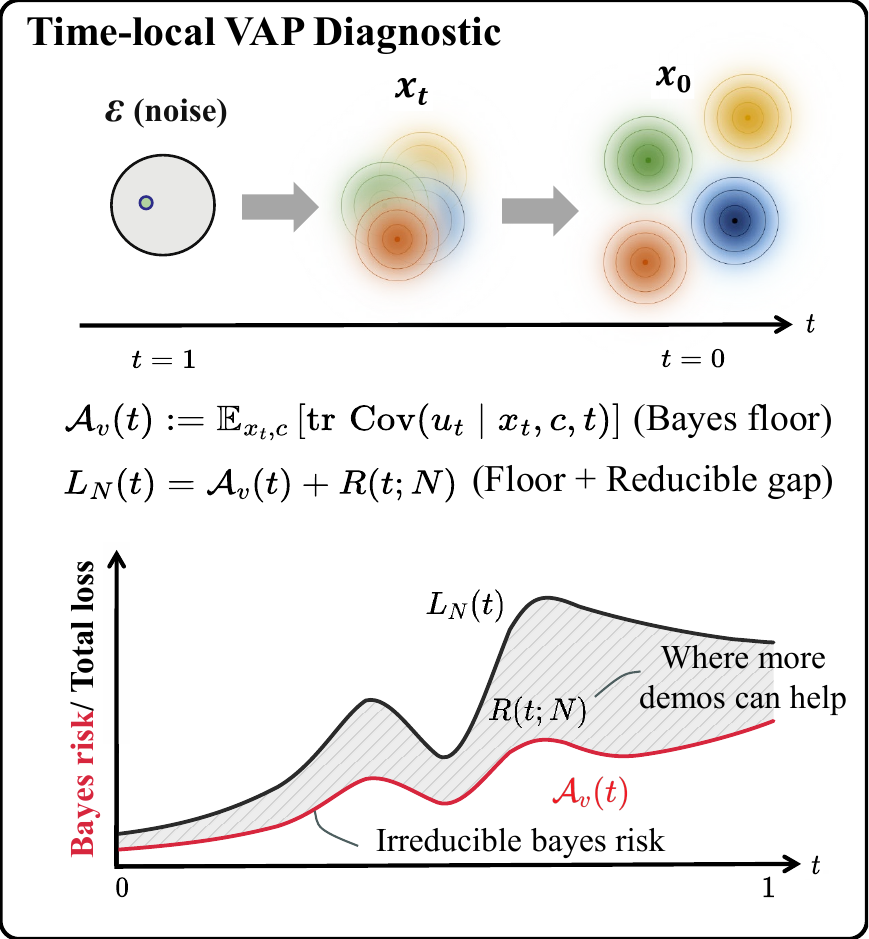}
  \caption{\textbf{Velocity ambiguity profiles localize where flow matching
    needs data.}
    The flow matching path defines a continuum of velocity prediction
    problems. VAP separates the irreducible ambiguity floor from the reducible
    excess error that more data or a different estimator can affect.}
  \label{fig:overview}
\end{wrapfigure}

VAP complements empirical scaling laws and time-aggregated generative-model
baselines~\citep{hu2025data,dass2025datamil,oko2023,weber2024}. It asks which
interpolation times are data-limited, which are ambiguity-limited, and where
controlled models separate these regimes. Figure~\ref{fig:overview} previews
this time-local view.

\paragraph{Main results.}
The paper develops VAP at three levels. First, we prove the exact Bayes-risk
decomposition above and instantiate a time-local intrinsic regression
benchmark: on compact time intervals, the local excess-risk rate has the usual
nonparametric exponent, with time dependence controlled by $\AV(t)$ up to
bounded factors. Second, in the separated regime of an isotropic Gaussian
mixture, we obtain a sharper label-oracle residual-regression benchmark: the
Bayes velocity is locally affine, the residual benchmark has parametric scale
$K\AV(t)/N$, matched by a separated residual Gaussian-location lower bound.
Third, controlled Gaussian-mixture experiments validate the diagnostic when
the ambiguity floor is observable. They show separated-regime parametric
scaling under a reliability-filtered readout, oracle-to-blind agreement with
the analytic between-component label cost, and small but measurable blind
neural residuals. A local time-sampler stress test further shows that a frozen
proxy from disjoint source runs can produce same-sign paired reducible-loss
reductions in the canonical balanced cell, while cross-cell sensitivity
delimits aggressive pure sampling. Fourth, a CIFAR-100 DINOv2 feature-flow
experiment tests transfer to natural feature distributions: guided sampling
reduces uniformly evaluated velocity loss by $6.78\%$ and global feature FD by
$23.6\%$. Macro conditional KID shows no gain, and lower recall and effective
covariance rank reveal a diversity cost.

\paragraph{Contributions.}
This paper makes the following five contributions.
\begin{enumerate}[leftmargin=*,itemsep=1pt]
  \item \textbf{A time-local Bayes-risk diagnostic for flow matching.}
    We define VAP, $\AV(t)$, and prove the exact decomposition
    $L_N(t)=\AV(t)+R(t;N)$. This turns one averaged flow matching loss into a
    time-resolved separation between irreducible ambiguity and reducible error.

  \item \textbf{Scoped statistical benchmarks for reducible error.}
    For an intrinsic local regression benchmark, $\AV(t)$ enters the standard
    nonparametric rate as the time-dependent noise level up to bounded factors.
    For separated isotropic Gaussian mixtures, we prove a sharper
    label-oracle residual-regression benchmark: the Bayes velocity is
    locally affine and the benchmark excess-risk scale is $K\AV(t)/N$,
    matched by a separated residual Gaussian-location subproblem lower bound.

  \item \textbf{Controlled Gaussian-mixture validation and sampler stress test.}
    Controlled Gaussian-mixture experiments validate VAP as a time-resolved
    diagnostic, showing analytic-empirical agreement, separated-regime
    parametric scaling, and documented sensitivity to the reliability readout.
    A local controlled time-sampler stress test further shows that a frozen
    reducible-error proxy from disjoint source runs can produce same-sign
    paired reducible-loss reductions, with cross-cell sensitivity delimiting
    aggressive pure sampling.

  \item \textbf{An oracle-to-blind bridge and proxy calibration.}
    Controlled bridge audits show that the label-cost decomposition is
    quantitatively accurate, while blind multilayer perceptron (MLP) residuals and empirical proxies
    are small but not exact and must be treated as calibrated diagnostics.

  \item \textbf{Natural-feature transfer with explicit scope.}
    In DINOv2 feature flow, guided sampling reduces uniformly evaluated
    velocity loss by $6.78\%$ and global feature FD by $23.6\%$. Macro conditional
    KID shows no gain, while recall and effective covariance rank decrease.
\end{enumerate}

\section{Preliminaries}\label{sec:prelim}

\subsection{Flow Matching}\label{sec:fm}

We adopt the conditional flow matching
(CFM) framework~\citep{lipman2023flow,liu2023flow,albergo2023building,albergo2023stochastic}.
Let $a\in\mathbb{R}^D$ denote an expert action
or data vector,
and let $\varepsilon\sim\mathcal{N}(0,I_D)$ be isotropic Gaussian noise.
The linear interpolation path is
\begin{equation}\label{eq:ot-path}
  x_t \;=\; t\,\varepsilon \;+\; (1{-}t)\,a,
  \qquad \forall t\in[0,1],
\end{equation}
with conditional velocity field
$u_t = \varepsilon - a$.
A neural network $v_\theta(x_t, t, c)$, conditioned on context
$c$ when present, is trained to regress onto $u_t$ by minimizing the
\emph{flow matching loss}
\begin{equation}\label{eq:fm-loss}
  \mathcal{L}_{\mathrm{FM}}(\theta)
  \;=\;
  \mathbb{E}_{t,\,a,\,\varepsilon}\Bigl[
    \bigl\lVert v_\theta(x_t,t,c) - u_t \bigr\rVert^2
  \Bigr],
\end{equation}
where $t$ is sampled from a training distribution
(for example uniform or a prescribed time-sampling schedule).
At inference, actions are generated by integrating $v_\theta$ from
$x_1\sim\mathcal{N}(0,I)$ to $x_0$ via an ordinary-differential equation (ODE) solver.


\section{Velocity Ambiguity Profile}
\label{sec:sigma}

We now introduce the central object of this paper: a single, time-resolved
quantity that captures the inherent difficulty of velocity estimation at each
point along the interpolation path.

\subsection{Definition and Interpretation}
\label{sec:sigma-def}

Recall the linear interpolation $x_t=(1-t)a+t\varepsilon$ with conditional
velocity target $u_t=\varepsilon-a$. For a fixed time $t$ and an estimator trained on $N$ samples, the Bayes-optimal
velocity is $v^*(x_t,c,t)=\mathbb{E}[u_t\mid x_t,c,t]$. Any estimator of this
conditional mean incurs a residual variance that cannot be reduced by additional
model capacity or training samples.

\begin{definition}[Velocity ambiguity profile]
\label{def:sigma}
The \emph{velocity ambiguity profile} (VAP) is
\begin{equation}\label{eq:sigma-def}
\begin{aligned}
    \AV(t)
    &\;\coloneqq\;
    \inf_{f_t}\,\mathbb{E}\!\left[\|f_t(x_t,c)-u_t\|^2\;\middle|\;t\right]
    \;=\;
    \mathbb{E}_{x_t,c}\!\left[
      \mathrm{tr}\,\mathrm{Cov}\!\left(u_t\mid x_t,c,t\right)
    \right] \\
    &\;=\;
    \frac{1}{t^2}\,
    \mathbb{E}_{x_t,c}\!\left[
      \mathrm{tr}\,\mathrm{Var}\!\left(a\mid x_t,c,t\right)
    \right]\qquad\text{for all }t>0,
\end{aligned}
\end{equation}
where the infimum is over all measurable functions $f_t$ that may depend on the
fixed time $t$, and $\AV(0)$ is interpreted as the one-sided limit $\AV(0^+)$
when it exists.
\end{definition}

Thus, $\AV(t)$ is the Bayes risk at time $t$: the smallest possible expected
squared velocity error given perfect knowledge of the data distribution. It is
a property of the time-local regression problem induced by the flow path. The
action-posterior form (third equality) follows from $u_t=\varepsilon-a=(x_t-a)/t$,
so, conditional on $(x_t,c,t)$, all residual uncertainty in $u_t$ comes from
posterior uncertainty in the original action $a$, rescaled by $1/t^2$.

Throughout the theory, we often condition on a fixed context
$c$ and suppress it from notation; all expectations and variances are then
implicitly conditional on $c$.

\begin{proposition}[Endpoint behavior and non-monotonicity of $\AV(t)$]
\label{prop:sigma-properties}
Under Assumption~\ref{asm:manifold}, $\AV(0^+)<\infty$ (in particular
$\AV(0^+)=d$ for the isotropic Gaussian mixture in \S\ref{sec:sigma-gmm});
if $\mathbb{E}[\|a\|^2]<\infty$, then
$\AV(1)=\operatorname{tr}\operatorname{Var}_{p_{\mathrm{data}}}(a)$;
and $\AV(t)$ is non-monotone in general.
\end{proposition}
The single-Gaussian closed form and full proof are in
Appendix~\ref{app:sigma-properties-proof}: near $t=0$ the observation
$x_t\approx a$ nearly determines the data realization, while near $t=1$ the
observation $x_t\approx\varepsilon$ carries little information about which
data realization generated the sample.

\subsection{Analytic Example: Gaussian Mixture}
\label{sec:sigma-gmm}

To build concrete intuition, consider a separate analytically tractable model:
a $K$-component isotropic Gaussian mixture
$p_{\mathrm{data}}=\sum_{k=1}^K\pi_k\mathcal N(\mu_k,\sigma^2I_d)$ in
$\mathbb R^d$. This Gaussian example is not an instance of the compact manifold
benchmark in Theorem~\ref{thm:minimax}; it is a structured model where the
posterior and Bayes velocity can be computed explicitly. Writing
$\sigma_t^2=(1-t)^2\sigma^2+t^2$, direct calculation yields
\begin{equation}\label{eq:sigma-gmm}
    \AV(t)
    =
    \underbrace{\frac{d\sigma^2}{\sigma_t^2}}_{\text{within-component}}
    +
    \underbrace{
      \frac{t^2}{\sigma_t^4}
      \mathbb{E}_{x_t}\!\left[
        \sum_{j<k}\gamma_j(x_t,t)\gamma_k(x_t,t)\|\mu_j-\mu_k\|^2
      \right]
    }_{\text{between-component}},
\end{equation}
where $\gamma_k(x_t,t)$ is the posterior responsibility of component $k$
(explicit softmax form in Appendix~\ref{sec:proof-theorem2prime}).

Equation~\eqref{eq:sigma-gmm} has a simple qualitative reading. The
within-component term is the residual velocity variance after the component is
known; for an isotropic component it equals $d\sigma^2/\sigma_t^2$. The
between-component term is small when $x_t$ identifies the component and grows
when several component means remain plausible under the same interpolant. Thus
small $t$ is mostly within-component, intermediate $t$ is where posterior
responsibilities mix, and large $t$ approaches the covariance scale of the data
because $x_t$ carries little component information. This is a mechanism-level
calculation, not an empirical claim. The claim-bearing controlled display is
deferred to Section~\ref{sec:controlled-exp}, where the same decomposition is
tested across the fixed dense separation sweep.

A useful consequence of the same posterior calculation is that, conditioned on
component $k$, the Bayes velocity is exactly affine
\begin{equation}\label{eq:gmm-component-affine}
  v_k^*(x_t,t)=A(t)x_t+b_k(t),\qquad
  A(t)=\frac{t-(1-t)\sigma^2}{\sigma_t^2}I_d,\qquad
  b_k(t)=-\frac{t}{\sigma_t^2}\mu_k.
\end{equation}
Theorem~\ref{thm:gmm} uses this affine form as the dominant
component approximation in the separated regime.

\subsection{Bias-Variance Decomposition of the Time-Resolved Loss}
\label{sec:decomposition}

\begin{proposition}[Loss decomposition]
\label{prop:decomposition}
For any fixed estimator $\hat v_N$ trained on $N$ samples, the population test
loss at time $t$ satisfies
\begin{equation}\label{eq:loss-decomp}
  L_N(t)
  \coloneqq
  \mathbb E\!\left[\|\hat v_N(x_t,c,t)-u_t\|^2\mid t\right]
  =
  \underbrace{\AV(t)}_{\text{Bayes risk}}
  +
  \underbrace{R(t;N)}_{\text{excess risk}},
\end{equation}
where
$R(t;N)=\mathbb E[\|\hat v_N(x_t,c,t)-v^*(x_t,c,t)\|^2\mid t]\ge0$.
If $\hat v_N$ is random through the training set, the same identity holds after
an additional expectation over the training set.
\end{proposition}

The proof is the standard conditional-mean orthogonality argument and is given
in Appendix~\ref{app:loss-decomposition-proof}.

\subsection{\texorpdfstring{Practical Estimation of $\AV(t)$}{Practical Estimation of VAP}}
\label{sec:sigma-estimation}

\paragraph{Large-budget reference profile.}
In applied flow matching settings, $\AV(t)$ is not directly observable. We
therefore use the largest-budget validation loss only as an upper-biased
reference profile,
$L_{\mathrm{ref}}(t)\equiv L_{N_{\max}}(t)$. At population risk,
Proposition~\ref{prop:decomposition} gives
$L_{\mathrm{ref}}(t)=\AV(t)+R(t;N_{\max})\geq\AV(t)$, so a finite-budget
reference is not an absolute Bayes-floor estimator. If a current model and
the reference are evaluated on the same distribution, their difference
removes the unknown floor exactly:
\begin{equation}\label{eq:reference-relative-gap}
  L_N(t)-L_{\mathrm{ref}}(t)=R(t;N)-R(t;N_{\max}).
\end{equation}
This reference-relative gap is a calibrated allocation diagnostic; it does
not identify the absolute Bayes floor. Consistency of the reference toward
$\AV(t)$ additionally requires vanishing training excess and a growing
validation sample under the usual moment conditions.

\paragraph{Collapse diagnostics.}
If $R(t;N)=\AV(t)g(N)$, then $L_N(t)/\AV(t)-1=g(N)$. Thus the clean
collapse diagnostic is normalized excess risk, $R(t;N)/\AV(t)$, or
$L_N(t)/\AV(t)-1$ when $\AV(t)$ is known. Raw losses $L_N(t)$ alone
saturate to the Bayes floor and should only be interpreted as approximate power
laws when the excess term dominates the floor.


\section{Main Theoretical Results}
\label{sec:theory}

We now ask when $\AV(t)$ predicts the sample-complexity scale, not only the
Bayes-risk floor. Theorem~\ref{thm:minimax} gives a compact local-regression
benchmark in which $\AV(t)$ enters a nonparametric rate up to bounded factors
in the fixed-positive-noise regime;
Theorem~\ref{thm:gmm} gives a separated-GMM label-oracle
residual-regression benchmark with parametric scale $K\AV(t)/N$, matched
inside the corresponding separated residual Gaussian-location subproblem. This
benchmark calibrates the residual-regression view of the flow objective; it is
not a lower bound over all action-aware estimators of the full supervised
tuple.

\paragraph{Oracle terminology.}
We reserve \emph{label-oracle residual benchmark} for the theorem-level
information set: the estimator receives true component labels and fits the
residual affine response in each separated cell. A \emph{label-and-action
oracle} is stronger: it also uses the supervised pair to recover
$a=x_t-t u_t$ and estimate component means directly. An \emph{analytic oracle}
uses the known mixture law only to compute diagnostic quantities such as
$\AV(t)$ and posterior responsibilities. The oracle-to-blind bridge below
compares these controlled oracle arms with estimators that do not receive
labels or analytic responsibilities.


\subsection{Time-Resolved Nonparametric Benchmark Bound}
\label{sec:thm1}

Our starting point is classical minimax theory for nonparametric regression
\citep{stone1982optimal,tsybakov2009nonparametric}, instantiated for the
velocity-estimation problem at fixed time $t$. The benchmark relies on the
regularity assumptions in Appendix~\ref{app:thm1-assumptions}. For each
$t\in[\delta,1-\delta]$, let $\mathcal G_t^{\mathrm{comp}}(s,d,B,\AV)$
denote the formal compact $d$-dimensional heteroscedastic regression
benchmark obtained from a finite atlas of $(1-t)\mathcal M$, with regression
function in a vector H\"older ball of radius $B(t)$, compact random design,
and pointwise conditional noise comparable to $\AV(t)$.

\begin{theorem}[Time-resolved compact-benchmark bound]
\label{thm:minimax}
Fix $t\in[\delta,1-\delta]$. Under
Assumptions~\ref{asm:manifold}-\ref{asm:noise-comparability} together with
the compact-design condition~\ref{asm:compact-design}, consider the
formal compact random-design local benchmark
$\mathcal G_t^{\mathrm{comp}}(s,d,B,\AV)$, defined by the finite-atlas,
boundary-Gram, and vector-H\"older conditions in
Appendix~\ref{app:thm1-assumptions}. Its integrated
prediction risk
\begin{equation}\label{eq:minimax-rate}
  R^*_{\mathrm{loc}}(t;N)
  \coloneqq
  \inf_{\hat v}\sup_{P\in\mathcal G_t^{\mathrm{comp}}}
  \mathbb E_{\mathcal D_N,X\sim P_{x_t}}\!
  \left[\|\hat v_{\mathcal D_N}(X,t)-v_P^*(X,t)\|^2\right]
\end{equation}
has the lower bound
\begin{equation}\label{eq:minimax-nonparametric}
  c_1 B(t)^{\frac{2d}{2s+d}}
  \left(\frac{\AV(t)}{N}\right)^{\frac{2s}{2s+d}}
  \le
  R^*_{\mathrm{loc}}(t;N)
  .
\end{equation}
For the same benchmark, the stabilized local-polynomial estimator gives
\begin{equation}\label{eq:minimax-upper-corrected}
  R^*_{\mathrm{loc}}(t;N)\le C_1\min\!\left\{B(t)^2,
  B(t)^{\frac{2d}{2s+d}}
  \left(\frac{\AV(t)}{N}\right)^{\frac{2s}{2s+d}}
  +B(t)^2\left(\frac{\log N}{N}\right)^{2s/d}\right\}.
\end{equation}
Here $c_1,C_1>0$ depend on $s,d$, the finite-atlas and boundary-Gram
constants, chart geometry, design constants, and noise-comparability
constants, but not on $N$. For fixed $t$ with $\AV(t)>0$, and sufficiently
large $N$, the coverage term is lower order and the usual upper rate is
recovered. If these constants are uniform on a compact time interval, the
threshold can be chosen uniformly. The statement is not uniform as
$\AV(t)\to0$ and is not a minimax claim for the full ambient flow-matching
observation model.
\end{theorem}

Theorem~\ref{thm:minimax} characterizes an intrinsic compact local benchmark.
The Gaussian-subclass lower bound and stabilized upper bound have the same
classical exponent only in the fixed-positive-noise, coverage-sufficient
regime; the finite-sample coverage term is essential at low noise. Exact
$\AV(t)$-normalized parametric scaling is proved only for the separated
Gaussian-mixture residual benchmark below. The proof and scope discussion are given in
Appendix~\ref{sec:proof-theorem1}.

\subsection{Interpretive Extension: Adaptive Rates for Structured Data}
\label{sec:adaptive-interpretation}

We use
the effective-rate vocabulary only to interpret structured data: if an
observed exponent in
$R(t;N)=\AV(t)\Theta(N^{-\alpha(t)})$ differs from the worst-case
Theorem~\ref{thm:minimax} exponent, Appendix~\ref{app:adaptive-conjecture}
describes the corresponding effective smoothness and dimension notation.
Theorems~\ref{thm:minimax} and~\ref{thm:gmm} stand independently of this
interpretive picture.

\paragraph{Merge-time vocabulary.}\label{sec:tmerge}
In the isotropic Gaussian-mixture setting of Theorem~\ref{thm:gmm}, let
$\Delta_{\min}=\min_{j\ne k}\|\mu_j-\mu_k\|$ and write
$\Delta_{\mathrm{eff}}(t)=(1-t)\Delta_{\min}/\sigma_t$ with
$\sigma_t^2=(1-t)^2\sigma^2+t^2$, and let
$t_{\mathrm{merge}}(\tau)=\inf\{t:\Delta_{\mathrm{eff}}(t)<\tau\}$.
In this separated GMM regime, below the threshold used in
Theorem~\ref{thm:gmm}, posterior mass concentrates on one mode and the Bayes
velocity is locally affine; above it, multiple affine components may mix. For
general multimodal distributions, we use ``merge time'' only as a heuristic
proxy for the point at which single-cell posterior dominance breaks down. The
rigorous posterior and entropy statements are in
Appendix~\ref{sec:proof-theorem2prime}.

\subsubsection{Oracle Result for Gaussian Mixtures}
\label{sec:thm2prime}

We now instantiate the structured picture for isotropic Gaussian mixtures in
the separated regime. The information set matters: in supervised flow
matching, observing $(x_t,u_t,t)$ also determines $a=x_t-tu_t$. The theorem
therefore studies a label-oracle residual-regression benchmark after
conditioning on separated mixture cells, not a full-information lower bound
for all action-aware estimators. Appendix~\ref{app:action-oracle} records the
stronger label-and-action oracle experiment.

\begin{theorem}[Label-Oracle Residual Benchmark for Separated Gaussian Mixtures]
\label{thm:gmm}
Let $p_{\mathrm{data}} = \sum_{k=1}^{K} \pi_k\,\mathcal{N}(\mu_k, \sigma^2 I_d)$
with minimum separation $\Delta_{\min}$ and diameter
$\Delta_{\max} \coloneqq \max_{j \neq k}\|\mu_j - \mu_k\|$, and let
$\Delta_{\mathrm{eff}}(t)=(1-t)\Delta_{\min}/\sigma_t$,
where $\sigma_t^2 = (1-t)^2 \sigma^2 + t^2$, and define
$\tau_{\mathrm{sep}}
\coloneqq C_{\mathrm{sep}}\sqrt{\log K+\log(1/\pi_{\min})}$.
Assume that $\Delta_{\mathrm{eff}}(t) \ge \tau_{\mathrm{sep}}$
for the time $t$ under consideration. Then there exist constants
$c_1, C_1, C_N > 0$ depending on $d, \sigma, \pi_{\min}$, and the
aspect ratio $\Delta_{\max}/\Delta_{\min}$ (but not on $N$) such that, for a
label-oracle residual-regression estimator given the true component labels and
the common variance $\sigma^2$:

\begin{enumerate}[label=(\alph*)]
\item \textbf{Velocity field structure.}
On a high-probability event (Lemma~\ref{lem:mode-dominance}), the posterior
concentrates on a dominant component and the Bayes-optimal velocity is
locally affine,
\begin{equation}\label{eq:gmm-local-affine}
    v^*(x_t, t)
    \;=\;
    A_{k^*}(t)\,x_t + b_{k^*}(t)
    + \varepsilon_{\mathrm{aff}}(x_t,t),
\end{equation}
with $A_k(t), b_k(t)$ as in Section~\ref{sec:sigma-gmm} and
$\|\varepsilon_{\mathrm{aff}}\|\le
C K e^{-c\,\Delta_{\mathrm{eff}}(t)^2}\Delta_{\max}/\sigma_t$
(Appendix~\ref{sec:proof-theorem2prime}).

\item \textbf{Label-oracle residual upper bound.}
For sample sizes
$N \ge C_N\pi_{\min}^{-1}(d+\log K)$, there exists a label-oracle residual estimator $\hat v_N^{\mathrm{orc}}$ with
\begin{equation}\label{eq:gmm-oracle-upper}
    R_{\mathrm{orc}}(t;N)
    \;\le\;
    C_1 \frac{K}{N}\AV(t)
    +
    \varepsilon_{\mathrm{sep}}(t)
    +
    \varepsilon_{\mathrm{count}}(t,N),
\end{equation}
where one can take
$\varepsilon_{\mathrm{sep}}(t)$ and $\varepsilon_{\mathrm{count}}(t,N)$ as
squared-risk terms exponentially small in the squared effective separation and
per-component sample count, respectively (Appendix~\ref{sec:proof-theorem2prime}).

\item \textbf{Separated residual-subproblem lower bound.}
For the corresponding separated-regime residual Gaussian-location subproblem,
\begin{equation}\label{eq:gmm-residual-lower}
    R_{\mathrm{sub}}^*(t;N)
    \;\ge\;
    c_1 \frac{K}{N}\AV(t).
\end{equation}
\end{enumerate}
Consequently, within this label-oracle residual-regression separated
subproblem, the upper bound and lower bound identify $K\,\AV(t)/N$ as the
parametric residual-regression scale, up to exponentially small separation and
count corrections.
\end{theorem}

The full proof is given in Appendix~\ref{sec:proof-theorem2prime}; the
extension of this picture to general (non-GMM) distributions is discussed in
Appendix~\ref{sec:general-adaptive}.

\subsection{Curve Collapse: A Normalization Principle}
\label{sec:normalized-collapse}

The following conditional factorization is the paper's directly testable
normalization principle.

\begin{proposition}[Regime-Wise Normalized Collapse]
\label{prop:normalized-collapse}
Define the normalized excess risk $\bar R(t;N) \coloneqq R(t;N)/\AV(t)$.
Suppose that on a time regime $I$ the excess risk factorizes as
\[
R(t;N) = \AV(t)\, g_I(N)\, \eta_I(t,N),
\qquad t \in I,
\]
where the exponent of $g_I(N)$ is independent of $t$ on $I$, and
$\eta_I(t,N)$ is either bounded above and below by positive constants
or differs from $1$ by at most a controlled small correction.
Then $\bar R(t;N)=g_I(N)\,\eta_I(t,N)$ for $t\in I$.
In particular, if $g_I(N)=\Theta(N^{-\alpha_I})$, then
$\bar R(t;N)=\Theta(N^{-\alpha_I})$
uniformly in $t \in I$ up to the residual variation in $\eta_I(t,N)$.

For the Gaussian-mixture separated regime, the residual variation is
exponentially small in $\Delta_{\mathrm{eff}}(t)^2$.
For the minimax baseline (Theorem~\ref{thm:minimax}), the collapse is only
approximate in the bounded-factor sense: the exponent $\alpha = 2s/(2s+d)$
is the same for all $t$, but the prefactor varies by up to
$(\sigma_u/\sigma_\ell)^{d/(2s+d)}$ across the evaluation interval (see
Remark~\ref{rem:factorization-constants}).
\end{proposition}

Thus Theorem~\ref{thm:gmm} gives exact collapse only for the separated
label-oracle residual-regression benchmark, while Theorem~\ref{thm:minimax}
motivates only bounded-factor common-exponent behavior. Regime-by-regime
factorization and reference-profile estimation are discussed in
Appendix~\ref{sec:proof-normalized-collapse}.


\section{Experiments}
\label{sec:experiments}

The experiments follow the claim hierarchy above: a controlled
Gaussian-mixture transition tests the predicted move from noisy overlap to
parametric scaling and finite-capacity attenuation; oracle-to-blind and proxy
audits ask which diagnostics survive without analytic access; a local
controlled-cell stress test asks whether a frozen reducible-error proxy can
guide time sampling; and a natural-image feature-flow bridge on CIFAR-100
(Section~\ref{sec:cifar}) instantiates the same diagnostic on a standard
public benchmark with an explicit coarse-to-fine hidden-mode hierarchy.
A leakage-free BEHAVIOR-1K panel is retained as application-context
diagnostics in Appendix~\ref{app:b1k-tasks}.

\subsection{Canonical Controlled Transition}
\label{sec:controlled-exp}

\paragraph{Controlled-core protocol.}
Our main display fixes a $d=4$, $K=4$, $\sigma=0.5$ isotropic Gaussian
mixture and varies geometric separation across $18$ values together with a
six-point demonstration-count ladder from $N=50$ to $N=2,000$. The primary
readout uses analytic $\AV(t)$ and the canonical mid-b reliability filter on
$t\in[0.05,0.80]$ with $\mathrm{SNR}\ge4$ and $R\ge0.016\,\AV$; full
settings, the complete separation ladder, and readout sensitivity are in
Appendix~\ref{sec:m2-dense-d4-appendix}. Post-hoc readouts from raw-positive
through very-strict filters preserve the same plateau-level conclusion while
making the low-overlap rows visibly threshold-sensitive
(Table~\ref{tab:m2-dense-d4-sensitivity}); the decoupled SNR-by-$R/\AV$
threshold-grid audit is reported in
Table~\ref{tab:m2-dense-d4-threshold-grid}.

\begin{figure}[t]
  \centering
  \safeincludegraphics[width=1\linewidth]{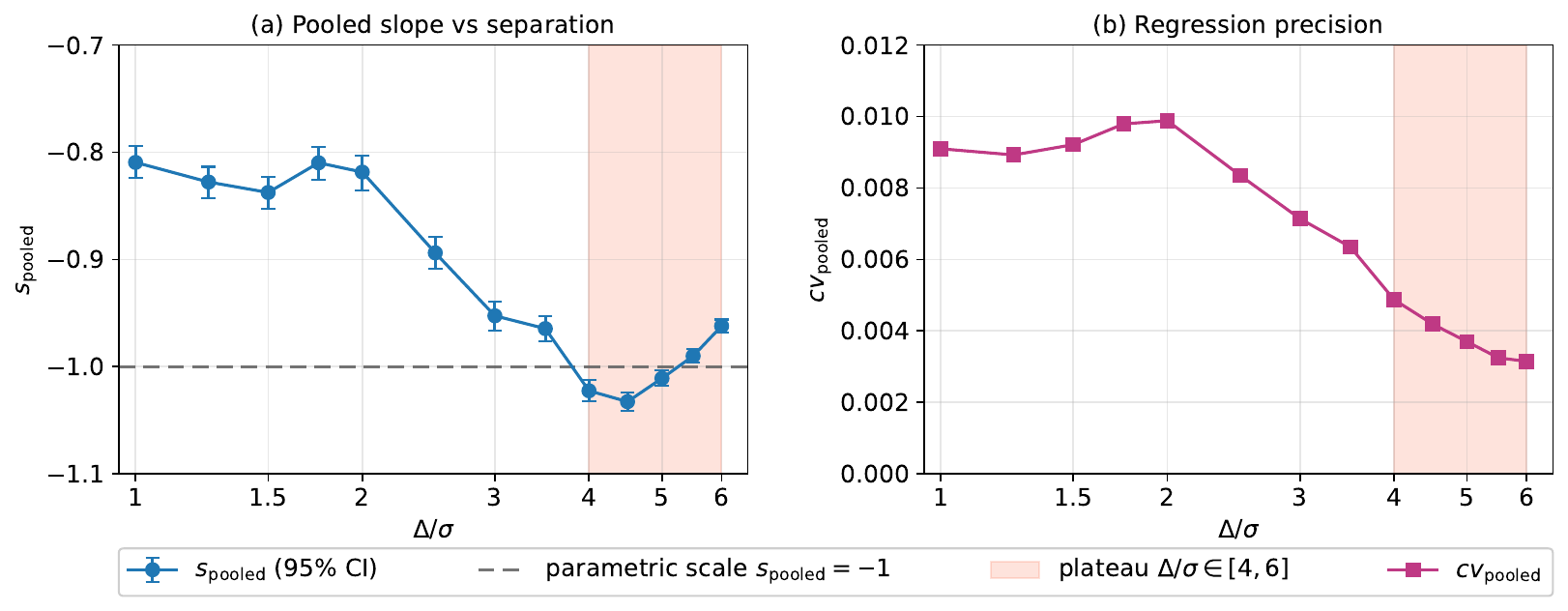}
  \caption{\textbf{Canonical $d{=}4$ dense controlled transition} on a
    $K{=}4$, $\sigma{=}0.5$ Gaussian mixture. The $13$-row front/plateau
    window shows the reliability-filtered signed log-log slope
    $s_{\mathrm{pooled}}=\partial\log \bar R/\partial\log N$ and regression
    precision; the shaded band marks $\Delta/\sigma\in[4,6]$, where
    $s_{\mathrm{pooled}}\approx -1$, equivalently the positive exponent
    $\alpha_{\mathrm{pooled}}=-s_{\mathrm{pooled}}\approx 1$. Full rows and
    filter sensitivity are in Appendix~\ref{sec:m2-dense-d4-appendix}.}
  \label{fig:m2-dense-d4}
\end{figure}

\paragraph{Dense transition diagram.}
Under the canonical mid-b readout, the dense sweep has three regimes. Across
the five sampled plateau rows between $\Delta/\sigma=4$ and $6$, the signed
slope $s_{\mathrm{pooled}}$ stays within roughly four percent of $-1$
(equivalently $\alpha_{\mathrm{pooled}}\approx 1$), so we describe the
plateau as \emph{consistent with the parametric scale}. At
$\Delta/\sigma=15$ the signed slope attenuates to $s_{\mathrm{pooled}}=-0.605$
($\alpha_{\mathrm{pooled}}=0.605$), consistent with finite-capacity bias
becoming comparable to the $C(t)/N$ variance term. At low overlap
($\Delta/\sigma=1.5$), the signed-slope magnitude shifts from about $0.98$
under the loosest readout to about $0.46$ under the tightest. 
Therefore, this region
is reported as low-SNR abstain-threshold sensitivity rather than an
unconditional overlap-regime claim. \begin{wraptable}[10]{r}{0.5\linewidth}
  \centering
  \small
   \setlength{\tabcolsep}{6pt}
  \caption{\textbf{Oracle-to-blind decomposition.} Per-row bridge
    audit metrics on five controlled GMM rows; column definitions
    in main text.}
  \label{tab:m6-bridge}
       \vspace{0.2cm}
  \begin{tabular*}{\linewidth}{@{\extracolsep{\fill}}lccc@{}}
    \toprule
    Row & $A_1$ dev (\%) & $z$ & $\epsilon_{\mathrm{MLP}}$ (\%) \\
    \midrule
    $d{=}2$ sep   & $-0.006$ & $0.27$ & $1.57$ \\
    $d{=}2$ p.m.  & $+0.068$ & $2.08$ & $0.64$ \\
    $d{=}4$       & $-0.021$ & $1.42$ & $1.51$ \\
    $d{=}8$       & $+0.024$ & $2.53$ & $1.49$ \\
    $d{=}16$      & $+0.005$ & $0.84$ & $1.71$ \\
    \bottomrule
  \end{tabular*}
\end{wraptable}
Complete rows, bootstrap intervals, and
normalization audits are in Appendices~\ref{sec:m2-dense-d4-appendix}
and~\ref{app:m4-norm-sanity}.

\subsection{Oracle-to-Blind and Proxy Calibration}
\label{sec:proxy-bridge}

\paragraph{Oracle-to-blind decomposition.}
The controlled transition above uses analytic $\AV(t)$ and a known mixture
structure. A separate five-row bridge audit compares the label-and-action
true-label ordinary least squares (OLS) plug-in estimator,
blind EM-OLS, and blind neural
readouts on a shared early-time window (Appendix~\ref{app:m6a-cr-sanity}).
Here blind EM-OLS fits a Gaussian mixture to unlabeled actions by
expectation-maximization (EM)~\citep{dempster1977em}, then uses the fitted
mixture in a closed-form velocity predictor that averages over posterior
component probabilities.
Table~\ref{tab:m6-bridge} reports the
empirical label-cost deviation $A_1$, its Bonferroni-paired statistic $z$,
and the blind MLP residual $\epsilon_{\mathrm{MLP}}$. The label-cost identity
holds within $0.07\%$ on every row, with a maximum $z=2.53$; the blind MLP
residual lies in $[0.64\%,1.71\%]$ and remains in $[1.57\%,2.06\%]$ under a
$d{=}8$ capacity ablation. This five-row audit therefore serves as a
bridge to marginal-Bayes attainability up to a bounded MLP residual, not as a blind
Cram\'er-Rao rate claim.

\paragraph{Proxy floor audit.}
When the analytic floor $\AV(t)$ is unavailable, a large-$N$
empirical reference $L_{\mathrm{largeN}}(t)$ supplies a
regime-dependent substitute, defining the proxy
$\hat R := L_N - L_{\mathrm{largeN}}$. We audit
$\rho_{\mathrm{Spearman}}(\hat R, R)$ across seventeen controlled
cells (Appendix Table~\ref{tab:m5-proxy}). Guard-passing cells retain high
rank agreement ($\rho_{\mathrm{Spearman}}\in[0.959,1]$), and the factor
diagnostic
$r_F := \max|L_{\mathrm{largeN}}-\AV|/\mathrm{std}(R \mid t\in[0.4,0.8])$
separates them from the isolated high-$r_F$ $d{=}2$ separated breakdown
documented in Appendix~\ref{app:m5-d2-limitation}. We therefore treat
$\hat R$ as a calibrated proxy under the empirical guard $r_F<2$, not as a
universal replacement for the analytic floor.

\subsection{Controlled Time-Sampler Stress Test}
\label{sec:envelope}

\paragraph{Setup.}
In a balanced partial-merge Gaussian mixture with $d=16$, $K=32$, and a
uniform-time baseline, we test whether changing the training-time sampler
$q(t)$ can reduce the remaining excess above the analytic floor
($L/\AV=1.037$ at the plateau). The proxy
$\hat R_{\mathrm{ind}}(t)$ is frozen from source baseline runs whose
demonstrations and evaluation actions are disjoint from the paired test; all
comparisons are paired and use sign-permutation tests.

\paragraph{Sampler-side envelope.}
A categorical $t$-sampler $q(t)$ on the same fine grid replaces the
training-time $\mathcal{U}(0.05, 0.95)$ with either a coverage-preserving
mixture
$q_\lambda = (1-\lambda)\,\mathcal{U} +
\lambda\,\hat R_{\mathrm{ind}}/\textstyle\sum\hat R_{\mathrm{ind}}$
or a pure sharpening
$q\propto \hat R_{\mathrm{ind}}^\alpha$, in the spirit of loss-aware
diffusion training schedules~\citep{hang2023minsnr,choi2022p2,esser2024sd3}.
The frozen proxy is reused without modification; the per-batch loss is the
standard mean-squared error (MSE) without importance correction, and sampler guards, including an
effective-sample-size floor~\citep{kong1994sequential}, enforce nondegenerate probability mass functions
(PMFs). All tested gap arms improve over the paired baseline with negative
paired seed-level reducible-loss deltas for every seed and
$p_{\mathrm{perm}}<0.003$~\citep{phipson2010permutation}; plateau gains span
$9.32\%$ to $20.21\%$; the complete paired table is in
Appendix~\ref{app:m9e-v5}. The learning curves in
Figure~\ref{fig:m9e-v4} show the peak-step ordering.

\paragraph{Synthesis.}
The effect strengthens along the mixture chain and peaks at pure $\alpha=2$
on the plateau and $\alpha=4$ near the peak-learning step; in the canonical
cell no tested arm reverses sign. Appendix~\ref{app:m9e-v5} adds a three-cell
$d$-axis sensitivity check: mixture arms and pure $\alpha\in\{1,2\}$ improve
across all cells, pure $\alpha=4$ weakens at the largest cell, and the most
aggressive pure sampler ($\alpha=8$) reverses in the late plateau there.

\begin{figure}[h]
  \centering
  \safeincludegraphics[width=\linewidth]{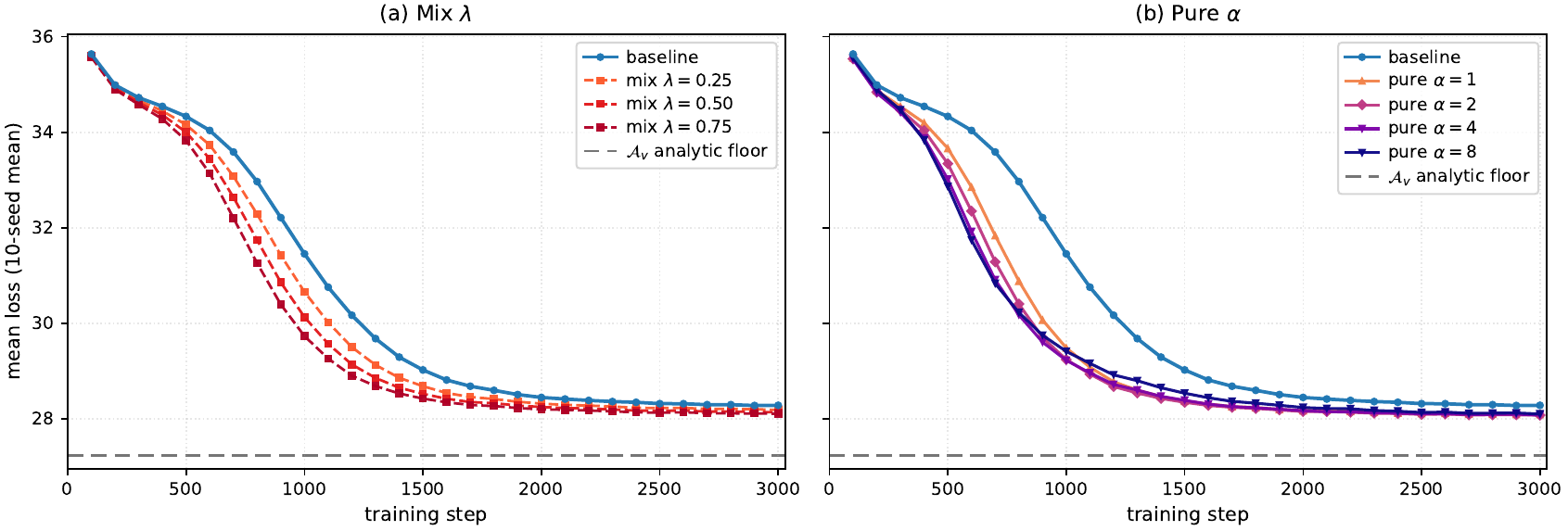}
  \caption{\textbf{Sampler-side $t$-sampler interventions on a
    controlled $d=16$, $K=32$ GMM.} Ten-seed mean loss for the baseline and
    gap-guided samplers using a proxy source disjoint from the paired test.
    Gap arms lie below baseline past the peak-learning step; pure
    $\alpha=2$ is lowest at the plateau, while pure $\alpha=4$ is lowest
    near the peak-learning step.}
  \label{fig:m9e-v4}
\end{figure}

\subsection{Natural-Data Application: CIFAR-100 DINOv2 Feature-Flow}
\label{sec:cifar}

\paragraph{Setup and training result.}
To test whether the diagnostic transfers beyond the controlled mixture, we
use the official CIFAR-100 split and extract frozen DINOv2 ViT-S/14
final-LayerNorm CLS features~\citep{oquab2023dinov2}. The resulting $384$-dimensional vectors are
centered and globally RMS calibrated using all $50,000$ official training rows;
no PCA, whitening, learned projection, or pixel decoder is used. Each target
seed uses $25$ examples per fine class ($2,500$ labeled examples total). We train a
fine-label-conditioned flow-matching field in this feature space and compare
uniform time sampling with VAP-guided reducible-error sampling using paired
training seeds. The guided arm uses $q_4(t)\propto\widehat R(t)^4$ on $64$
categorical bins without importance correction. On a uniformly weighted
$25$-time evaluation grid, the guided field has $6.78\%$ lower velocity loss at
the common step $11,500$, with all five paired seeds favoring the guided arm.
\begin{wrapfigure}[20]{r}{0.5\linewidth}
  \vspace{0.3cm}
  \centering
  \includegraphics[width=\linewidth]{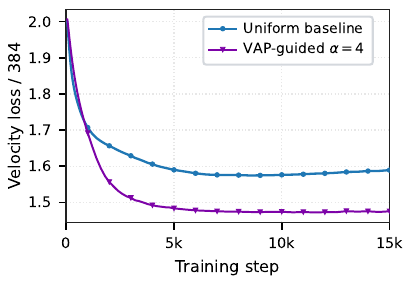}
  \vspace{-0.5cm}
  \caption{\textbf{Uniform-time evaluation during DINOv2 feature-flow training.}
    The gap-guided sampling lowers uniformly evaluated velocity loss.
    Curves show means over five paired training seeds.}
  \label{fig:cifar-dinov2-training}
\end{wrapfigure}
This result supports the transfer of the time-local ambiguity diagnostic to a
natural feature distribution while remaining a feature-space statement
rather than an image-generation claim.

\paragraph{Closed-loop feature-space deployment.}
We integrate each learned field as an ODE generator from the same initial
noise and evaluate the resulting feature distributions against held-out
CIFAR-100 features. Global feature FD falls by $23.6\%$, alongside gains in
macro precision, macro density, macro coverage, and fine-label macro accuracy
(Table~\ref{tab:cifar-dinov2-summary}). Macro conditional KID shows no
improvement, and both improved recall and effective covariance rank decrease,
indicating a diversity cost alongside the gains in FD and label accuracy.
These are five paired training seeds and feature-space
metrics; feature FD is not image FID, and the experiment does not establish
pixel-space realism or a universal sampler.
\begin{table}[!htb]
  \centering
  \small
  \setlength{\tabcolsep}{4pt}
  \caption{\textbf{CIFAR-100 DINOv2 feature-flow summary.} Values are
    aggregate deployment metrics over five paired training seeds. Arrows
    indicate the preferred direction.}
  \label{tab:cifar-dinov2-summary}
  \begin{tabular}{@{}lccc@{}}
    \toprule
    Metric & Uniform & VAP-guided & Change \\
    \midrule
    Global feature FD $\downarrow$ & $33.333$ & $\mathbf{25.476}$ & $-23.6\%$ \\
    Macro precision $\uparrow$ & $0.151$ & $\mathbf{0.690}$ & $+0.539$ \\
    Macro density $\uparrow$ & $0.122$ & $\mathbf{1.226}$ & $+1.104$ \\
    Macro coverage $\uparrow$ & $0.234$ & $\mathbf{0.728}$ & $+0.494$ \\
    Fine-label macro accuracy $\uparrow$ & $0.791$ & $\mathbf{0.859}$ & $+6.76$ pp \\
    Macro conditional KID $\downarrow$ & $0.1988$ & $0.2062$ & $+0.0074$ \\
    Improved recall $\uparrow$ & $0.722$ & $0.221$ & $-0.501$ \\
    Effective covariance rank $\uparrow$ & $47.25$ & $40.14 $& $-7.11$ \\
    \bottomrule
  \end{tabular}
\end{table}

The deployment comparison uses a frozen checkpoint protocol and paired
initialization. Each method generates $6,000$ features, compared with a disjoint
$6,000$-feature held-out test reference library. Checkpoint-selection details,
per-seed trajectories, solver checks, and the full conditional tables are deferred to
Appendix~\ref{app:cifar}. The natural-data result therefore strengthens the
case for VAP-guided time allocation through improvements in the training
objective, global feature FD, and label accuracy, with the conditional-distribution
and diversity limitations quantified above.


\section{Related Work}\label{sec:related}

\paragraph{Flow matching, imitation learning, and schedules.}
Flow matching and rectified-flow objectives transport noise to data
\citep{lipman2023flow,liu2023flow,albergo2023building,tong2024improving,pooladian2023multisample,feng2026multi}.
Imitation scaling and behavioral-cloning theory study data composition and
policy error~\citep{hu2025data,dass2025datamil,ross2010efficient,ross2011dagger,foster2024bc,block2023provable},
while schedule methods change the training-time distribution
\citep{karras2022edm,hang2023minsnr,choi2022p2,esser2024sd3}. VAP instead
diagnoses the induced regression problems along interpolation time.

\paragraph{Statistical theory and adaptive structure.}
Diffusion and score theory characterize minimax rates under manifold and
smoothness assumptions~\citep{oko2023,weber2024,chen2023sampling,debortoli2023convergence,lee2022convergence},
and adaptive regression theory studies intrinsic or compositional structure
\citep{schmidt2020nonparametric,suzuki2019adaptivity,nakada2020adaptive,pope2021intrinsic}.
Our intrinsic result is a time-resolved velocity analog; the separated
Gaussian-mixture result is the scoped label-oracle residual benchmark.



\section{Conclusion}\label{sec:conclusion}

VAP turns the aggregate flow-matching loss into a time-resolved diagnostic by
separating the Bayes-risk floor from reducible excess error. The theory gives
two scoped benchmarks: a local intrinsic regression rate with bounded-factor
time dependence and a sharper separated-GMM label-oracle residual scale
$K\AV(t)/N$. Controlled experiments validate the analytic floor,
    separated-regime collapse prediction, oracle-to-blind calibration, proxy-floor
    guards, and local time-sampler intervention, whose sensitivity boundary is
    given in Appendix~\ref{app:m9e-v5}.
On CIFAR-100 DINOv2 feature flow, guided time sampling reduces uniformly
evaluated velocity loss by $6.78\%$ and global feature FD by $23.6\%$.
Macro conditional KID shows no gain, while lower recall and effective
covariance rank reveal a diversity cost. Turning these diagnostics into a budget rule or intervention policy remains open.

\paragraph{Limitations.}
The intrinsic result is for the local benchmark
$\mathcal{G}_t^{\mathrm{comp}}$, not the full ambient observation, with $\AV(t)$
entering as a bounded-factor prefactor. The GMM result is a separated
label-oracle residual benchmark, not a full-information lower bound because
$(x_t,u_t,t)$ determines $a=x_t-t u_t$. The natural-data result is limited to the 
frozen DINOv2 feature space: it makes no pixel-quality, universal-sampler, or
uniform-metric-improvement claims.

\begin{ack}
The work was supported in part by the National Natural Science Foundation of China under Grants U23B2059 and 62688201, the BIT Research and Innovation Promoting Project under Grant 2025YCXZ033, and the Beijing Natural Science Foundation under Grant QY26254. 
\end{ack}

\bibliography{references}
\bibliographystyle{abbrvnat}

\newpage
\appendix
\hypersetup{linkcolor=black, citecolor=mydarkblue, urlcolor=black}
\renewcommand{\appendixpagename}{\centering \LARGE Appendix}
\appendixpage

\startcontents[section]
\printcontents[section]{l}{1}[2]{}

\paragraph{Appendix Roadmap.}
The appendix follows the paper's claim hierarchy:
\begin{itemize}[leftmargin=*, itemsep=1pt]
\item \textbf{Intrinsic benchmark.} Appendix~\ref{sec:proof-theorem1}
states the assumptions and proves the time-resolved minimax benchmark behind
Theorem~\ref{thm:minimax}.
\item \textbf{Separated GMM benchmark.} Appendix~\ref{sec:proof-theorem2prime}
proves Theorem~\ref{thm:gmm} through mode dominance, the label-oracle residual
upper bound, and the residual lower bound.
\item \textbf{Curve collapse.} Appendix~\ref{sec:proof-normalized-collapse} derives the
normalized-collapse prediction used by the controlled Gaussian-mixture
experiments.
\item \textbf{Experimental protocols.} Appendix~\ref{app:implementation-details}
gives the controlled protocols and audit checks behind the main figures.
\item \textbf{Bridge, proxy, and sampler auxiliary audits.}
Appendices~\ref{app:m5-d2-limitation}, \ref{app:m7-heterogeneity},
\ref{app:m9e-v5}, \ref{app:bridge-aux}, \ref{app:m6a-cr-sanity},
\ref{app:blind-attain-panel}, and~\ref{app:two-stage-sweep} record
proxy-floor and observable-guard calibration, controlled heterogeneity,
cross-cell sampler sensitivity, oracle-to-blind sanity checks,
blind-attainability, and the two-stage attainability sweep.
\item \textbf{Application materials.} Appendix~\ref{app:b1k-tasks} gives the
BEHAVIOR-1K task context used to interpret the diagnostic panel.
\end{itemize}

\hypersetup{linkcolor=mydarkblue, citecolor=mydarkblue, urlcolor=mydarkblue}
\vspace{20ex}

\newpage


\section{Basic VAP Facts and Intrinsic Benchmark Proofs}
\label{sec:proof-theorem1}

This appendix collects: (\S\ref{app:thm1-assumptions}) the formal statement of
the five assumptions, Assumptions~\ref{asm:manifold}-\ref{asm:compact-design},
used by Theorem~\ref{thm:minimax};
(\S\ref{app:loss-decomposition-proof}) the proof of
Proposition~\ref{prop:decomposition} (loss decomposition);
(\S\ref{app:sigma-properties-proof}) the proof of
Proposition~\ref{prop:sigma-properties} (boundary behavior and non-monotonicity
of $\AV$); (\S\ref{app:a4-sufficient}) verifiable sufficient conditions
for Assumption~\ref{asm:noise-comparability}, used implicitly throughout the
proof of Theorem~\ref{thm:minimax}; (\S\ref{app:thm1-dim-discussion})
expanded discussion of the intrinsic-vs-ambient dimension distinction; and
the main minimax proof itself.

Throughout this appendix we follow the conditional convention of
\S\ref{sec:sigma-def}: a fixed context $c$ is implicit and suppressed from
notation, with all expectations and variances conditional on $c$.

We prove the time-resolved baseline result at fixed time $t$ by analyzing the
local heteroscedastic regression benchmark $\mathcal{G}_t^{\mathrm{comp}}$ induced by the
flow matching geometry (Lemma~\ref{lem:local-reduction}).
The lower bound uses an Assouad construction within $\mathcal{G}_t^{\mathrm{comp}}$, and the
upper bound uses kernel regression with the standard bias-variance tradeoff.
Because $\AV(t)$ is only converted into a multiplicative prefactor after
restricting to compact time intervals on which it is bounded above and below,
the result should be read as a bounded-factor time-resolved minimax characterization.

\subsection{Assumptions for Theorem~\ref{thm:minimax}}
\label{app:thm1-assumptions}

\begin{assumption}\label{asm:manifold}
The data distribution $p_{\mathrm{data}}$ is supported on a compact
$d$-dimensional $C^2$ submanifold $\mathcal{M} \subset \mathbb{R}^D$ with positive reach~\citep{federer1959curvature,niyogi2008finding} and with
$0 < c_\ell \leq p_{\mathrm{data}}(a) \leq c_u$ in local coordinates on $\mathcal{M}$.
Note that for $t > 0$, the marginal $p(x_t \mid t)$ has full
$D$-dimensional support due to the Gaussian noise $\varepsilon$;
however, the regularity of the Bayes-optimal velocity $v^*(\cdot, t)$ is
governed by the $d$-dimensional intrinsic structure of~$\mathcal{M}$
(Assumption~\ref{asm:smoothness}).
\end{assumption}

\begin{assumption}\label{asm:smoothness}
Fix a finite atlas with chart domains $Q_j\subset\mathbb R^d$ and fixed
compact neighborhoods $Q_j^+\supset Q_j$. For each
$t\in[\delta,1-\delta]$, let $f_{t,j}:Q_j\to\mathbb R^D$ denote the
coordinate representation of the Bayes velocity on chart $j$. For a fixed
finite $s>0$, write $m=\lceil s\rceil-1$ and $\beta=s-m\in(0,1]$ (with
$C^{s-1,1}$ when $s$ is an integer). There are extensions
$\widetilde f_{t,j}\in C^{m,\beta}(Q_j^+;\mathbb R^D)$ satisfying
\begin{equation}\label{eq:finite-atlas-smoothness}
\sup_{t\in[\delta,1-\delta]}\max_j
\|\widetilde f_{t,j}\|_{C^{m,\beta}(Q_j^+;\mathbb R^D)}
\le B_\delta<\infty.
\end{equation}
The chart overlap, cutoff, and boundary-Gram constants are fixed uniformly
on this compact time interval. This is an explicit finite-smoothness
condition on the compact local benchmark. It is not a global ambient bound,
not a consequence of Gaussian smoothing, and does not assert $s=\infty$.
For a Gaussian mixture, the field is smooth and hence satisfies this
condition for every fixed finite $s$ on each bounded readout neighborhood;
the parametric result for that model instead comes from the separated
finite-dimensional residual theorem~\ref{thm:gmm}.
\end{assumption}

\begin{assumption}[Compact random design and boundary Gram]
\label{asm:compact-design}
For every chart and every admissible bandwidth $h\le h_0$, the intrinsic
design density is bounded above and below on the compact chart domain. The
population moment matrix of the degree-$m$ local-polynomial basis has a
uniformly positive minimum eigenvalue, including boundary queries after the
fixed cutoff and stabilization rule. The atlas has finite overlap and the
cutoffs have a positive boundary margin. These are benchmark assumptions on
the random-design experiment, not consequences of Assumption~\ref{asm:manifold}.
\end{assumption}

\begin{assumption}\label{asm:bounded-sigma}
On every compact subinterval $[\delta, 1-\delta] \subset (0,1)$,
the velocity ambiguity profile satisfies
$0 < \sigma_\ell(\delta) \leq \AV(t) \leq \sigma_u(\delta) < \infty$
for all $t \in [\delta, 1-\delta]$.
\end{assumption}

Assumption~\ref{asm:bounded-sigma} is a compact-interval regularity condition: it holds whenever $p_{\mathrm{data}}$
has finite second moments and the conditional velocity problem is non-degenerate on the interval.
It ensures that the noise level $\AV(t)$ neither vanishes nor diverges
on compact time intervals, which is needed for interval-uniform control of
the time-resolved baseline.

\begin{assumption}\label{asm:noise-comparability}
\emph{Uniform noise comparability.}
For each compact subinterval $[\delta, 1-\delta] \subset (0,1)$,
the pointwise noise variance satisfies
\[
    c_n\,\AV(t) \;\leq\; \mathrm{tr}\,\mathrm{Cov}[u_t \mid x_t, t]
    \;\leq\; C_n\,\AV(t)
    \qquad \text{for $p_{x_t}$-a.e.\ } x_t,
\]
for constants $0 < c_n \leq C_n$ depending only on $\delta$ and $p_{\mathrm{data}}$.
\end{assumption}

Assumption~\ref{asm:noise-comparability} requires a uniform comparison between
the pointwise conditional noise level and its average value $\AV(t)$.
This is a structural assumption, not automatic from compact support; it holds
in the separated Gaussian-mixture regime, where the conditional covariance is
asymptotically spatially uniform up to exponentially small corrections
(Remark~\ref{rem:sigma-behavior}), and can fail for distributions with
bridge-like multimodal geometries where the posterior assigns weight to
multiple distant modes.
Verifiable sufficient conditions for
Assumption~\ref{asm:noise-comparability} beyond the Gaussian-mixture setting
are formalized in Appendix~\ref{app:a4-sufficient}.

\begin{remark}[Scope and interpretation]
\label{rem:thm1-scope}
Theorem~\ref{thm:minimax} characterizes the lower and corrected upper bounds
for the \emph{intrinsic compact local regression benchmark}
$\mathcal{G}_t^{\mathrm{comp}}$ induced by the stated benchmark assumptions,
not the minimax risk over the full ambient
distribution class.
This parallels the approach of \citet{oko2023} for score
estimation, with the key difference being the \emph{time-resolved}
perspective: $\AV(t)$ enters as the noise level at each point along
the interpolation path.
A formal constructive reduction from the full ambient flow matching
observation model to $\mathcal{G}_t^{\mathrm{comp}}$ would require verifying that the
ambient estimator class projects faithfully onto the benchmark class; we
leave this extension to future work.
For structured data, the benchmark rate can be beaten: the Gaussian-mixture
theorem (Theorem~\ref{thm:gmm}) achieves a parametric rate in the separated
regime.
\end{remark}

\subsection{Proof of Proposition~\ref{prop:decomposition}}
\label{app:loss-decomposition-proof}

\begin{proof}
Expand $\|\hat v_N-u_t\|^2=\|\hat v_N-v^*+v^*-u_t\|^2$. The cross-term vanishes
because $v^*=\mathbb E[u_t\mid x_t,c,t]$, so
$\mathbb E[v^*-u_t\mid x_t,c,t]=0$.
\end{proof}

\subsection{Proof of Proposition~\ref{prop:sigma-properties}}
\label{app:sigma-properties-proof}

\begin{proof}[Proof of Proposition~\ref{prop:sigma-properties}]
(ii) At $t=1$, $x_1 = \varepsilon$ is independent of $a$, so
$p(a \mid x_1, 1) = p_{\mathrm{data}}(a)$ and
$\AV(1) = \mathrm{tr}\,\mathrm{Var}[a]$.

(iii) For a single Gaussian $\mathcal{N}(\mu, \sigma^2 I_d)$,
$\AV(t) = d\sigma^2/\sigma_t^2 = d\sigma^2/((1{-}t)^2\sigma^2 + t^2)$,
which attains a maximum at $t = \sigma^2/(1+\sigma^2) \in (0,1)$ and therefore first increases and then decreases.

(i) Under Assumption~\ref{asm:manifold}, $\mathcal{M}$ is compact, so
$\|a\| \leq R$ for some $R > 0$.  The posterior
$p(a \mid x_t, t) \propto p_{\mathrm{data}}(a)\,\phi_D(x_t; (1{-}t)a, t^2 I)$
concentrates in an $O(t/(1{-}t))$-neighborhood of the point
$a^* \in \mathcal{M}$ nearest to $x_t/(1{-}t)$ as $t \to 0^+$.
Since $\mathcal{M}$ is a $C^2$ submanifold and the density $p_{\mathrm{data}}$
is bounded above and below, a Laplace-type argument on the tangent space
shows that $\mathrm{tr}\,\mathrm{Var}[a \mid x_t, t] = O(t^2)$ for
$p_{x_t}$-a.e.\ $x_t$.  Consequently,
$\AV(t) = (1/t^2)\,\mathbb{E}_{x_t}[\mathrm{tr}\,\mathrm{Var}[a \mid x_t, t]]
= O(1)$, hence $\AV(0^+) < \infty$.
\end{proof}

\subsection{Verifiable sufficient conditions for Assumption~\ref{asm:noise-comparability}}
\label{app:a4-sufficient}

The proof below repeatedly invokes Assumption~\ref{asm:noise-comparability}
(uniform comparability of the pointwise conditional noise to its average
$\AV(t)$). For completeness we record verifiable sufficient conditions
under which Assumption~\ref{asm:noise-comparability} holds beyond the
Gaussian-mixture setting.

\begin{remark}[Verifiable sufficient conditions for Assumption~\ref{asm:noise-comparability}]
\label{rem:a4-sufficient}
A convenient sufficient condition for Assumption~\ref{asm:noise-comparability}
is the existence, on a compact interval $[\delta,1-\delta]$, of a scalar
profile $\nu(t)$ and constants $0<c_0\le C_0<\infty$ such that
\[
  c_0 \,\nu(t)
  \;\le\;
  \mathrm{tr}\,\mathrm{Cov}[u_t\mid x_t,t]
  \;\le\;
  C_0 \,\nu(t)
  \qquad \text{for $p_{x_t}$-a.e.\ } x_t,
\]
and such that $\AV(t)=\mathbb{E}[\mathrm{tr}\,\mathrm{Cov}[u_t\mid x_t,t]]$
is itself comparable to $\nu(t)$ on that interval.  In that case
Assumption~\ref{asm:noise-comparability} follows
immediately.

This sufficient condition is verifiable beyond the Gaussian-mixture setting in
at least two common situations: (i)~unimodal families for which the posterior
$p(a\mid x_t,t)$ is uniformly strongly log-concave and its Hessian is bounded
above and below on the relevant region, implying posterior covariance bounds
that are uniform in $x_t$; and (ii)~finite mixtures of such components away
from the merge region, where posterior mass concentrates on a single component
and the same covariance bounds are inherited componentwise.
By contrast, A4 is expected to fail in bridge-like multimodal geometries,
where a small subset of locations can carry posterior mass on multiple distant
modes and hence exhibit conditional variance much larger than the average
$\AV(t)$.
\end{remark}

\subsection{Intrinsic vs.\ ambient dimension}
\label{app:thm1-dim-discussion}

\begin{remark}[Intrinsic dimension vs.\ ambient dimension]
\label{rem:manifold-dimension}
Although $x_t$ lives in $\mathbb{R}^D$, the minimax rate depends on the
\emph{intrinsic} dimension~$d$ of~$\mathcal{M}$, not the ambient~$D$,
paralleling \citet{oko2023} for score estimation. Empirical estimates of the
intrinsic dimension of natural data~\citep{pope2021intrinsic,stanczuk2024diffusion}
support the gap between intrinsic $d$ and ambient $D$. The mechanism
is the $O(t)$-tubular concentration of $p(x_t\mid t)$ around $(1{-}t)\mathcal{M}$.
\end{remark}

For moderate $t$, the marginal density $p(x_t\mid t)$ concentrates in an
$O(t)$-tubular neighborhood of $(1{-}t)\mathcal{M}$, and the velocity field
$v^*$ varies primarily along the $d$ tangent directions of this manifold
rather than along all $D$ ambient directions. The benchmark class
$\mathcal{G}_t^{\mathrm{comp}}(s,d,B,\AV)$ constructed in
Lemma~\ref{lem:local-reduction} formalizes this intrinsic picture: both the
Assouad lower bound and the kernel-regression upper bound operate within
this $d$-dimensional benchmark, so the resulting minimax rate exponent
depends only on $d$. The gap between this intrinsic benchmark and the full
ambient flow matching observation model is discussed in
Remark~\ref{rem:thm1-scope}.

\subsection{Problem Setup}

Fix $t \in (0, 1-\delta]$.  Under the linear interpolation $x_t = (1-t)\,a + t\,\varepsilon$,
we observe $N$ i.i.d.\ triples $(x_t^{(i)}, u_t^{(i)}, t)$, where $u_t^{(i)} = \varepsilon^{(i)} - a^{(i)}$.
The regression model at time $t$ takes the form:
\begin{equation}\label{eq:regression-model-t1}
    u_t^{(i)} = v^*(x_t^{(i)}, t) + \xi^{(i)},
\end{equation}
where $v^*(x_t, t) = \mathbb{E}[u_t \mid x_t, t]$ is the Bayes-optimal velocity and
$\xi^{(i)} = u_t^{(i)} - v^*(x_t^{(i)}, t)$ is the residual noise satisfying
$\mathbb{E}[\xi^{(i)} \mid x_t^{(i)}, t] = 0$ and
$\mathbb{E}[\|\xi^{(i)}\|^2 \mid x_t^{(i)}, t] = \mathrm{tr}\,\mathrm{Cov}[u_t \mid x_t^{(i)}, t]$.

By Definition~\ref{def:sigma}, the average noise level is
$\mathbb{E}_{x_t}[\mathrm{tr}\,\mathrm{Cov}[u_t \mid x_t, t]] = \AV(t)$.
Thus \eqref{eq:regression-model-t1} is a heteroscedastic nonparametric regression
problem with average noise level $\AV(t)$.

The following lemma formalizes the local $d$-dimensional regression benchmark
induced by the ambient observation geometry.

\begin{lemma}[Benchmark construction via local coordinates]\label{lem:local-reduction}
Under Assumptions~\ref{asm:manifold}-\ref{asm:smoothness}, fix
$t \in [\delta, 1{-}\delta]$.  Let $\{(U_j, \phi_j)\}$ be a finite atlas
of $(1{-}t)\mathcal{M}$ and let $\mathcal{T}_j(t)$ denote the
$O(t)$-tubular neighborhood of $(1{-}t)U_j$.
Then the following properties define the compact benchmark class
$\mathcal{G}_t^{\mathrm{comp}}$:
\begin{enumerate}[label=(\roman*)]
\item The marginal density $p(x_t \mid t)$ restricted to $\mathcal{T}_j(t)$
  is bounded above and below by constants depending only on
  $c_\ell, c_u, t, \sigma$, and the chart Jacobian.
\item The velocity field $v^*(\cdot, t)$ expressed in the local coordinates
  $z = \phi_j(\cdot)$ has H\"{o}lder norm $\|v^*\|_{C^s} \leq B(t)$
  by Assumption~\ref{asm:smoothness}.
\item Under Assumption~\ref{asm:noise-comparability}, the conditional noise
  $\mathrm{tr}\,\mathrm{Cov}[u_t \mid x_t, t]$
  for $x_t \in \mathcal{T}_j(t)$ satisfies
  $c_n\,\AV(t) \leq \mathrm{tr}\,\mathrm{Cov}[u_t \mid x_t, t] \leq C_n\,\AV(t)$.
\end{enumerate}
Together, these properties specify a compact $d$-dimensional
heteroscedastic regression benchmark
$\mathcal{G}_t^{\mathrm{comp}}(s,d,B,\AV)$ with controlled design density,
smoothness, boundary Gram, and noise level.
Theorem~\ref{thm:minimax} establishes the lower and corrected upper bounds
of this benchmark;
a full constructive reduction from the ambient flow matching observation
model to $\mathcal{G}_t^{\mathrm{comp}}$ (i.e., showing that the ambient minimax risk is
of the same order) would require additional arguments about the faithfulness
of the tubular projection, which we leave to future work.
\end{lemma}

The minimax risk of the intrinsic benchmark is:
\begin{equation}\label{eq:minimax-nonparametric-def}
    R^*_{\mathrm{loc}}(t;\, N) \;\coloneqq\;
    \inf_{\hat{v}} \;\sup_{P \in \mathcal{G}_t^{\mathrm{comp}}(s,d,B,\AV)} \;
    \mathbb{E}_{\mathcal D_N,X\sim P_{x_t}}\!\left[\|\hat{v}_{\mathcal D_N}(X,t) - v_P^*(X,t)\|^2\right],
\end{equation}
where the infimum is over all estimators $\hat{v}$ based on the $N$ training samples.

\subsection{Lower Bound via Assouad's Method}

We adapt the classical Assouad lower bound \citep{tsybakov2009nonparametric,yang1999information}
to the associated local regression experiment, following the strategy of
\citet{stone1982optimal}. The role of Assumptions~\ref{asm:manifold} and
\ref{asm:smoothness} is to justify a local coordinate reduction in which the
Bayes velocity varies only along $d$ intrinsic directions while the ambient
Gaussian smoothing contributes a controlled noise term. We therefore construct
the least-favorable family inside the formal compact benchmark. The resulting lower bound is
a benchmark lower bound for this induced local experiment, not a lower bound
for the full ambient flow matching observation model.

\paragraph{Step 1: Local coordinate reduction.}
Fix a chart $\phi$ on a patch of the base manifold $\mathcal{M}$ and extend it
to a tubular-neighborhood parameterization around $(1-t)\mathcal{M}$.
Within this patch, the signal part of the regression problem is represented by
a $d$-dimensional coordinate $z = \phi(a)$ together with ambient Gaussian
smoothing whose effect is absorbed into the conditional noise level.
The lower bound is therefore proved for this local $d$-dimensional regression
experiment; the constants in the reduction depend on the chart Jacobian and the
tube geometry but not on $N$.

\paragraph{Step 2: Hypothesis construction in local coordinates.}
Choose a bandwidth $h > 0$ (to be optimized) and partition the coordinate
domain into $M \asymp (1/h)^d$ cubes of side length $h$.  Let $\{z_m\}_{m=1}^M$
denote the centers of these cubes.  For each binary vector
$\omega = (\omega_1, \ldots, \omega_M) \in \{0,1\}^M$, define the velocity
perturbation:
\begin{equation}\label{eq:assouad-hypotheses}
    v_\omega(x, t) = v_0(x, t) + c_0\,B(t)\,h^s \sum_{m=1}^{M} \omega_m\,\psi\!\left(\frac{\phi(x) - z_m}{h}\right) e_1,
\end{equation}
where $v_0$ is a baseline velocity field in $C^s(\mathcal{M}_t)$,
$\psi: \mathbb{R}^d \to \mathbb{R}$ is a smooth bump function supported on
$[-1/2, 1/2]^d$ with $\|\psi\|_{C^s} \leq 1$,
$e_1$ is a unit vector,
and $c_0 > 0$ is chosen small enough that $v_\omega \in C^s(\mathcal{M}_t)$
with norm bounded by $B(t)$ for all $\omega$.

By construction, the least-favorable family is defined directly inside the
benchmark class $\mathcal{G}_t^{\mathrm{comp}}(s,d,B,\AV)$, which is the object of
Theorem~\ref{thm:minimax}.

The $L^2$ separation between hypotheses differing in coordinate $m$ is:
\begin{equation}\label{eq:hypothesis-separation}
    \|v_\omega - v_{\omega^{(m)}}\|^2_{L^2(\mathcal{M}_t)}
    \;\geq\;
    c_1\,B(t)^2\,h^{2s}\,h^d
    \;=\;
    c_1\,B(t)^2\,h^{2s+d},
\end{equation}
where $\omega^{(m)}$ differs from $\omega$ only in coordinate $m$.

\paragraph{Step 3: KL divergence bound via Gaussian subclass.}
For the purpose of the lower bound, we restrict to the Gaussian-noise
subclass $\mathcal{G}_t^{\mathrm{Gauss}} \subset \mathcal{G}_t^{\mathrm{comp}}$ consisting
of benchmark experiments with conditionally Gaussian noise:
$\xi \mid x_t \sim \mathcal{N}(0,\, \sigma^2_{\mathrm{noise}}(x_t)\, I_d)$,
where $c_n\,\AV(t) \leq \sigma^2_{\mathrm{noise}}(x_t) \leq C_n\,\AV(t)$
for $p_{x_t}$-a.e.\ $x_t$ (Assumption~\ref{asm:noise-comparability}).
Since $\mathcal{G}_t^{\mathrm{Gauss}} \subset \mathcal{G}_t^{\mathrm{comp}}$, the minimax risk
of $\mathcal{G}_t^{\mathrm{comp}}$ is at least that of $\mathcal{G}_t^{\mathrm{Gauss}}$, so
a lower bound for the Gaussian subclass yields a lower bound for the full
compact benchmark class.

For the Gaussian subclass, the KL divergence between location-shifted
experiments satisfies
\begin{equation}\label{eq:kl-bound}
\begin{split}
    D_{\mathrm{KL}}\!\left(P_\omega^{\otimes N} \,\|\, P_{\omega^{(m)}}^{\otimes N}\right)
    &= \frac{N}{2} \int \frac{\|v_\omega(x) - v_{\omega^{(m)}}(x)\|^2}
        {\sigma^2_{\mathrm{noise}}(x)}\, p_{x_t}(x)\,\mathrm{d}x \\
    &\leq \frac{N\,c_0^2\,B(t)^2\,h^{2s}}{c_n\,\AV(t)} \cdot
        \int_{\text{cube}_m}
        \psi^2\!\left(\frac{\phi(x) - z_m}{h}\right) p_{x_t}(x)\,\mathrm{d}x \\
    &\leq \frac{c_2\,N\,B(t)^2\,h^{2s+d}}{\AV(t)}.
\end{split}
\end{equation}
where the first equality is the standard Gaussian KL formula for
location-shift experiments, and the constant $c_2$ absorbs the
noise comparability ratio $1/c_n$.
The key factor $\AV(t)$ in the denominator reflects the noise
level of the velocity regression: higher $\AV(t)$ means noisier
observations, increasing the difficulty of distinguishing hypotheses.

\emph{Remark.}
The upper bound (Subsection~\ref{subsec:upper-bound-kernel}) holds for the
full compact benchmark class $\mathcal{G}_t^{\mathrm{comp}}$ using only the second-moment noise
comparability in Assumption~\ref{asm:noise-comparability}, without
requiring Gaussian noise.  Thus the Gaussian-subclass restriction is
needed only for the lower bound and does not affect the upper bound.

\paragraph{Step 4: Applying Assouad's lemma.}
By Assouad's lemma \citep[Theorem~2.12]{tsybakov2009nonparametric}, for any
estimator $\hat{v}$:
\begin{equation}\label{eq:assouad-application}
    \sup_\omega \,\mathbb{E}_\omega\!\left[\|\hat{v} - v_\omega\|^2_{L^2}\right]
    \;\geq\;
    \frac{M}{2}\,c_1\,B(t)^2\,h^{2s+d}\,\left(1 - \sqrt{\frac{c_2\,N\,B(t)^2\,h^{2s+d}}{2\,\AV(t)}}\right).
\end{equation}

Setting $h \asymp \{\AV(t)/(N B(t)^2)\}^{1/(2s+d)}$ balances the KL divergence against the number of hypotheses. Assouad's lemma then gives
\begin{align}
    R^*_{\mathrm{loc}}(t;\, N)
    &\;\geq\;
    c_3\,(1/h)^d \cdot B(t)^2 h^{2s+d}
    \;=\;
    c_3\,B(t)^2 h^{2s} \notag \\
    &\;=\;
    c_3\,B(t)^{\frac{2d}{2s+d}}
      \left(\frac{\AV(t)}{N}\right)^{\frac{2s}{2s+d}}.
      \label{eq:lower-bound-final}
\end{align}
This is the standard nonparametric lower-bound scaling with H\"older radius $B(t)$ and noise level $\AV(t)$. On compact intervals where both $B(t)$ and $\AV(t)$ are bounded above and below, this implies a common exponent $2s/(2s+d)$ with interval-dependent constants.

\subsection{Upper Bound via Stabilized Local Polynomial Regression}
\label{subsec:upper-bound-kernel}

The upper bound is for the formal compact random-design benchmark
$\mathcal G_t^{\mathrm{comp}}$, rather than for the full ambient
flow-matching observation model. In each chart, let $Z$ denote the intrinsic
coordinate, let $Y=f(Z)+\xi\in\mathbb R^D$, and use a boundary-aware local
polynomial estimator of degree $m=\lceil s\rceil-1$. The benchmark assumes a
compact chart domain, a design density bounded above and below, a uniform
population Gram lower bound for the local polynomial basis, and the vector
$C^{m,\beta}$ condition in Assumption~\ref{asm:smoothness}, with
$\beta=s-m$ (and the $C^{s-1,1}$ convention for integer $s$).

For every admissible bandwidth $0<h\le h_0$, empirical Gram concentration and
polynomial reproduction give, after stabilization on the bad Gram event,
\begin{equation}\label{eq:stabilized-local-bound}
  \mathbb E_{\mathcal D_N,Z}
  \|\widehat f_h(Z)-f(Z)\|_2^2
  \le C_b B(t)^2h^{2s}
      +C_v\frac{\AV(t)}{Nh^d}
      +C_cB(t)^2e^{-cNh^d}.
\end{equation}
The variance calculation uses the trace of the conditional covariance, so no
additional output-dimension factor is hidden in $\AV(t)$. The last term is the
random-design coverage failure; it cannot be dropped in the zero-noise limit.
The expectation is integrated over an independent test point (Tonelli's
 theorem), and does not assert a simultaneous empirical event for all queries.
A finite atlas is combined with fixed-overlap cutoffs and a partition of unity;
its constants are absorbed into $C_b,C_v,C_c$.

Choosing $h$ on the noisy branch and using the coverage bandwidth
$h_{\rm cov}\asymp(\log N/N)^{1/d}$ on the low-noise branch yields
\begin{equation}\label{eq:upper-bound-final}
  R^*_{\mathrm{loc}}(t;N)
  \le C_2\min\left\{B(t)^2,
  B(t)^{\frac{2d}{2s+d}}
  \left(\frac{\AV(t)}{N}\right)^{\frac{2s}{2s+d}}
  +B(t)^2\left(\frac{\log N}{N}\right)^{2s/d}\right\}.
\end{equation}
For fixed $t$ with $\AV(t)>0$, the coverage term is eventually lower order,
recovering the usual upper exponent. Uniform constants and a uniform threshold
require uniform atlas, Gram, design, noise, and smoothness bounds over the
compact time interval.

\subsection{Combining the Bounds}

The lower bound~\eqref{eq:lower-bound-final} and the stabilized upper bound
~\eqref{eq:upper-bound-final} have the same classical exponent only in the
fixed-positive-noise, coverage-sufficient regime. At finite $N$ or as
$\AV(t)\to0$, the coverage term in the upper bound is essential, so a global
two-sided $\AV(t)$-power law is not claimed. On compact time intervals where
$B(t)$ and $\AV(t)$ are bounded above and below and the uniform coverage
threshold is met, the result gives a bounded-factor common exponent; it is not
an exact linear factorization in $\AV(t)$.

\begin{remark}[Role of $\AV(t)$]
The appearance of $\AV(t)$ inside the standard noise-level factor is the key structural insight: $\AV(t)$ plays the role of the noise variance in classical nonparametric regression theory.
Unlike the diffusion model setting \citep{oko2023}, where the analogous
quantity is the score function's noise level at each diffusion time, here $\AV(t)$
has a direct interpretation as the Bayes risk of velocity prediction
(Definition~\ref{def:sigma}).  This yields a bounded-factor common-exponent structure on compact intervals and motivates the
normalization principle in Proposition~\ref{prop:normalized-collapse}.
\end{remark}

\begin{remark}[Constants and the factorization]
\label{rem:factorization-constants}
The raw Assouad bound yields $(\AV(t)/N)^{2s/(2s+d)}$; the multiplicative
form $\AV(t) \cdot N^{-2s/(2s+d)}$ absorbs $\AV(t)^{-d/(2s+d)}$ into
the constant.  On $[\delta, 1{-}\delta]$, this absorbed factor lies in
$[\sigma_u^{-d/(2s+d)},\, \sigma_\ell^{-d/(2s+d)}]$ and is thus bounded.
Consequently, the ``collapse'' of the normalized excess risk
$R^*_{\mathrm{loc}}(t;N)/\AV(t)$ is exact only up to a slowly varying factor
$\AV(t)^{-d/(2s+d)} \in [(\sigma_u/\sigma_\ell)^{-d/(2s+d)},\, 1]$.
\end{remark}

\begin{remark}[Comparison with score estimation]
\citet{oko2023} establish the minimax rate for score estimation in
diffusion models.  Our result differs in three respects:
(i) we work with the velocity parameterization rather than the score;
(ii) we use the linear interpolation path rather than the variance-preserving SDE;
(iii) we obtain a \emph{time-resolved} rate that explicitly tracks $\AV(t)$
across the interpolation path, rather than integrating over $t$.
The time-resolved perspective is essential for our curve collapse prediction.
\end{remark}


\section{Proof of the Gaussian-Mixture Benchmark}
\label{sec:proof-theorem2prime}

We prove Theorem~\ref{thm:gmm}, the Gaussian-mixture result in the separated
regime.
First, we derive the exact posterior and the Bayes-optimal velocity field.
Second, we show that when the effective separation is large, the posterior
concentrates on a single component with exponentially small leakage, so the
Bayes velocity is locally affine up to exponentially small error.
This reduces the label-oracle residual estimation problem to a collection of
local Gaussian location subproblems, yielding a label-oracle residual upper
bound of parametric order and a matching lower bound for the corresponding
separated-regime subproblem.

Throughout this appendix we follow the conditional convention of
\S\ref{sec:sigma-def}: a fixed context $c$ is implicit and suppressed from
notation. The Gaussian-mixture setting below has no context variable, so the
definitions of $\AV(t)$ and the velocity field reduce to context-free forms
without notational change.

\subsection{Setup and Notation}

\begin{setting}\label{setting:gmm}
Let $p_{\mathrm{data}} = \sum_{k=1}^{K} \pi_k \,\mathcal{N}(\mu_k,\, \sigma^2 I_d)$ be a $K$-component
isotropic Gaussian mixture in $\mathbb{R}^d$, with mixing weights $\pi_k > 0$, $\sum_k \pi_k = 1$,
means $\mu_k \in \mathbb{R}^d$, and common variance $\sigma^2 > 0$.
Define the minimum separation $\Delta_{\min} \coloneqq \min_{j \neq k} \|\mu_j - \mu_k\|$
and the diameter $\Delta_{\max} \coloneqq \max_{j \neq k} \|\mu_j - \mu_k\|$.

Consider the linear interpolation path:
\begin{equation}\label{eq:ot-path-app}
    x_t = (1-t)\,a + t\,\varepsilon, \qquad a \sim p_{\mathrm{data}},\quad \varepsilon \sim \mathcal{N}(0, I_d),\quad t \in (0,1).
\end{equation}
The conditional velocity target is $u_t = \varepsilon - a$, and the Bayes-optimal velocity field is
$v^*(x_t, t) = \mathbb{E}[u_t \mid x_t, t]$.

Given $N$ i.i.d.\ samples $\{a_i\}_{i=1}^N \sim p_{\mathrm{data}}$, an estimator $\hat{v}_N$ is trained
to approximate $v^*$. The excess risk at time $t$ is:
\begin{equation}\label{eq:excess-risk}
    R(t;\, N) \coloneqq \mathbb{E}\bigl[\|\hat{v}_N(x_t, t) - v^*(x_t, t)\|^2\bigr] = L_N(t) - \AV(t),
\end{equation}
where $\AV(t) = \mathbb{E}_{x_t}[\mathrm{tr}\,\mathrm{Cov}[u_t \mid x_t, t]]$ is the Bayes risk.
\end{setting}

We also define the total variance at time $t$:
\begin{equation}\label{eq:sigma-t-sq-app}
    \sigma_t^2 \coloneqq (1-t)^2 \sigma^2 + t^2,
\end{equation}
and the posterior variance (within each component):
\begin{equation}\label{eq:posterior-var}
    \hat{\sigma}_t^2 \coloneqq \frac{\sigma^2 t^2}{\sigma_t^2} = \frac{\sigma^2 t^2}{(1-t)^2\sigma^2 + t^2}.
\end{equation}

\subsection{Information Sets and Label-and-Action Oracles}
\label{app:action-oracle}

The Gaussian-mixture theorem is a residual-regression benchmark. It should be
distinguished from stronger observation models available in supervised flow
matching. The relevant information sets are:
\begin{center}
\small
\begin{tabular}{p{0.23\linewidth}p{0.68\linewidth}}
\toprule
Information set & Role in this paper \\
\midrule
Analytic oracle & Uses the true mixture law to compute $\AV(t)$, the Bayes
velocity field, and posterior responsibilities for diagnostics. \\
Label-oracle residual benchmark & Receives true component labels and fits the
local affine velocity response in each separated cell. This is the benchmark
used in Theorem~\ref{thm:gmm}. \\
Label-and-action oracle & Uses the full supervised tuple. Since
$(x_t,u_t,t)$ determines $a=x_t-t u_t$, this oracle can estimate component
means directly from recovered actions. \\
Blind neural estimator & Trains a neural velocity model from the usual flow
matching loss without mixture labels or analytic responsibilities. \\
\bottomrule
\end{tabular}
\end{center}

For the label-and-action oracle, suppose true component labels are available.
Here $\AVw(t)\coloneqq d\sigma^2/\sigma_t^2$ denotes the within-component
contribution to the ambiguity profile. We recover
$a_i=x_{t,i}-t u_{t,i}$ from each supervised training pair. The
component mean estimator satisfies
\[
  \mathbb{E}\|\hat\mu_k-\mu_k\|^2 = \frac{d\sigma^2}{N_k}.
\]
Since the local affine offset is $b_k(t)=-t\mu_k/\sigma_t^2$, the induced
velocity-offset risk is
\[
  \mathbb{E}\|\hat b_k-b_k\|^2
  = \frac{t^2}{\sigma_t^4}\,
    \mathbb{E}\|\hat\mu_k-\mu_k\|^2
  = \frac{t^2 d\sigma^2}{N_k\sigma_t^4}.
\]
Aggregating over separated components gives the label-and-action scale
\[
  R_{\mathrm{act}}(t;N)
  \asymp
  \frac{K}{N}\frac{t^2}{\sigma_t^2}\,\AVw(t),
\]
up to the same separated-regime constants and exponentially small leakage
terms. This can be smaller than the residual-regression benchmark by the
factor $t^2/\sigma_t^2$. Thus, Theorem~\ref{thm:gmm} is not a lower bound over
all estimators of the full supervised tuple; it is the sharp benchmark for the
residual Gaussian-location subproblem defined below.

\begin{figure}[t]
  \centering
  \safeincludegraphics[width=\linewidth]{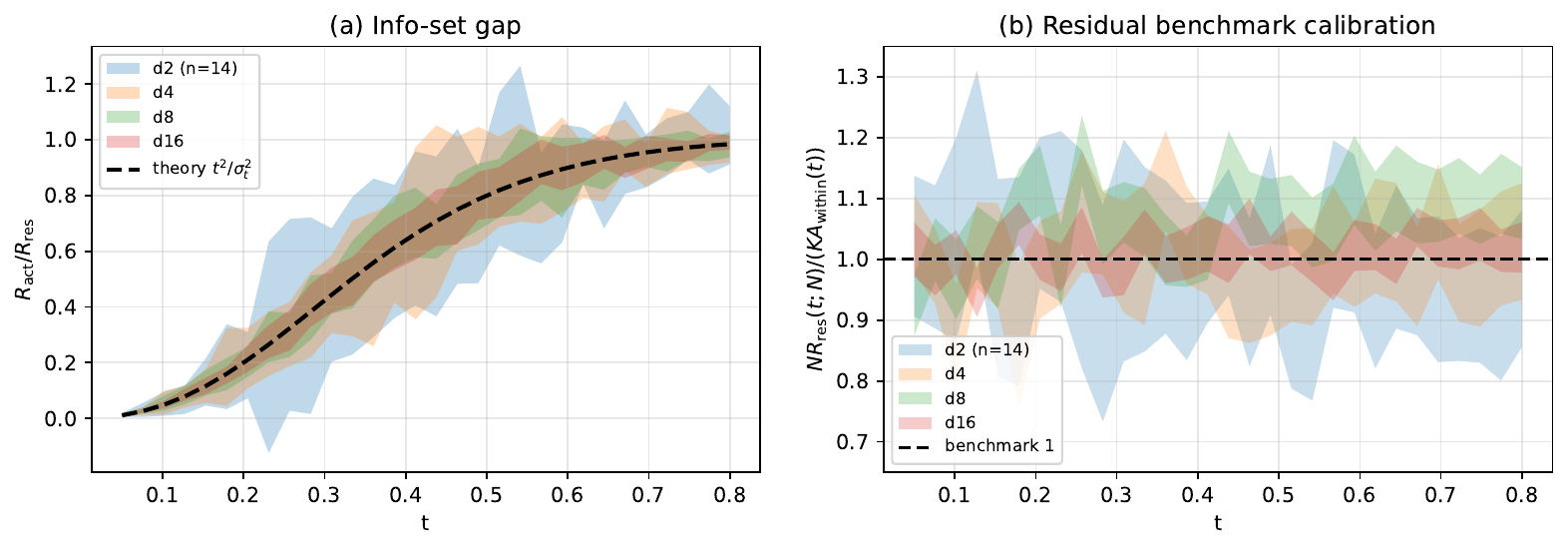}
  \caption{\textbf{Info-set decomposition and residual benchmark calibration.}
    (a) The empirical ratio
    $R_{\mathrm{act}}(t;N) / R_{\mathrm{res}}(t;N)$
    from a closed-form ablation
    ($N{=}2000$, $K{=}4$, $\sigma{=}0.5$, no MLP)
    collapses onto the analytic curve $t^2/\sigma_t^2$ (dashed) across
    $d \in \{2, 4, 8, 16\}$. (b) The residual-regression oracle,
    normalized as $N R_{\mathrm{res}}(t;N)/(K\AVw(t))$, is centered at
    the unit benchmark from Theorem~\ref{thm:gmm}. All scalar summaries
    in the text use the full $20$-seed run for every $d$. Panel (a) shows seed-level spread for the
    ratio readout, while panel (b) shows standard error around the mean
    for residual calibration. Together these panels show that the
    label-and-action oracle is strictly tighter at small $t$, while the
    residual oracle is calibrated to the theorem's sharp
    residual-regression benchmark.}
  \label{fig:info-set-decomp}
\end{figure}

\paragraph{Empirical realization of the info-set gap.}
We numerically verify the analytic decomposition above on closed-form
ablations of the label-oracle residual and label-and-action information sets,
with no MLP training. For separated GMM
configurations with $K{=}4$, $\sigma{=}0.5$, and $d \in \{2, 4, 8, 16\}$,
we generate per-cluster mean estimators from the residual-regression
arm (mean of $u_{t,i} - \beta_1(t) x_{t,i}$ within each component, with
$\beta_1(t) = (t - (1-t)\sigma^2)/\sigma_t^2$ known) and from the
label-and-action arm (per-cluster sample mean of recovered actions
$a_i = x_{t,i} - t u_{t,i}$). Each arm's velocity prediction error
$\hat v(x_t,k) - v^*(x_t,k) = -(t/\sigma_t^2)(\hat\mu_k - \mu_k)$
is constant in $x_t$, so the cell-level excess MSE is exact in closed
form given $\hat\mu_k$. Across $20$ training seeds and an $N$ ladder
$\{50, 100, 200, 500, 1000, 2000\}$, on the readout window
$t \in [0.10, 0.80]$ at $N{=}2000$ the residual estimator satisfies
$N R_{\mathrm{res}}(t;N) / (K \AVw(t))$ within $5.9\%$ of unity at
$d{=}16$ (within $13\%$ at $d{=}4$ and $24\%$ at $d{=}2$, where
finite-dimension averaging variance dominates), and the empirical ratio
$R_{\mathrm{act}}(t;N) / R_{\mathrm{res}}(t;N)$ tracks the analytic
$t^2/\sigma_t^2$ pointwise to within $8.5\%$ at $d{=}16$ across the
window (Figure~\ref{fig:info-set-decomp}). At $t{=}0.05$ the label-and-action
oracle is approximately $90\times$ tighter than the label-oracle residual estimator;
the two converge to within $1.5\%$ by $t{=}0.80$. The factor
$t^2/\sigma_t^2$ gap predicted above is therefore realized pointwise
across $d$, and Theorem~\ref{thm:gmm} is the sharp benchmark for the
residual-regression info set rather than for the full supervised tuple.

\subsection{Main Result}

\begin{theorem}[Restatement of Theorem~\ref{thm:gmm} with explicit hypotheses]
Under Setting~\ref{setting:gmm}, define the effective separation
\[
    \Delta_{\mathrm{eff}}(t) \coloneqq \frac{(1-t)\Delta_{\min}}{\sigma_t},
\]
equivalently the separation signal-to-noise scale is proportional to $\Delta_{\mathrm{eff}}(t)^2$, and the merge
time
\[
    t_{\mathrm{merge}}(\tau) \coloneqq \inf\{t \in [0,1] :
    \Delta_{\mathrm{eff}}(t) < \tau\}
\]
for a chosen threshold $\tau > 0$.  The theorem-level separated statements use
$\tau_{\mathrm{sep}}
\coloneqq C_{\mathrm{sep}}\sqrt{\log K+\log(1/\pi_{\min})}$.
\par The upper bound below is a label-oracle residual-regression upper bound:
it uses the true generating component labels for the residual OLS fit. The
lower bound is information-theoretic for the explicitly defined
separated-regime residual Gaussian-location subproblem.

Then there exist constants $c_1, C_1, C_N > 0$ depending only on
$d, \sigma, \pi_{\min}$, and the aspect ratio
$\Delta_{\max}/\Delta_{\min}$ such that:

\begin{enumerate}[label=(\alph*)]
\item \textbf{Velocity field structure.}
With probability at least
$1 - 2K e^{-c\,\Delta_{\mathrm{eff}}(t)^2}$ over $x_t$,
on this event the posterior responsibility of the dominant component satisfies
$\gamma_{k^*} \geq 1 - K e^{-c\,\Delta_{\mathrm{eff}}(t)^2}$,
and the Bayes-optimal velocity is locally affine up to exponentially small error
(Theorem~\ref{thm:local-affine}).

\item \textbf{Label-oracle residual upper bound.}
For times satisfying $\Delta_{\mathrm{eff}}(t) \ge \tau_{\mathrm{sep}}$ and
$N \ge C_N\pi_{\min}^{-1}(d+\log K)$, there exists a label-oracle residual estimator such that
\begin{equation}\label{eq:parametric-rate-thm2}
    R_{\mathrm{orc}}(t;\, N)
    \;\leq\;
    \frac{C_1\, K}{N}\,\AV(t)
    + \varepsilon_{\mathrm{sep}}(t)
    + \varepsilon_{\mathrm{count}}(t,N),
\end{equation}
where one can take
$\varepsilon_{\mathrm{sep}}(t)=O\!\left(K e^{-c\Delta_{\mathrm{eff}}(t)^2}\Delta_{\max}^2/\sigma_t^2\right)$,
and
$\varepsilon_{\mathrm{count}}(t,N)=
O\!\left(K e^{-c\pi_{\min}N}\Delta_{\max}^2/\sigma_t^2\right)$.

\item \textbf{Separated residual-subproblem lower bound.}
For the corresponding separated-regime residual Gaussian-location subproblem,
\begin{equation}\label{eq:collapse-parametric}
    R_{\mathrm{sub}}^*(t;\, N)
    \;\geq\;
    \frac{c_1\, K}{N}\,\AV(t).
\end{equation}

\end{enumerate}
Consequently, within this label-oracle residual-regression separated
subproblem, the upper bound and lower bound identify $K\,\AV(t)/N$ as the
parametric residual-regression scale, up to exponentially small separation and
count corrections.
The separated-regime normalized collapse (Proposition~\ref{prop:normalized-collapse})
follows as a direct consequence.
\end{theorem}

\begin{remark}[Comparison with the Intrinsic Benchmark]
The intrinsic benchmark minimax rate (Theorem~\ref{thm:minimax}) gives a bounded-factor
time-resolved scaling for general distributions with finite Hölder smoothness
$s$. For the Gaussian mixture, the velocity field is smooth on every bounded
readout neighborhood and therefore satisfies the benchmark assumption for each
fixed finite $s$. The label-oracle residual separated-regime benchmark has the
parametric rate $N^{-1}$ for a different reason: it is a finite-parameter
residual-regression problem. We do not obtain this rate by taking an
$s\to\infty$ limit of Theorem~\ref{thm:minimax}.
\end{remark}

\begin{remark}[Interpretation of $\alpha \approx 1$ in experiments]
The $2$D separated-row sanity audit with $t_{\max}=0.50$ gives
$\alpha = 1.078 \pm 0.157$ and $\mathrm{CV}=0.146$, matching the
parametric scale predicted by Theorem~\ref{thm:gmm}. The main controlled
experiment in Section~\ref{sec:controlled-exp} uses a denser $d=4$ sweep
and shows where this parametric behavior appears within a broader transition:
the plateau occurs at intermediate separation, while extreme overlap and
extreme separation attenuate the observed exponent for different reasons.
\end{remark}

\begin{remark}[Scope and prescriptive limits]
\label{rem:thm2prime-scope}
Theorem~\ref{thm:gmm} is an \emph{label-oracle residual ceiling} in the
Gaussian-mixture separated regime: below
$t_{\mathrm{merge}}(\tau_{\mathrm{sep}})$, the label-oracle residual estimator
achieves the parametric residual-benchmark scale $K\AV(t)/N$; above
$t_{\mathrm{merge}}(\tau_{\mathrm{sep}})$, we only retain
the nonparametric minimax guarantee of Theorem~\ref{thm:minimax}. This makes
$t_{\mathrm{merge}}(\tau_{\mathrm{sep}})$ an \emph{interpretive} phase boundary for the
separated-to-unresolved transition in the separated-GMM label-oracle residual
picture; it is not a
proved sharp theorem about blind estimators on general distributions. It is
also not a prescriptive recipe: the present result does not imply that
$\AV(t)$-informed time reweighting will improve a practical blind
flow matching pipeline. We document this empirically in
\S\ref{sec:envelope}, where a controlled time-sampler stress test gives
same-sign paired improvements across the tested mixture-to-pure arms in
the main controlled cell with a demonstration-seed-disjoint proxy source.
Appendix~\ref{app:m9e-v5} adds a three-cell $d$-axis sensitivity check:
mixture arms and lower pure exponents preserve the improvement direction,
while the most aggressive pure sampler exposes a boundary at the largest
cell. Broader generality beyond this partial-merge $d$-axis family
remains open; whether structured non-time signals such as stage-aware
reweighting yield further gains is also untested.
\end{remark}

\subsection{Proof Map}

The appendix proof follows the three clauses of Theorem~\ref{thm:gmm}.
\begin{enumerate}[label=(\alph*)]
\item \textbf{Mode dominance and local affine structure.}
Lemma~\ref{lem:mode-dominance} and Theorem~\ref{thm:local-affine} show that,
on the separated event, one component dominates and the Bayes velocity is
locally affine.
\item \textbf{Label-oracle residual upper bound.}
The label-oracle residual construction gives the $K\AV(t)/N$ upper scale with
separation and count remainders.
\item \textbf{Residual lower bound.}
The separated residual Gaussian-location subproblem gives the matching
$K\AV(t)/N$ lower scale.
\end{enumerate}

\subsection{Discussion: Beyond Gaussian Mixtures}
\label{sec:general-adaptive}

For general distributions $p_{\mathrm{data}}$ supported on a $d$-dimensional
$C^2$ manifold with velocity field $v^* \in C^s$ on the stated compact
coordinate domains,
Conjecture~\ref{conj:adaptive} predicts the adaptive rate
\begin{equation}
    R(t;\, N) = \AV(t) \cdot \Theta\!\left(N^{-\frac{2s_{\mathrm{eff}}(t)}{2s_{\mathrm{eff}}(t) + d_{\mathrm{eff}}(t)}}\right),
\end{equation}
where $s_{\mathrm{eff}}(t)$ and $d_{\mathrm{eff}}(t)$ are effective parameters
that interpolate between two regimes around a model-dependent transition time.
In the isotropic-GMM proxy this transition is denoted
$t_{\mathrm{merge}}(\tau)$ for a chosen threshold $\tau$:
\begin{itemize}
    \item $t < t_{\mathrm{merge}}(\tau)$: if $p_{\mathrm{data}}$ has modal structure,
    the local residual model is finite-dimensional (described
    phenomenologically by a large $s_{\mathrm{eff}}$ with
    $d_{\mathrm{eff}}=d$), and the rate approaches the parametric scale
    $N^{-1}$;
    \item $t > t_{\mathrm{merge}}(\tau)$: $s_{\mathrm{eff}}(t)$ is determined by the
    effective finite Hölder smoothness of $v^*$ ($s_{\mathrm{eff}} = s$,
    $d_{\mathrm{eff}} = d$), yielding the nonparametric rate
    $\AV(t) \cdot N^{-2s_{\mathrm{eff}}/(2s_{\mathrm{eff}}+d_{\mathrm{eff}})}$.
\end{itemize}
The adaptive mechanism is motivated by \citet{schmidt2020nonparametric}:
deep networks exploit the piecewise smooth structure of $v^*$ without
requiring prior knowledge of $s_{\mathrm{eff}}$. We treat this as a
conjectural extension and use it only as interpretive vocabulary for
controlled and appendix diagnostics.

\paragraph{Controlled SGD support.}
A controlled flow-matching experiment on a separated Gaussian mixture
(Appendix~\ref{app:adaptive-sgd-evidence}) provides empirical support for the
trained-network direction of this picture. An SGD-trained velocity network has a
data-end exponent $\alpha \approx 0.94$ (approaching the parametric scale
$N^{-1}$) and a slower nonparametric exponent toward the noise end, with the
ordering invariant across two time samplers and two training-budget conventions
once a minimum-gradient-step floor removes the small-sample undertraining
confound. A complementary copying estimator (the empirical-velocity
nearest-endpoint rule) instead localizes with a one-nearest-neighbor exponent
$\alpha = 2/D$ in the intrinsic dimension $D$, tracking $2/D$ to within $0.02$
across $D \in \{4,5,6,8\}$. The two estimators therefore separate by their
data-end exponent: the trained network approaches the parametric scale
independent of dimension, whereas the copying reference tracks the
dimension-dependent localization exponent.

\subsection{Effective parameters and adaptive-rate conjecture}
\label{app:adaptive-conjecture}

Real-world action distributions are far from worst-case: they exhibit
multi-modality, low intrinsic dimension, and piecewise smooth structure.
This motivates a conjectural adaptive-rate picture in which sufficiently
expressive estimators may exploit this structure to converge faster than the
worst-case minimax rate of Theorem~\ref{thm:minimax}.

\begin{remark}[Effective smoothness and dimension-phenomenological parameters]
\label{rem:effective-params}
When the excess risk at time~$t$ takes the power-law form
$R(t;N) = \AV(t) \cdot \Theta(N^{-\alpha(t)})$,
we \emph{define} the effective smoothness $s_{\mathrm{eff}}(t)$ and dimension
$d_{\mathrm{eff}}(t)$ via the identity
$\alpha(t) = 2s_{\mathrm{eff}}(t)/(2s_{\mathrm{eff}}(t) + d_{\mathrm{eff}}(t))$:
\begin{equation}\label{eq:adaptive-rate}
    R(t;\, N)
    \;=\;
    \AV(t) \cdot \Theta\!\left(N^{-\frac{2\,s_{\mathrm{eff}}(t)}{2\,s_{\mathrm{eff}}(t) + d_{\mathrm{eff}}(t)}}\right).
\end{equation}
These are \emph{phenomenological} parameters inferred from observed
scaling behavior, not a priori structural quantities. The observed exponent
$\alpha(t)$ alone does not uniquely determine $s_{\mathrm{eff}}$ and
$d_{\mathrm{eff}}$; additional structural assumptions (such as fixing
$d_{\mathrm{eff}} = d$) are needed for identifiability. The
Gaussian-mixture residual benchmark can be encoded by the limiting
phenomenological notation $s_{\mathrm{eff}}\to\infty$, $d_{\mathrm{eff}}=d$
in the parametric regime, yielding $\alpha=1$. This notation is not an
Assumption~\ref{asm:smoothness} claim and is recorded in
Appendix~\ref{app:eff-params-gmm}.
\end{remark}

When $s_{\mathrm{eff}}(t)$ exceeds the nominal smoothness $s$ in
Theorem~\ref{thm:minimax}, the estimator achieves a rate faster than
the minimax bound, a phenomenon known as \emph{adaptation}
\citep{schmidt2020nonparametric}.

\begin{conjecture}[Adaptive Rate for General Structured Data]
\label{conj:adaptive}
Under Assumptions~\ref{asm:manifold}-\ref{asm:smoothness}, suppose additionally
that $p_{\mathrm{data}}$ has structured support (e.g., finite mixture,
piecewise manifold) such that the conditional velocity field $v^*(\cdot, t)$
has effective smoothness $s_{\mathrm{eff}}(t) > s$.
Then a neural network estimator $\hat{v}_N$ with appropriate architecture
achieves:
\begin{equation}\label{eq:adaptive-main}
    R(t;\, N)
    \;=\;
    \AV(t) \cdot \Theta\!\left(N^{-\frac{2\,s_{\mathrm{eff}}(t)}{2\,s_{\mathrm{eff}}(t) + d_{\mathrm{eff}}(t)}}\right),
\end{equation}
where $s_{\mathrm{eff}}(t)$ and $d_{\mathrm{eff}}(t)$ depend on the local structure
of $p_{\mathrm{data}}$ at time step $t$.
\end{conjecture}

\emph{Rigorous for Gaussian mixtures:} Theorem~\ref{thm:gmm} below establishes
a separated-regime residual benchmark of parametric order together with a
matching lower bound for the corresponding residual Gaussian-location
subproblem.
\emph{General case:} a theory-motivated extrapolation from the adaptive
deep nonparametric regression results of \citet{schmidt2020nonparametric},
used as interpretive vocabulary in the empirical diagnostics
(Section~\ref{sec:experiments}).

This conjecture is included as an interpretive extension rather than a
required step for the proved results. It motivates the possibility that
$s_{\mathrm{eff}}(t)$ varies with $t$, creating regimes separated by a
critical time $t_{\mathrm{merge}}$, but Theorems~\ref{thm:minimax}
and~\ref{thm:gmm} stand independently of that conjectural picture. A controlled
SGD experiment (Appendix~\ref{app:adaptive-sgd-evidence}) provides evidence
consistent with this $t$-varying $\alpha(t)$ in the trained-network direction;
it supports the direction of the conjecture without elevating it to a theorem.

\subsection{Effective parameters: Gaussian-mixture instance}
\label{app:eff-params-gmm}

The phenomenological pair
$(s_{\mathrm{eff}}, d_{\mathrm{eff}})$ defined in
Remark~\ref{rem:effective-params} takes a concrete value for the
Gaussian-mixture model under Setting~\ref{setting:gmm}: in the parametric
(separated) regime $\Delta_{\mathrm{eff}}(t) \ge \tau_{\mathrm{sep}}$,
Theorem~\ref{thm:gmm} delivers the residual-benchmark rate
$R(t;N) = \Theta(K\,\AV(t)/N)$ up to exponentially small separation and
component-count corrections, which can be represented phenomenologically by
$s_{\mathrm{eff}}\to\infty$, $d_{\mathrm{eff}} = d$, hence $\alpha = 1$.
This is a finite-dimensional residual result, not an infinite-smoothness
limit of Theorem~\ref{thm:minimax}.
Outside the separated regime the same pair serves only as a descriptive
framework consistent with Conjecture~\ref{conj:adaptive}.

\subsection{Controlled adaptive-rate evidence: trained network and copying estimator}
\label{app:adaptive-sgd-evidence}

We report a controlled flow-matching experiment that separates the
trained-network excess-risk rate from the copying (empirical-velocity) rate on a
common setting. The target is a separated isotropic Gaussian mixture ($K=12$,
$\sigma=0.5$, centers on a radius-$3$ sphere), with the path
$x_t = t\,\varepsilon + (1-t)\,a$ so that data sit at $t=0$. For each time bin we
estimate the held-out velocity-risk exponent $\alpha(t)$ from the slope of the
log paired excess risk against $\log N$ over a demonstration-count ladder, using
common random numbers so that the dominant $\AV(t)$ variance cancels. Two
estimators are compared.

\paragraph{Copying estimator (population-floor reference).}
The empirical-velocity rule copies the velocity of the nearest training
endpoint. Measured against the analytic Bayes-velocity population floor (which
removes the finite-sample bias of an empirical floor), its data-end exponent
tracks the one-nearest-neighbor localization exponent $2/D$ across dimension.

\begin{center}
\begin{tabular}{ccc}
\toprule
$D$ & target $2/D$ & measured $\alpha_{\mathrm{data}}$ \\
\midrule
$4$ & $0.500$ & $0.480$ \\
$5$ & $0.400$ & $0.398$ \\
$6$ & $0.333$ & $0.331$ \\
$8$ & $0.250$ & $0.272$ \\
\bottomrule
\end{tabular}
\end{center}

A regression of the measured exponent on $2/D$ has slope $0.85$ and $R^2=0.997$;
the monotone trend is significant under an isotonic test ($p=0.0002$) and a
label-shuffle null does not reproduce it. The case $D=3$ is omitted as degenerate
($2/D \approx 1$ coincides with the averaging exponent).

\paragraph{Trained network (SGD).}
An SGD-trained velocity network, evaluated against a self-referenced floor that
cancels its $N$-independent approximation error, shows the opposite ordering:
a data-end exponent $\alpha_{\mathrm{data}} \approx 0.94$ (approaching the
parametric scale $N^{-1}$) and a slower nonparametric exponent toward the noise
end ($\alpha_{\mathrm{noise}} \approx 0.5$). This is the direction predicted by
Conjecture~\ref{conj:adaptive}: separated modes near the data end give a locally
affine field and an exponent close to $1$, while merged modes near the noise end
give a slower nonparametric exponent. The ordering is invariant across two time
samplers (uniform and beta) and two budget conventions (equal gradient steps and
equal epochs) once a minimum-gradient-step floor lifts the smallest-sample models
out of the undertraining regime; all four sampler-by-budget cells agree in sign.
The data-end exponent is significant in this direction under the isotonic test
($p=0.0002$) with a clean shuffle null.

\paragraph{Reading.}
The two estimators separate along their data-end rate: the trained network
approaches the parametric exponent independent of dimension, whereas the copying
reference tracks the dimension-dependent localization exponent $2/D$. We report
the trained-network profile as controlled support for the direction of
Conjecture~\ref{conj:adaptive}, not as a proof. The noise-end exponent is
resolved in point estimate but with wide intervals, and the budget invariance
holds in sign with small magnitude under the equal-epoch budget.

\subsection{Comparison with Theorem~\ref{thm:minimax}}
\label{app:thm2prime-vs-thm1}

The intrinsic benchmark minimax rate (Theorem~\ref{thm:minimax}) gives the
usual nonparametric exponent $2s/(2s+d)$, with time dependence controlled
by $\AV(t)$ up to interval-dependent constants. For the Gaussian
mixture, $v^*$ is $C^\infty$, so the nonparametric rate with any finite
$s$ is slower than the parametric rate $N^{-1}$. This is not a
contradiction: Theorem~\ref{thm:minimax} is a worst-case intrinsic
nonparametric benchmark, whereas Theorem~\ref{thm:gmm} exploits the
additional finite-dimensional structure of the separated GMM regime.


\subsection{Upper Bound (Parametric Regime)}
\label{sec:thm2prime-upper}

Throughout this section we work under the following assumptions.

\begin{assumption}[Gaussian Mixture Model]\label{assump:gmm}
The data distribution is a $K$-component isotropic Gaussian mixture in $\mathbb{R}^d$:
\[
  p_{\mathrm{data}}(a) = \sum_{k=1}^{K} \pi_k \,\mathcal{N}(a;\, \mu_k,\, \sigma^2 I_d),
\]
with mixing weights $\pi_k > 0$, $\sum_{k=1}^{K}\pi_k = 1$,
component means $\mu_k \in \mathbb{R}^d$, and common variance $\sigma^2 > 0$.
\end{assumption}

\begin{assumption}[Separation]\label{assump:separation}
The minimum pairwise separation satisfies
\[
  \Delta \;:=\; \min_{j \neq k} \|\mu_j - \mu_k\| \;>\; 0.
\]
\end{assumption}

\begin{assumption}[Linear Interpolation]\label{assump:interpolation}
For $a \sim p_{\mathrm{data}}$ and $\varepsilon \sim \mathcal{N}(0, I_d)$ independent,
\[
  x_t = (1-t)\,a + t\,\varepsilon, \qquad t \in (0,1).
\]
The conditional velocity target is $u_t = \varepsilon - a$, and the Bayes-optimal
velocity field is $v^*(x_t,t) = \mathbb{E}[u_t \mid x_t, t]$.
\end{assumption}

We define the \emph{total variance at time $t$}:
\begin{equation}\label{eq:sigma-t-sq-upper}
  \sigma_t^2 \;:=\; (1-t)^2 \sigma^2 + t^2,
\end{equation}
and the \emph{velocity ambiguity profile}:
\begin{equation}\label{eq:Sigma-t-def}
  \AV(t) \;:=\; \mathbb{E}_{x_t}\!\bigl[\operatorname{tr}\operatorname{Cov}[u_t \mid x_t,t]\bigr].
\end{equation}

\subsubsection{Posterior Computation}
\label{subsec:posterior}

We derive the closed-form posterior $p(a \mid x_t, t)$.

\begin{lemma}[Conditional distribution of $x_t$]\label{lem:xt-conditional}
Under Assumptions~\ref{assump:gmm}-\ref{assump:interpolation},
conditioned on $a$ belonging to component $k$,
\[
  x_t \mid k \;\sim\; \mathcal{N}\!\bigl((1-t)\mu_k,\; \sigma_t^2 I_d\bigr).
\]
\end{lemma}

\begin{proof}
Given $a \mid k \sim \mathcal{N}(\mu_k, \sigma^2 I_d)$ and $\varepsilon \sim \mathcal{N}(0, I_d)$
independent, $x_t = (1-t)a + t\varepsilon$ is a sum of two independent Gaussians:
\[
  (1-t)a \mid k \;\sim\; \mathcal{N}\!\bigl((1-t)\mu_k,\; (1-t)^2\sigma^2 I_d\bigr),
  \qquad
  t\varepsilon \;\sim\; \mathcal{N}(0,\; t^2 I_d).
\]
Since the sum of independent Gaussians is Gaussian with mean and covariance
given by the respective sums,
$x_t \mid k \sim \mathcal{N}\!\bigl((1-t)\mu_k,\, [(1-t)^2\sigma^2 + t^2]\,I_d\bigr)
= \mathcal{N}\!\bigl((1-t)\mu_k,\, \sigma_t^2 I_d\bigr)$.
\end{proof}

\begin{definition}[Posterior responsibilities]\label{def:responsibilities}
The posterior weight of component $k$ given $(x_t, t)$ is
\begin{equation}\label{eq:gamma-k}
  \gamma_k(x_t, t)
  \;:=\;
  \frac{
    \pi_k \,\phi_d\!\bigl(x_t;\, (1-t)\mu_k,\, \sigma_t^2 I_d\bigr)
  }{
    \sum_{j=1}^{K} \pi_j \,\phi_d\!\bigl(x_t;\, (1-t)\mu_j,\, \sigma_t^2 I_d\bigr)
  },
\end{equation}
where $\phi_d(\cdot\,; m, \Sigma)$ denotes the $d$-dimensional Gaussian density
with mean $m$ and covariance $\Sigma$.
\end{definition}

\begin{proposition}[Posterior distribution of $a$]\label{prop:posterior-a}
The posterior $p(a \mid x_t, t)$ is a Gaussian mixture:
\begin{equation}\label{eq:posterior-a}
  p(a \mid x_t, t)
  \;=\;
  \sum_{k=1}^{K} \gamma_k(x_t, t)
  \;\mathcal{N}\!\Bigl(
    a;\;
    \hat{m}_k(x_t, t),\;
    \frac{\sigma^2 t^2}{\sigma_t^2}\, I_d
  \Bigr),
\end{equation}
where the component-wise posterior mean is
\begin{equation}\label{eq:mk-hat}
  \hat{m}_k(x_t, t)
  \;=\;
  \frac{(1-t)\sigma^2\, x_t + t^2\, \mu_k}{\sigma_t^2}.
\end{equation}
\end{proposition}

\begin{proof}
Fix component $k$.  The prior on $a$ is $p(a \mid k) = \mathcal{N}(a; \mu_k, \sigma^2 I_d)$,
and the likelihood is $p(x_t \mid a, t) = \mathcal{N}(x_t; (1-t)a, t^2 I_d)$.
Viewed as a function of $a$, the likelihood is proportional to
$\exp\!\bigl(-\frac{1}{2t^2}\|x_t - (1-t)a\|^2\bigr)$,
which is a Gaussian in $a$ with mean $x_t/(1-t)$ and variance $t^2/(1-t)^2$ per coordinate.

By standard Gaussian conjugacy, the posterior $a \mid k, x_t, t$ is Gaussian with
\begin{align}
  \text{precision:} \quad
  \hat{\sigma}_t^{-2}
  &= \sigma^{-2} + \frac{(1-t)^2}{t^2}
  = \frac{t^2 + (1-t)^2\sigma^2}{t^2 \sigma^2}
  = \frac{\sigma_t^2}{t^2 \sigma^2}, \label{eq:post-precision} \\[4pt]
  \text{variance:} \quad
  \hat{\sigma}_t^2
  &= \frac{t^2 \sigma^2}{\sigma_t^2}, \label{eq:post-var} \\[4pt]
  \text{mean:} \quad
  \hat{m}_k
  &= \hat{\sigma}_t^2
    \Bigl(
      \frac{\mu_k}{\sigma^2} + \frac{(1-t)\,x_t}{t^2}
    \Bigr)
  = \frac{t^2 \sigma^2}{\sigma_t^2}
    \cdot
    \frac{t^2 \mu_k + (1-t)\sigma^2 x_t}{t^2 \sigma^2}
  = \frac{(1-t)\sigma^2\, x_t + t^2 \mu_k}{\sigma_t^2}. \label{eq:post-mean}
\end{align}
The full posterior is the mixture $\sum_k \gamma_k(x_t,t)\,\mathcal{N}(a;\hat{m}_k, \hat{\sigma}_t^2 I_d)$,
where $\gamma_k$ are the posterior responsibilities from Definition~\ref{def:responsibilities}.
\end{proof}

\subsubsection{Velocity Field Closed Form}
\label{subsec:velocity-closed-form}

\begin{proposition}[Optimal velocity field]\label{prop:velocity-field}
Under Assumptions~\ref{assump:gmm}-\ref{assump:interpolation},
the Bayes-optimal velocity field is
\begin{equation}\label{eq:v-star}
  v^*(x_t, t)
  \;=\;
  \frac{1}{t}\bigl(x_t - \bar{a}(x_t, t)\bigr),
\end{equation}
where $\bar{a}(x_t,t) := \mathbb{E}[a \mid x_t, t]$ is given explicitly by
\begin{equation}\label{eq:a-bar}
  \bar{a}(x_t, t)
  \;=\;
  \sum_{k=1}^{K} \gamma_k(x_t, t)\,\hat{m}_k(x_t, t)
  \;=\;
  \frac{(1-t)\sigma^2\, x_t + t^2\, \bar{\mu}(x_t, t)}{\sigma_t^2},
\end{equation}
with $\bar{\mu}(x_t,t) := \sum_{k=1}^{K} \gamma_k(x_t,t)\,\mu_k$
the posterior-weighted mean of the component centers.

Equivalently,
\begin{equation}\label{eq:v-star-explicit}
  v^*(x_t, t)
  \;=\;
  \frac{\bigl[t - (1-t)\sigma^2\bigr]\,x_t - t\,\bar{\mu}(x_t,t)}{\sigma_t^2}.
\end{equation}
\end{proposition}

\begin{proof}
Since $u_t = \varepsilon - a$ and $\varepsilon = (x_t - (1-t)a)/t$, we have
$u_t = (x_t - a)/t$.  Thus $v^*(x_t,t) = \mathbb{E}[u_t \mid x_t,t] = (x_t - \bar{a})/t$.

For the explicit form of $\bar{a}$, by the tower property over component indices:
\[
  \bar{a}(x_t,t)
  = \sum_{k=1}^{K} \gamma_k(x_t,t)\,\hat{m}_k(x_t,t)
  = \sum_{k=1}^{K} \gamma_k \cdot
    \frac{(1-t)\sigma^2\, x_t + t^2\, \mu_k}{\sigma_t^2}
  = \frac{(1-t)\sigma^2\, x_t + t^2\, \bar{\mu}(x_t,t)}{\sigma_t^2},
\]
where we used $\sum_k \gamma_k = 1$.
Substituting into~\eqref{eq:v-star}:
\begin{align*}
  v^*(x_t,t)
  &= \frac{1}{t}\biggl(x_t - \frac{(1-t)\sigma^2\,x_t + t^2\,\bar{\mu}}{\sigma_t^2}\biggr) \\
  &= \frac{1}{t} \cdot
     \frac{\sigma_t^2\,x_t - (1-t)\sigma^2\,x_t - t^2\,\bar{\mu}}{\sigma_t^2} \\
  &= \frac{1}{t} \cdot
     \frac{\bigl[(1-t)^2\sigma^2 + t^2 - (1-t)\sigma^2\bigr]x_t - t^2\,\bar{\mu}}{\sigma_t^2}.
\end{align*}
The coefficient of $x_t$ in the numerator simplifies as
\[
  (1-t)^2\sigma^2 + t^2 - (1-t)\sigma^2
  = (1-t)\sigma^2\bigl[(1-t) - 1\bigr] + t^2
  = -t(1-t)\sigma^2 + t^2
  = t\bigl[t - (1-t)\sigma^2\bigr],
\]
yielding~\eqref{eq:v-star-explicit}.
\end{proof}

\begin{remark}[Structural observation]\label{rem:affine-structure}
The velocity field $v^*(x_t,t)$ is affine in $x_t$ if and only if
$\bar{\mu}(x_t,t) = \sum_k \gamma_k(x_t,t)\,\mu_k$ is constant (i.e.,
independent of $x_t$).  This occurs when the posterior responsibilities
$\gamma_k$ are effectively degenerate-one component dominates throughout.
The ``locally affine'' structure in the parametric regime exploits precisely
this observation (Theorem~\ref{thm:local-affine}).
\end{remark}

\subsubsection{\texorpdfstring{$\AV(t)$}{Sigma(t)} Decomposition}
\label{subsec:sigma-decomposition}

\begin{proposition}[Conditional variance of $u_t$]\label{prop:cond-var}
The conditional covariance of $u_t$ given $(x_t, t)$ satisfies
\begin{equation}\label{eq:cov-ut}
  \operatorname{Cov}[u_t \mid x_t, t]
  \;=\;
  \frac{1}{t^2}\,\operatorname{Cov}[a \mid x_t, t].
\end{equation}
\end{proposition}

\begin{proof}
Since $u_t = (x_t - a)/t$ and $x_t$ is fixed when conditioning,
$\operatorname{Cov}[u_t \mid x_t,t] = \operatorname{Cov}[-a/t \mid x_t,t]
= (1/t^2)\,\operatorname{Cov}[a \mid x_t,t]$.
\end{proof}

\begin{proposition}[Law of total variance decomposition]\label{prop:total-var}
The conditional covariance of $a$ decomposes as
\begin{equation}\label{eq:var-a-decomp}
  \operatorname{Cov}[a \mid x_t, t]
  \;=\;
  \underbrace{\frac{t^2 \sigma^2}{\sigma_t^2}\, I_d}_{\text{within-component}}
  \;+\;
  \underbrace{
    \sum_{k=1}^{K} \gamma_k\,
    (\hat{m}_k - \bar{a})\,(\hat{m}_k - \bar{a})^\top
  }_{\text{between-component}},
\end{equation}
where $\bar{a} = \sum_k \gamma_k \hat{m}_k$.
\end{proposition}

\begin{proof}
By the law of total variance applied to the mixture index $k$:
\[
  \operatorname{Cov}[a \mid x_t, t]
  = \mathbb{E}_k\!\bigl[\operatorname{Cov}[a \mid x_t, t, k]\bigr]
  + \operatorname{Cov}_k\!\bigl[\mathbb{E}[a \mid x_t, t, k]\bigr].
\]
The first term is $\sum_k \gamma_k \cdot \hat{\sigma}_t^2 I_d
= (t^2\sigma^2/\sigma_t^2)\,I_d$,
since each within-component posterior has the same variance
$\hat{\sigma}_t^2 = t^2\sigma^2/\sigma_t^2$
(Proposition~\ref{prop:posterior-a}).
The second term is $\sum_k \gamma_k (\hat{m}_k - \bar{a})(\hat{m}_k - \bar{a})^\top$.
\end{proof}

\begin{theorem}[$\AV(t)$ decomposition]\label{thm:sigma-decomp}
The velocity ambiguity profile admits the decomposition
\begin{equation}\label{eq:sigma-decomp}
  \AV(t)
  \;=\;
  \underbrace{\frac{d\,\sigma^2}{\sigma_t^2}}_{\AVw(t)}
  \;+\;
  \underbrace{
    \frac{t^2}{\sigma_t^4}\,
    \mathbb{E}_{x_t}\!\biggl[\sum_{j < k} \gamma_j \gamma_k\,
    \|\mu_j - \mu_k\|^2\biggr]
  }_{\AVb(t)}.
\end{equation}
\end{theorem}

\begin{proof}
Combining Propositions~\ref{prop:cond-var} and~\ref{prop:total-var}, we have
\[
  \operatorname{tr}\operatorname{Cov}[u_t \mid x_t, t]
  = \frac{d\,\sigma^2}{\sigma_t^2}
  + \frac{1}{t^2}\sum_{k=1}^{K}\gamma_k\,\|\hat{m}_k - \bar{a}\|^2.
\]
The second term can be rewritten using the identity
$\sum_k \gamma_k\,\|\hat{m}_k - \bar{a}\|^2
= \sum_{j<k}\gamma_j\gamma_k\,\|\hat{m}_j - \hat{m}_k\|^2$
(which follows from $\bar{a} = \sum_k \gamma_k \hat{m}_k$ and direct expansion).

From~\eqref{eq:mk-hat}, the difference of posterior means is
\[
  \hat{m}_j - \hat{m}_k
  = \frac{t^2(\mu_j - \mu_k)}{\sigma_t^2},
\]
so $\|\hat{m}_j - \hat{m}_k\|^2 = t^4\|\mu_j - \mu_k\|^2/\sigma_t^4$.  Therefore
\[
  \frac{1}{t^2}\sum_{j<k}\gamma_j\gamma_k\,\|\hat{m}_j - \hat{m}_k\|^2
  = \frac{t^2}{\sigma_t^4}\sum_{j<k}\gamma_j\gamma_k\,\|\mu_j - \mu_k\|^2.
\]
Taking $\mathbb{E}_{x_t}$ of both sides and using~\eqref{eq:Sigma-t-def}
yields~\eqref{eq:sigma-decomp}.
\end{proof}

\begin{remark}[Behavior of the two terms]\phantomsection\label{rem:sigma-behavior}
\begin{enumerate}[label=(\roman*)]
\item $\AVw(t) = d\sigma^2/\sigma_t^2$ is deterministic
  and monotonically varies with $t$.  It represents the irreducible
  per-component posterior uncertainty.
\item $\AVb(t)$ is non-negative and depends on the
  expected product $\mathbb{E}[\gamma_j\gamma_k]$.
  When modes are well separated ($\Delta_{\mathrm{eff}}(t) \gg 1$,
  cf.\ Definition~\ref{def:delta-eff-upper}), the posteriors concentrate on a
  single component and $\gamma_j\gamma_k \approx 0$ for $j \neq k$,
  so $\AVb(t) \approx 0$.
  When modes overlap ($\Delta_{\mathrm{eff}}(t) \lesssim 1$),
  $\AVb(t)$ becomes the dominant contribution.
\end{enumerate}
\end{remark}

\subsubsection{Parametric Regime Upper Bound}
\label{subsec:parametric-upper}

We now establish that when the effective signal-to-noise ratio is large
(modes are well separated in $x_t$-space), the velocity field is locally
affine, and the estimation problem reduces to ordinary least squares (OLS)
within each component, yielding a parametric $N^{-1}$ rate.

\begin{definition}[Effective separation]\label{def:delta-eff-upper}
The effective separation at time $t$ is
\begin{equation}\label{eq:delta-eff-upper}
  \Delta_{\mathrm{eff}}(t)
  \;:=\;
  \frac{(1-t)\,\Delta_{\min}}{\sigma_t}
  \;=\;
  \frac{(1-t)\,\Delta_{\min}}{\sqrt{(1-t)^2\sigma^2 + t^2}},
\end{equation}
where $\Delta_{\min} = \min_{j \neq k}\|\mu_j - \mu_k\|$.
\end{definition}

\begin{definition}[Merge time]\label{def:t-merge}
For a chosen threshold parameter $\tau > 0$, the merge time is
\begin{equation}\label{eq:t-merge}
  t_{\mathrm{merge}}(\tau)
  \;:=\;
  \inf\!\bigl\{t \in (0,1) : \Delta_{\mathrm{eff}}(t) < \tau\bigr\}.
\end{equation}
For $\Delta_{\min} > \tau\sigma$ (the non-trivial regime), the merge time admits the
explicit formula
\begin{equation}\label{eq:t-merge-formula}
  t_{\mathrm{merge}}(\tau)
  = \frac{\sqrt{\Delta_{\min}^2 - \tau^2\sigma^2}}
         {\sqrt{\Delta_{\min}^2 - \tau^2\sigma^2} + \tau}.
\end{equation}
\end{definition}

\begin{lemma}[Mode dominance]\label{lem:mode-dominance}
Let
$\tau_{\mathrm{sep}}
\coloneqq C_{\mathrm{sep}}\sqrt{\log K+\log(1/\pi_{\min})}$.
Fix $t$ with $\Delta_{\mathrm{eff}}(t) \geq \tau_{\mathrm{sep}}$.
Let $z$ be the generating component and let
$k^*(x_t) := \arg\max_k \gamma_k(x_t, t)$.
Then, with probability at least
$1-2K\exp(-c\Delta_{\mathrm{eff}}(t)^2)$ over $x_t \sim p(x_t\mid t)$,
$z=k^*(x_t)$ and, for all $k \neq k^*$,
\begin{equation}\label{eq:mode-dominance}
  \gamma_k(x_t, t)
  \;\leq\;
  \exp\!\bigl(-c\,\Delta_{\mathrm{eff}}(t)^2\bigr),
\end{equation}
and hence $\sum_{k\neq k^*}\gamma_k(x_t,t)
\le K\exp(-c\Delta_{\mathrm{eff}}(t)^2)$.
\end{lemma}

\begin{proof}
Suppose $x_t$ was generated from component $z$, i.e.,
$x_t \mid z \sim \mathcal{N}\!\bigl((1-t)\mu_z,\, \sigma_t^2 I_d\bigr)$.
Write $x_t = (1-t)\mu_z + \sigma_t g$ with $g \sim \mathcal{N}(0, I_d)$.
The log-likelihood ratio between the true component $z$ and a competing
component $k$ is
\begin{align*}
  \log\frac{\gamma_z}{\gamma_k}
  &= \log\frac{\pi_z}{\pi_k}
    + \frac{\|x_t - (1-t)\mu_k\|^2 - \|x_t - (1-t)\mu_z\|^2}{2\sigma_t^2}.
\end{align*}
The quadratic difference evaluates to
\begin{align*}
  \frac{\|x_t - (1-t)\mu_k\|^2 - \|x_t - (1-t)\mu_z\|^2}{2\sigma_t^2}
  &= \frac{(1-t)^2\|\mu_z - \mu_k\|^2}{2\sigma_t^2}
    + \frac{(1-t)}{\sigma_t}\bigl\langle g,\, \mu_z - \mu_k\bigr\rangle \\
  &=: S_1 + S_2.
\end{align*}
The deterministic term satisfies
$S_1 \geq \frac{1}{2}\Delta_{\mathrm{eff}}(t)^2$
since $\|\mu_z - \mu_k\| \geq \Delta_{\min}$.
The stochastic term $S_2$ is Gaussian with mean $0$
and variance
$\sigma_{S_2}^2 = (1-t)^2\|\mu_z-\mu_k\|^2/\sigma_t^2 \leq C'\,\Delta_{\mathrm{eff}}(t)^2$
for a constant $C'$ depending on $\Delta_{\max}/\Delta_{\min}$.

By the Gaussian tail bound $\mathbb{P}(|S_2| > s) \leq 2\exp(-s^2/(2\sigma_{S_2}^2))$,
choosing $s = \frac{1}{4}\Delta_{\mathrm{eff}}(t)^2$,
we obtain $|S_2| \leq \frac{1}{4}\Delta_{\mathrm{eff}}(t)^2$ with probability
at least $1 - 2\exp\!\bigl(-c\Delta_{\mathrm{eff}}(t)^2\bigr)$.

On this event,
$\log(\gamma_z/\gamma_k) \geq S_1 - |S_2| - |\log(\pi_z/\pi_k)|
\geq c\Delta_{\mathrm{eff}}(t)^2$,
where the final inequality uses
$\Delta_{\mathrm{eff}}(t)\ge\tau_{\mathrm{sep}}$ with
$C_{\mathrm{sep}}$ large enough to absorb the mixing-weight ratio.
Thus $\gamma_k \leq \exp(-c\Delta_{\mathrm{eff}}(t)^2)\gamma_z$.
A union bound over the $K-1$ competing components gives the stated event; on
that event $z$ is the posterior mode $k^*(x_t)$ and
\eqref{eq:mode-dominance} follows after adjusting constants.
\end{proof}

\begin{theorem}[Local affine structure]\label{thm:local-affine}
Fix $t$ with $\Delta_{\mathrm{eff}}(t) \geq \tau_{\mathrm{sep}}$.
For $x_t$ in the high-probability region of component $k^*$,
$v^*(x_t, t)$ is $\epsilon$-close to an affine function:
\begin{equation}\label{eq:local-affine}
  \bigl\|v^*(x_t, t) - \bigl(A_{k^*}(t)\,x_t + b_{k^*}(t)\bigr)\bigr\|
  \;\leq\;
  \epsilon_{\mathrm{affine}}(t),
\end{equation}
where
\begin{equation}\label{eq:Ak-bk}
  A_k(t) = \frac{t - (1-t)\sigma^2}{\sigma_t^2}\, I_d,
  \qquad
  b_k(t) = -\frac{t\,\mu_k}{\sigma_t^2},
\end{equation}
and the approximation error is exponentially small:
\begin{equation}\label{eq:eps-affine}
  \epsilon_{\mathrm{affine}}(t)
  \;=\;
  O\!\Biggl(K\exp\!\Bigl(-\tfrac{1}{8}\Delta_{\mathrm{eff}}(t)^2\Bigr)
    \cdot \frac{\Delta_{\max}}{\sigma_t}\Biggr).
\end{equation}
This bound holds with probability at least
$1 - 2K\exp(-c\Delta_{\mathrm{eff}}(t)^2)$ over $x_t$.
\end{theorem}

\begin{proof}
From~\eqref{eq:v-star-explicit}, the velocity field is
\[
  v^*(x_t,t)
  = \frac{[t-(1-t)\sigma^2]\,x_t - t\,\bar{\mu}(x_t,t)}{\sigma_t^2}.
\]
When the posterior is concentrated on component $k^*$,
we write $\bar{\mu}(x_t,t) = \mu_{k^*} + \sum_{k \neq k^*}\gamma_k(\mu_k - \mu_{k^*})$,
since $\gamma_{k^*} = 1 - \sum_{k \neq k^*}\gamma_k$.  Thus
\[
  v^*(x_t,t)
  = \underbrace{
      \frac{[t-(1-t)\sigma^2]\,x_t - t\,\mu_{k^*}}{\sigma_t^2}
    }_{A_{k^*}(t)\,x_t + b_{k^*}(t)}
  - \frac{t}{\sigma_t^2}\sum_{k \neq k^*}\gamma_k\,(\mu_k - \mu_{k^*}).
\]
The remainder term has norm bounded by
\[
  \frac{t}{\sigma_t^2}\sum_{k \neq k^*}\gamma_k\,\|\mu_k - \mu_{k^*}\|
  \;\leq\;
  \frac{t}{\sigma_t^2}\cdot K\exp\!\Bigl(-\tfrac{1}{8}\Delta_{\mathrm{eff}}(t)^2\Bigr)
  \cdot \max_{k\neq k^*}\|\mu_k - \mu_{k^*}\|,
\]
by Lemma~\ref{lem:mode-dominance}.  Since
$\max_{k\neq k^*}\|\mu_k - \mu_{k^*}\| \leq \Delta_{\max}$, where
$\Delta_{\max} = \max_{j \neq k}\|\mu_j - \mu_k\|$ is the mixture diameter,
and $t/\sigma_t^2 \leq 1/\sigma_t$,
we obtain~\eqref{eq:eps-affine} with constants depending on
$\Delta_{\max}/\Delta_{\min}$.
\end{proof}

We now state and prove the main upper bound.

\begin{theorem}[Label-oracle residual parametric upper bound]\label{thm:parametric-upper}
There exist constants $C_N, C_{\mathrm{orc}}>0$ depending only on fixed mixture
parameters such that the following holds.
Fix $t$ with $\Delta_{\mathrm{eff}}(t) \geq \tau_{\mathrm{sep}}$ and
$N \ge C_N\pi_{\min}^{-1}(d+\log K)$.
There exists a label-oracle residual estimator $\hat{v}_N$
(OLS within each true component label)
such that the excess risk satisfies
\begin{equation}\label{eq:parametric-rate-upper}
  R(t; N)
  \;:=\;
  \mathbb{E}\bigl\|\hat{v}_N(x_t, t) - v^*(x_t, t)\bigr\|^2
  \;\leq\;
  C_{\mathrm{orc}}\frac{K}{N}\;\AV(t)
  + \varepsilon_{\mathrm{sep}}(t)
  + \varepsilon_{\mathrm{count}}(t,N),
\end{equation}
where $\pi_{\min} = \min_k \pi_k$ and
\[
  \begin{aligned}
  \varepsilon_{\mathrm{sep}}(t)
  &= O\!\Bigl(
       K\exp\!\bigl(-c\,\Delta_{\mathrm{eff}}(t)^2\bigr)
       \Delta_{\max}^2/\sigma_t^2
     \Bigr),\\
  \varepsilon_{\mathrm{count}}(t,N)
  &= O\!\Bigl(
       K\exp(-c\pi_{\min}N)
       \Delta_{\max}^2/\sigma_t^2
     \Bigr)
  \end{aligned}
\]
are the posterior-boundary leakage and component-count failure terms,
respectively.

In particular, for fixed $K, d, \sigma, \{\pi_k\}$:
\begin{equation}\label{eq:rate-compact}
  \boxed{R(t; N) \;=\; O\!\bigl(\AV(t) \cdot N^{-1}\bigr)
  \qquad \text{whenever } \Delta_{\mathrm{eff}}(t)\ge\tau_{\mathrm{sep}}.}
\end{equation}
\end{theorem}

\begin{proof}
The proof proceeds in four steps.  Since this is a label-oracle residual upper bound,
the estimator uses the true component labels $k_i^*$ (the generating
component of each training sample), not the MAP estimates.

\medskip\noindent
\textbf{Step 1: Label-oracle assignment.}
Given $N$ i.i.d.\ training samples $\{(a_i, \varepsilon_i)\}_{i=1}^N$
from the generative model, we observe the pairs
$(x_{t,i}, u_{t,i})$ where $x_{t,i} = (1-t)a_i + t\varepsilon_i$
and $u_{t,i} = \varepsilon_i - a_i$.
The label-oracle estimator assigns each sample to its true generating component $k_i^*$.
Let $N_k = |\{i : k_i^* = k\}|$ denote the count; then
$\mathbb{E}[N_k] = \pi_k N$, and by a Chernoff bound,
$N_k \geq \frac{1}{2}\pi_k N$ with probability
$\geq 1 - \exp(-c\,\pi_k N)$ for each $k$.

\medskip\noindent
\textbf{Step 2: OLS within each component.}
Within the samples from component $k$, the velocity field
is (approximately) affine: $v^*(x_t, t) = A_k(t)\,x_t + b_k(t) + r_k(x_t,t)$,
where $\|r_k\| \leq \epsilon_{\mathrm{affine}}(t)$ by Theorem~\ref{thm:local-affine}.

Each sample satisfies the linear model
\begin{equation}\label{eq:ols-model}
  u_{t,i} = A_k x_{t,i} + b_k + \xi_i,
  \qquad
  \xi_i := u_{t,i} - v^*(x_{t,i}, t),
\end{equation}
where $\xi_i$ has zero mean conditioned on $x_{t,i}$ and
per-coordinate variance
\begin{equation}\label{eq:noise-var}
  \sigma_{\xi}^2
  = \frac{\hat{\sigma}_t^2}{t^2}
  = \frac{\sigma^2}{\sigma_t^2}
\end{equation}
(from the within-component posterior variance, Proposition~\ref{prop:posterior-a}).

Since $A_k = \alpha(t) I_d$ is a scalar multiple of the identity
(see~\eqref{eq:Ak-bk}), the only unknown is
$b_k = -(t/\sigma_t^2)\mu_k \in \mathbb{R}^d$,
a $d$-dimensional parameter.
The OLS estimator for $b_k$ from $N_k$ samples is
\[
  \hat{b}_k
  = \frac{1}{N_k}\sum_{i:k_i^*=k}
    \bigl(u_{t,i} - A_k x_{t,i}\bigr).
\]
Because the estimator uses true labels, there is no misclassification error
in the training set; each cell contains only samples from the correct component.

\medskip\noindent
\textbf{Step 3: Per-component risk.}
The estimation error of $\hat{b}_k$ satisfies
\[
  \mathbb{E}\bigl[\|\hat{b}_k - b_k\|^2 \;\big|\; N_k\bigr]
  = \frac{d\,\sigma_\xi^2}{N_k}
  = \frac{d\,\sigma^2}{N_k\,\sigma_t^2}.
\]
The per-component excess risk (including the test-time affine approximation error) is
\[
  R_k(t; N)
  = \frac{d\,\sigma^2}{N_k\,\sigma_t^2}
    + O(\epsilon_{\mathrm{affine}}(t)^2).
\]

\medskip\noindent
\textbf{Step 4: Aggregation over components.}
Let $\mathcal{E}_{\mathrm{count}}$ denote the event that
$N_k \geq \frac{1}{2}\pi_k N$ for all $k$.
By a Chernoff bound and union over $K$ components,
$\Pr(\mathcal{E}_{\mathrm{count}}) \geq 1 - K e^{-c\,\pi_{\min}N}$.

On $\mathcal{E}_{\mathrm{count}}$, each test point $x_t$ falls in the basin
of some dominant component $k^*$ with probability $\pi_{k^*}$
(up to exponentially small separation corrections from
Theorem~\ref{thm:local-affine}).
The conditional risk is:
\begin{align*}
  \mathbb{E}\bigl[R \mid \mathcal{E}_{\mathrm{count}}\bigr]
  &= \sum_{k=1}^{K} \pi_k\, R_k(t; N)
    + O\!\bigl(K e^{-c\Delta_{\mathrm{eff}}^2}\,\Delta_{\max}^2/\sigma_t^2\bigr) \\
  &\leq \frac{d\,\sigma^2}{\sigma_t^2}
    \cdot \sum_{k=1}^{K}\frac{\pi_k}{\frac{1}{2}\pi_k N}
    + \varepsilon_{\mathrm{sep}} \\
  &= \frac{2Kd\,\sigma^2}{N\,\sigma_t^2}
    + \varepsilon_{\mathrm{sep}}
  \;=\; \frac{2K}{N}\,\AVw(t)
    + \varepsilon_{\mathrm{sep}}.
\end{align*}

On the complement $\mathcal{E}_{\mathrm{count}}^c$ (probability
$\leq K e^{-c\,\pi_{\min}N}$), the crude risk bound
$O(\Delta_{\max}^2/\sigma_t^2)$ contributes
$K e^{-c\,\pi_{\min}N} \cdot O(\Delta_{\max}^2/\sigma_t^2)$, which is
the component-count term $\varepsilon_{\mathrm{count}}(t,N)$.

In the parametric regime, $\AV(t) = \AVw(t)
+ \AVb(t)$ with $\AVb(t)$
exponentially small (Remark~\ref{rem:sigma-behavior}),
so $\AVw(t) = \AV(t)\,(1 + o(1))$.
This yields~\eqref{eq:parametric-rate-upper}, where
$\varepsilon_{\mathrm{sep}}(t)$ collects the test-time affine approximation error
$O(K e^{-c\Delta_{\mathrm{eff}}^2}\,\Delta_{\max}^2/\sigma_t^2)$
(independent of $N$), and $\varepsilon_{\mathrm{count}}(t,N)$ records the
exponentially-small-in-$N$ count concentration term.
\end{proof}

\begin{remark}[Oracle vs.\ adaptive estimator]
\label{rem:oracle}
Theorem~\ref{thm:parametric-upper} is a \emph{label-oracle residual} upper bound:
Step~1 uses the true generating component label of each training sample.
A fully adaptive two-stage residual estimator may be able to achieve the same
order; one possible route is as follows.

\emph{Stage~1 (parameter recovery).}
When $\Delta_{\min} \geq C\sigma\tau_{\mathrm{sep}}$ and
$N \geq C' K\pi_{\min}^{-1}(d+\log K)$, spectral methods
\citep{vempala2004spectral} or the EM algorithm recover the component means
to accuracy $\|\hat\mu_k - \mu_k\| = O(\sigma\sqrt{d/N_k})$ per component.
The resulting misclassification probability per training point is bounded by
$K\exp(-c\,\Delta_{\mathrm{eff}}(t)^2) + K\exp(-c'\,N\pi_{\min}\Delta^2/\sigma^2)$,
which is exponentially small under the separation condition.

\emph{Stage~2 (OLS with estimated assignments).}
Running OLS within each estimated cell, the additional excess risk from
misassigned training points is
$O(K\exp(-c''\Delta_{\mathrm{eff}}(t)^2)\cdot\Delta_{\max}^2/\sigma_t^2)$,
which is absorbed into $\varepsilon_{\mathrm{sep}}(t)$.
The parameter-estimation error in $\hat\mu_k$ propagates through
$\hat{b}_k = -(t/\sigma_t^2)\hat\mu_k$, contributing
$O(t^2 d\sigma^2/(\sigma_t^4 N_k))$ per component. This is no larger than the
residual-regression benchmark scale and can be smaller by the factor
$t^2/\sigma_t^2$ discussed in Appendix~\ref{app:action-oracle}.

This heuristic suggests that a fully adaptive two-stage residual estimator
could match the label-oracle residual order $R(t;N) = O(K\,\AV(t)/N)$ for
$N \geq C_0 K\pi_{\min}^{-1}(d+\log K)$, although
we do not make this extension fully rigorous here.
An empirical anchor for this two-stage construction, sweeping
$\Delta_{\min}/\sigma$ to span merge through separated regimes on a
controlled $d{=}2$, $K{=}4$, $\sigma{=}0.5$ family, is given in
Appendix~\ref{app:two-stage-sweep}.
We present the label-oracle residual form in the main theorem for clarity.
\end{remark}

\begin{remark}[Separated-subproblem lower bound]\label{rem:lower-bound}
The lower bound in Theorem~\ref{thm:gmm}(c) applies to the
\emph{separated-regime residual Gaussian-location subproblem}, not to the full
flow matching estimation problem.
Specifically, for the canonical subproblem in which each cell receives
$N_k$ Gaussian observations with per-coordinate variance $\sigma_\xi^2$,
the per-component estimation error for the $d$-dimensional intercept $b_k$
is at least $d\,\sigma_\xi^2/N_k = d\sigma^2/(N_k \sigma_t^2)$.
Aggregating over $K$ cells gives
$R_{\mathrm{sub}}^*(t;N) \geq (K/N)\,\AVw(t) = \Omega(K\,\AV(t)/N)$
up to the same separated-regime correction.
Together with the label-oracle residual upper bound
(Theorem~\ref{thm:parametric-upper}),
this identifies $K\,\AV(t)/N$ as the parametric scale for the label-oracle
residual-regression separated subproblem.
\end{remark}

\begin{remark}[Role of $d$ in the prefactor]\label{rem:d-prefactor}
Since $A_k(t) = \alpha(t)\,I_d$ is fully determined by the known interpolation
schedule (independent of the data), the only parameters to estimate are the
$d$-dimensional offsets $b_k = -(t/\sigma_t^2)\mu_k$, one per component.
Consequently the prefactor in~\eqref{eq:parametric-rate-upper} is $2K/\pi_{\min}$,
which depends on $K$ but \emph{not} on $d$ beyond the $d$ already present
inside $\AV(t) = d\sigma^2/\sigma_t^2 + \AVb(t)$.
For the separated-regime label-oracle residual statement, this constant contributes to the
leading $K\,\AV(t)/N$ term, with any remaining deviation absorbed into the
separation correction.
The $N^{-1}$ rate may be represented by the limiting phenomenological
notation $s_{\mathrm{eff}}\to\infty$ in the general formula, but here it
comes directly from finite-dimensional affine regression.
\end{remark}


\subsection{Lower Bound and Transition}\label{sec:lower-bound-transition}

\subsubsection{Merge Time: Rigorous Definition}\label{sec:tmerge-rigorous}

We formalize the transition between the parametric and nonparametric regimes
via the \emph{signal-to-noise ratio} (SNR) of the component structure in $x_t$-space.

\begin{definition}[Effective separation]\label{def:delta-eff-lower}
For a $K$-component isotropic Gaussian mixture $p_{\mathrm{data}} = \sum_{k=1}^{K}\pi_k\,\mathcal{N}(\mu_k,\sigma^2 I_d)$
with minimum separation $\Delta_{\min} \triangleq \min_{j\neq k}\|\mu_j - \mu_k\|$
and total variance $\sigma_t^2 \triangleq (1-t)^2\sigma^2 + t^2$,
define the \textbf{effective separation} at time~$t$:
\begin{equation}\label{eq:delta-eff-lower}
  \Delta_{\mathrm{eff}}(t) \;\triangleq\; \frac{(1-t)\,\Delta_{\min}}{\sigma_t}
  \;=\; \frac{(1-t)\,\Delta_{\min}}{\sqrt{(1-t)^2\sigma^2 + t^2}}\,.
\end{equation}
\end{definition}

The quantity $\Delta_{\mathrm{eff}}(t)$ is the Mahalanobis separation between the nearest pair
of component means $\{(1-t)\mu_k\}$ in the marginal $p(x_t\mid t)$, which has per-component
covariance $\sigma_t^2 I_d$.

\begin{definition}[Merge time]\label{def:tmerge-lower}
For a chosen threshold $\tau > 0$, the \textbf{merge time} is
\begin{equation}\label{eq:tmerge-def-lower}
  t_{\mathrm{merge}}(\tau) \;\triangleq\; \inf\!\Big\{t \in (0,1) :
    \Delta_{\mathrm{eff}}(t) < \tau\Big\}\,.
\end{equation}
\end{definition}

A posterior-domination scale is
$O(\sqrt{\log K+\log(1/\pi_{\min})})$, and the theorem-level
separated regime uses
$\tau_{\mathrm{sep}}
=C_{\mathrm{sep}}\sqrt{\log K+\log(1/\pi_{\min})}$
(Lemma~\ref{lem:mode-dominance}).  For
$t < t_{\mathrm{merge}}(\tau)$ the posterior is effectively degenerate at
threshold $\tau$; for $t > t_{\mathrm{merge}}(\tau)$, the threshold no longer
certifies single-mode dominance and multiple modes may contribute
nontrivially.

\begin{lemma}[Explicit formula for $t_{\mathrm{merge}}$]\label{lem:tmerge-formula}
Suppose $\Delta_{\min} > \tau\sigma$ for the chosen threshold $\tau$.
Then
\begin{equation}\label{eq:tmerge-explicit}
  t_{\mathrm{merge}}(\tau)
  \;=\; \frac{\sqrt{\Delta_{\min}^2 - \tau^2\sigma^2}}
             {\sqrt{\Delta_{\min}^2 - \tau^2\sigma^2}\;+\;\tau}
  \;=\; \frac{1}{1 + \tau\big/\!\sqrt{\Delta_{\min}^2 - \tau^2\sigma^2}}\,.
\end{equation}
\end{lemma}

\begin{proof}
Setting $\Delta_{\mathrm{eff}}(t) = \tau$ in~\eqref{eq:delta-eff-lower} and squaring:
\[
  (1-t)^2\Delta_{\min}^2 \;=\; \tau^2\bigl[(1-t)^2\sigma^2 + t^2\bigr].
\]
Rearranging,
\[
  (1-t)^2\bigl(\Delta_{\min}^2 - \tau^2\sigma^2\bigr) \;=\; \tau^2\,t^2.
\]
Since $\Delta_{\min} > \tau\sigma$, the left factor is positive, so
\[
  \frac{1-t}{t} \;=\; \frac{\tau}{\sqrt{\Delta_{\min}^2 - \tau^2\sigma^2}}\,,
\]
and solving for $t$ yields~\eqref{eq:tmerge-explicit}.
\end{proof}

\begin{corollary}[Limiting regimes]\label{cor:tmerge-limits}
\leavevmode
\begin{enumerate}[label=(\alph*)]
  \item \textbf{Large separation} ($\Delta_{\min} \gg \tau\sigma$):
    $t_{\mathrm{merge}}(\tau) \to 1 - \tau/\Delta_{\min}$.
    Modes remain separated until close to $t=1$.
  \item \textbf{Moderate separation} ($\Delta_{\min}$ on the order of
    $\tau\sigma$ with a nontrivial margin):
    $t_{\mathrm{merge}}(\tau)$ lies in the interior of $(0,1)$.
  \item \textbf{Marginal separation} ($\Delta_{\min} \lesssim \tau\sigma$):
    $t_{\mathrm{merge}}(\tau) \to 0$.
    The parametric regime vanishes.
\end{enumerate}
\end{corollary}

\begin{proof}
Direct substitution into~\eqref{eq:tmerge-explicit}.
For (a), $\sqrt{\Delta_{\min}^2 - \tau^2\sigma^2}\approx \Delta_{\min}$, so
$t_{\mathrm{merge}}(\tau) \approx \Delta_{\min}/(\Delta_{\min}+\tau) = 1 - \tau/\Delta_{\min} + O(\tau^2/\Delta_{\min}^2)$.
Parts (b) and (c) follow similarly.
\end{proof}

\subsubsection{Separated Residual-Subproblem Lower Bound in the Parametric Regime}\label{sec:CR-lower}

We now establish a lower bound of parametric order for a canonical
separated-regime residual Gaussian-location subproblem. The purpose of this section is
to identify the correct $N^{-1}$ scale in the regime where posterior mass is
concentrated on a single component. We do not claim that the argument below, by
itself, yields a sharp lower bound for the full flow matching estimation
problem outside that separated residual subproblem reduction.

\begin{lemma}[Within-component Gaussian location experiment]\label{lem:fisher}
Fix $t$ in the separated regime and condition on the event
$E_k = \{k^*(x_t)=k\}$
that sample $x_t$ is correctly assigned to component~$k$.
On $E_k$, the velocity field is affine:
\begin{equation}\label{eq:vstar-affine}
  v^*(x_t,t) \;=\; A(t)\,x_t + b_k(t),
  \qquad
  A(t) = \frac{t-(1-t)\sigma^2}{\sigma_t^2}\,I_d,
  \qquad
  b_k(t) = -\frac{t\,\mu_k}{\sigma_t^2}\,,
\end{equation}
up to an approximation error
$\epsilon_{\mathrm{affine}}(t)
= O(K\exp(-c\Delta_{\mathrm{eff}}(t)^2)\Delta_{\max}/\sigma_t)$
that is exponentially small in $\Delta_{\mathrm{eff}}(t)^2$ (Theorem~\ref{thm:local-affine}).

The only unknown parameter in each cell is $\mu_k$ (entering through $b_k$).
The observations within cell~$k$ are
\[
  y_i \;=\; u_{t,i} \;=\; v^*(x_{t,i},t) + \xi_i,
  \qquad
  \xi_i \sim \mathcal{N}\!\bigl(0,\;\sigma_{\mathrm{noise}}^2\,I_d\bigr),
  \qquad
  \sigma_{\mathrm{noise}}^2 \;=\; \frac{\sigma^2}{\sigma_t^2}\,,
\]
where $\sigma_{\mathrm{noise}}^2 = \hat{\sigma}_t^2/t^2 = \sigma^2 t^2/(\sigma_t^2 t^2) = \sigma^2/\sigma_t^2$
is the per-coordinate residual variance from the within-component posterior (Proposition~\ref{prop:cond-var}).

Conditioned on $E_k$ and after subtracting the known affine term $A(t)x_t$, this
reduces to a $d$-dimensional Gaussian location experiment for the intercept
$b_k(t)$ (equivalently for $\mu_k$). The remainder of the section lower-bounds
that canonical experiment.

The Fisher information for $\mu_k$ from a single observation is
\begin{equation}\label{eq:fisher-single}
  \mathcal{I}_1(\mu_k)
  \;=\; \frac{1}{\sigma_{\mathrm{noise}}^2}
        \left\|\frac{\partial v^*}{\partial \mu_k}\right\|^2_{\mathrm{per\text{-}coord}}
        \cdot I_d
  \;=\; \frac{\sigma_t^2}{\sigma^2}\cdot\frac{t^2}{\sigma_t^4}\,I_d
  \;=\; \frac{t^2}{\sigma^2\,\sigma_t^2}\,I_d\,,
\end{equation}
since $\partial b_k/\partial\mu_k = -(t/\sigma_t^2)\,I_d$ and the sensitivity per coordinate is $t/\sigma_t^2$.
\end{lemma}

\begin{proof}
From~\eqref{eq:vstar-affine}, $v^*$ depends on $\mu_k$ solely through $b_k(t) = -t\mu_k/\sigma_t^2$.
The likelihood of observation $y_i$ given $\mu_k$ (within cell~$k$) is
\[
  p(y_i \mid \mu_k) \;\propto\;
  \exp\!\left(-\frac{\|y_i - A(t)x_{t,i} - b_k(t)\|^2}{2\sigma_{\mathrm{noise}}^2}\right).
\]
This is a standard location model with sensitivity matrix $\partial b_k/\partial\mu_k = -(t/\sigma_t^2)I_d$.
The Fisher information per observation per coordinate is
\[
  \bigl[\mathcal{I}_1(\mu_k)\bigr]_{jj}
  = \frac{(t/\sigma_t^2)^2}{\sigma^2/\sigma_t^2}
  = \frac{t^2}{\sigma^2\sigma_t^2}\,.  \qedhere
\]
\end{proof}

\begin{lemma}[Residual Gaussian subproblem lower bound on velocity MSE]\label{lem:CR-bound}
Let $N_k$ denote the number of observations in cell~$k$.
In the residual Gaussian-location subproblem induced by cell~$k$, any estimator of the
intercept parameter has mean-squared error at least
\begin{equation}\label{eq:CR-velocity}
  \inf_{\hat{v}_k}
  \sup_{\mu_k}
  \mathbb{E}\bigl[\|\hat{v}_k(x_t) - v^*(x_t,t)\|^2 \;\big|\; E_k\bigr]
  \;\geq\;
  \frac{d\,\sigma^2}{N_k\,\sigma_t^2}
  \;=\;
  \frac{\AVw(t)}{N_k}\,,
\end{equation}
where $\AVw(t) = d\sigma^2/\sigma_t^2$ is the within-component
contribution to the velocity ambiguity~$\AV(t)$.
This identifies the minimax scale of the separated residual Gaussian-location
subproblem. The proof uses the fact that, in the residual Gaussian-location model, the
sample mean attains the Cram\'er-Rao bound and is minimax optimal.
\end{lemma}

\begin{proof}
With $N_k$ i.i.d.\ observations, the total Fisher information is $\mathcal{I}_{N_k}(\mu_k)
= N_k\,\mathcal{I}_1(\mu_k) = (N_k t^2)/(\sigma^2\sigma_t^2)\,I_d$.
The Cram\'er-Rao bound for estimating $\mu_k$ in the Gaussian location model gives
\begin{equation}\label{eq:CR-mu}
  \mathbb{E}\bigl[\|\hat{\mu}_k - \mu_k\|^2\bigr]
  \;\geq\; \mathrm{tr}\bigl(\mathcal{I}_{N_k}^{-1}\bigr)
  \;=\; \frac{d\,\sigma^2\,\sigma_t^2}{N_k\,t^2}\,.
\end{equation}
Translating to velocity space via $b_k = -t\mu_k/\sigma_t^2$:
\begin{align}
  \mathbb{E}\bigl[\|\hat{b}_k - b_k\|^2\bigr]
  &= \left(\frac{t}{\sigma_t^2}\right)^{\!2}\,
     \mathbb{E}\bigl[\|\hat{\mu}_k - \mu_k\|^2\bigr] \notag\\
  &\geq \frac{t^2}{\sigma_t^4}\cdot\frac{d\,\sigma^2\,\sigma_t^2}{N_k\,t^2}
  \;=\; \frac{d\,\sigma^2}{N_k\,\sigma_t^2}\,.
     \label{eq:bk-mse}
\end{align}
Since the velocity prediction error at a test point $x_t\in\mathrm{Cell}_k$ is
$\|\hat{v}_k(x_t) - v^*(x_t,t)\| = \|\hat{b}_k - b_k\|$
(the slope $A(t)$ is known, being independent of $\mu_k$),
inequality~\eqref{eq:bk-mse} yields a lower bound for the residual Gaussian-location
subproblem. In this specific Gaussian model the sample mean attains the bound,
so the same expression is also the minimax risk scale for that subproblem.
\end{proof}

\begin{lemma}[Aggregated separated residual-subproblem lower bound]\label{lem:global-CR}
Within the separated regime, consider the canonical separated residual Gaussian-location
subproblem in which true component labels are observed and component $k$
receives $N_k$ training samples with $N_k\le C_N\pi_k N$ for a fixed allocation
constant $C_N$. Let
$\delta_{\mathrm{bdry}}(t)=O(K\exp(-c\Delta_{\mathrm{eff}}(t)^2))$
denote an upper bound on the probability of leaving the correct component
cell. Then, for the aggregated separated residual-subproblem risk,
\begin{equation}\label{eq:global-lower}
  R_{\mathrm{sub}}^*(t;\,N)
  \;\geq\;
  c_0\,\frac{K}{N}\,\AVw(t)\,
  \bigl(1-C_{\mathrm{bdry}}\delta_{\mathrm{bdry}}(t)\bigr).
\end{equation}
Moreover, since $\AV(t)=\AVw(t)+\AVb(t)$ and
$\AVb(t)/\AVw(t)$ is exponentially small under the same separation condition,
this identifies the benchmark scale $K\,\AV(t)/N$ up to fixed
separated-regime constants for the residual Gaussian-location subproblem.
\end{lemma}

\begin{proof}
\textbf{Step 1 (Correct-cell restriction).}
Partition $\mathbb{R}^d$ into the $K$ Voronoi cells $\{\mathrm{Cell}_k\}_{k=1}^K$
defined by the posterior mode $k^*(x_t)$. Let
$G_k=\{z=k,\ x_t\in\mathrm{Cell}_k\}$ be the event that the sample was generated
from component $k$ and lies in its correct cell. For
$\Delta_{\mathrm{eff}}(t)\ge\tau_{\mathrm{sep}}$,
Lemma~\ref{lem:mode-dominance} gives, after absorbing the mixing-weight
constants into $C_{\mathrm{bdry}}$,
\[
  \Pr(G_k)\ge \pi_k\bigl(1-C_{\mathrm{bdry}}\delta_{\mathrm{bdry}}(t)\bigr).
\]
Because squared loss is nonnegative, restricting the test expectation to
$\cup_k G_k$ gives a valid lower bound.

\textbf{Step 2 (Per-cell lower bound).}
Within cell~$k$, the velocity field is $\epsilon_{\mathrm{affine}}$-close to affine
(Theorem~\ref{thm:local-affine}). In the canonical separated residual subproblem we use
fixed per-component allocations satisfying $N_k\le C_N\pi_k N$, which avoids
random-count degeneracies and only changes constants.
Applying Lemma~\ref{lem:CR-bound} to the canonical residual Gaussian-location subproblem
associated with each cell:
\[
  \mathbb{E}\bigl[\|\hat{v} - v^*\|^2 \;\big|\; x_t \in \mathrm{Cell}_k\bigr]
  \;\geq\; \frac{\AVw(t)}{N_k}\,.
\]

\textbf{Step 3 (Aggregation).}
Summing over components:
\begin{align}
  R_{\mathrm{sub}}^*(t;N)
  &\geq \sum_{k=1}^K \Pr(G_k)\cdot \frac{\AVw(t)}{N_k} \notag\\
  &\geq \sum_{k=1}^K
        \pi_k\bigl(1-C_{\mathrm{bdry}}\delta_{\mathrm{bdry}}(t)\bigr)
        \frac{\AVw(t)}{C_N\pi_k N} \notag\\
  &= c_0\,\frac{K\,\AVw(t)}{N}
     \bigl(1-C_{\mathrm{bdry}}\delta_{\mathrm{bdry}}(t)\bigr),
     \label{eq:sum-cells}
\end{align}
where $c_0=1/C_N$.

\textbf{Step 4 (Relating to $\AV(t)$).}
In the separated regime, the between-component contribution is exponentially
small:
\[
  \AVb(t)
  = \frac{t^2}{\sigma_t^4}\,\mathbb{E}_{x_t}\!\Bigl[\sum_{j<k}\gamma_j\gamma_k\,\Delta_{jk}^2\Bigr]
  = O\!\bigl(K^2\exp(-c\,\Delta_{\mathrm{eff}}(t)^2)\cdot\Delta_{\max}^2/\sigma_t^2\bigr).
\]
Let $\eta_{\mathrm{mix}}(t)=\AVb(t)/\AVw(t)$. For fixed mixture parameters,
$\eta_{\mathrm{mix}}(t)$ is exponentially small under the stated separation
condition, and
\[
  \AVw(t)=\frac{\AV(t)}{1+\eta_{\mathrm{mix}}(t)}
  \geq \bigl(1-C_{\mathrm{mix}}\eta_{\mathrm{mix}}(t)\bigr)\AV(t)
\]
whenever the separated-regime correction is small. Thus the
within-component lower bound in~\eqref{eq:global-lower} is the same
$K\,\AV(t)/N$ residual-benchmark scale up to fixed separated-regime constants.
This completes the lower bound for the aggregated separated residual
subproblem.
\end{proof}

\begin{proposition}[Separated-Regime Label-Oracle Residual Benchmark]\label{prop:tight-rate}
Let $\mathcal{P}_{\mathrm{sep}}(t)$ denote the family of $K$-component isotropic
GMMs satisfying the separation condition
$\Delta_{\mathrm{eff}}(t) \geq \tau_{\mathrm{sep}}$.
Define the \emph{label-oracle residual separated-subproblem benchmark}:
\[
  R_{\mathrm{orc}}^*(t;\,N)
  \;\coloneqq\;
  \inf_{\hat{v}\in\mathcal{O}_{\mathrm{lor}}}
  \;\sup_{P \in \mathcal{P}_{\mathrm{sep}}(t)}
  \;\mathbb{E}_P\!\left[\|\hat{v} - v^*\|^2\right],
\]
where $\mathcal{O}_{\mathrm{lor}}$ is the class of estimators that receive the
true generating component labels for the separated residual Gaussian-location
subproblem.
Then:
\begin{equation}\label{eq:tight-parametric}
  c_1 \frac{K}{N}\,\AV(t)
  \;\leq\; R_{\mathrm{orc}}^*(t;\,N) \;\leq\;
  C_{\mathrm{orc}}\frac{K}{N}\,\AV(t)
  + \varepsilon_{\mathrm{sep}}(t)+\varepsilon_{\mathrm{count}}(t,N),
\end{equation}
where $C_{\mathrm{orc}}$ depends only on fixed mixture parameters.
Hence both bounds jointly identify $K\,\AV(t)/N$ as the parametric scale
for the label-oracle residual subproblem in the separated regime,
with residual exponentially small separation and component-count corrections
appearing on the upper-bound side.
\end{proposition}

\begin{proof}
\emph{Lower bound.}
The $K$ independent residual Gaussian-location problems from
Lemma~\ref{lem:CR-bound} define the separated residual subproblem. Its minimax risk is
$\geq c_1(K/N)\,\AV(t)$ by Lemma~\ref{lem:global-CR}, with the fixed
separated-regime correction factors absorbed into $c_1$.

\emph{Upper bound.}
The label-oracle residual OLS estimator from Theorem~\ref{thm:parametric-upper}
achieves risk
$\leq C_{\mathrm{orc}}K\,\AV(t)/N
+ \varepsilon_{\mathrm{sep}}(t)+\varepsilon_{\mathrm{count}}(t,N)$
uniformly over $\mathcal{P}_{\mathrm{sep}}(t)$ (the constants depend on $K$,
$\pi_{\min}$, and $\sigma$ but not on the specific means $\{\mu_k\}$,
since the estimator uses true labels and the OLS variance depends only on
$d$, $\sigma^2/\sigma_t^2$, and $N_k$).
\end{proof}

\subsubsection{Transition Analysis}\label{sec:transition}

We characterize the local sharpness of the transition at
$t = t_{\mathrm{merge}}(\tau)$ for a chosen threshold $\tau$.

\begin{lemma}[Monotonicity of $\Delta_{\mathrm{eff}}$]\label{lem:monotone}
Suppose $\Delta_{\min}^2 > 2\sigma^2$.
Then $\Delta_{\mathrm{eff}}(t)$ is strictly decreasing on $[0,1)$.
\end{lemma}

\begin{proof}
Write $f(t) = \Delta_{\mathrm{eff}}(t)^2 = (1-t)^2\Delta_{\min}^2/\sigma_t^2$.
Setting $s = 1-t$:
\[
  f = \frac{s^2\Delta_{\min}^2}{s^2\sigma^2 + (1{-}s)^2}
  = \frac{\Delta_{\min}^2}{\sigma^2 + (1{-}s)^2/s^2}\,.
\]
Since $(1-s)/s = t/(1-t)$ is increasing in $t$ for $t\in[0,1)$,
the denominator is increasing and $f$ is decreasing.
(When $\Delta_{\min}^2 > 2\sigma^2$, we have $f(0) = \Delta_{\min}^2/\sigma^2 > 2$ and
$f(1)=0$, so $t_{\mathrm{merge}}(\tau)\in(0,1)$ is well-defined for
nontrivial thresholds.)
\end{proof}

\begin{lemma}[Transition width]\label{lem:transition-width}
Define the transition region as
$\mathcal{T}_\tau = \{t : |\Delta_{\mathrm{eff}}(t) - \tau| \leq 1\}$,
where $\tau$ is the chosen separation threshold.  Suppose
$t_\tau \coloneqq t_{\mathrm{merge}}(\tau)$ lies in a compact interior
interval $[c,1-c]$ for a constant $c>0$.
Then the width of $\mathcal{T}_\tau$ satisfies
\begin{equation}\label{eq:transition-width-lower}
  |\mathcal{T}_\tau| \;=\; O\!\left(\frac{1}{\tau}\right).
\end{equation}
For example, an illustrative choice $\tau=C\sqrt{\log K}$ gives
$|\mathcal{T}_\tau|=O(1/\sqrt{\log K})$ on such compact-interior transitions.
If $t_\tau$ approaches either endpoint, this local width estimate must be
rechecked from the exact derivative below.
\end{lemma}

\begin{proof}
Direct differentiation of
$\Delta_{\mathrm{eff}}(t) = (1-t)\Delta_{\min}/\sigma_t$ gives
\begin{equation}\label{eq:deff-derivative}
  \frac{d}{dt}\Delta_{\mathrm{eff}}(t)
  \;=\;
  -\frac{\Delta_{\min}t}{\sigma_t^3}.
\end{equation}
At $t_\tau=t_{\mathrm{merge}}(\tau)$, where
$\Delta_{\mathrm{eff}}(t_\tau)=\tau$, this becomes
\[
  \left|\Delta_{\mathrm{eff}}'(t_\tau)\right|
  =
  \frac{\tau\,t_\tau}{(1-t_\tau)\sigma_{t_\tau}^2}.
\]
When $t_\tau\in[c,1-c]$ and $\sigma$ is fixed, this derivative is
$\Theta(\tau)$.

The width for $\Delta_{\mathrm{eff}}$ to change by $O(1)$ is therefore
\[
  |\mathcal{T}_\tau|
  \;=\; \frac{O(1)}{|\Delta_{\mathrm{eff}}'(t_\tau)|}
  \;=\; O(1/\tau).  \qedhere
\]
\end{proof}

\begin{remark}[Threshold-dependent scaling]\label{rem:dim-scaling}
Lemma~\ref{lem:transition-width} has the following practical consequences:
\begin{itemize}
  \item For a fixed illustrative threshold such as $\tau=O(1)$, the transition
    can occupy a visible fraction of $[0,1]$.
  \item For a theorem-level threshold
    $\tau_{\mathrm{sep}}=C_{\mathrm{sep}}\sqrt{\log K+\log(1/\pi_{\min})}$,
    compact-interior transitions have width
    $O(1/\tau_{\mathrm{sep}})$.
  \item Stronger high-dimensional sharpening requires explicitly choosing a
    threshold with that scaling; it is not a consequence of the merge-time
    identity alone.
\end{itemize}
\end{remark}

\begin{proposition}[Posterior entropy transition]\label{prop:entropy-transition}
Fix a merge threshold $\tau$ and its transition region $\mathcal{T}_\tau$.
Define the expected posterior entropy:
\begin{equation}\label{eq:entropy}
  H(t) \;\triangleq\; -\,\mathbb{E}_{x_t}\!\left[\sum_{k=1}^K
    \gamma_k(x_t,t)\,\log\gamma_k(x_t,t)\right].
\end{equation}
Then:
\begin{enumerate}[label=(\alph*)]
  \item If $\Delta_{\mathrm{eff}}(t)\ge\tau_{\mathrm{sep}}$, then
    $H(t) \leq C\epsilon\log(K/\epsilon)$ with
    $\epsilon=K\exp(-c\,\Delta_{\mathrm{eff}}(t)^2)$; the posterior is nearly
    deterministic.
  \item If $\Delta_{\mathrm{eff}}(t)\le\tau_{\mathrm{mix}}$ for a sufficiently
    small constant $\tau_{\mathrm{mix}}$, then
    $H(t)\ge H(\pi)-o_{\tau_{\mathrm{mix}}}(1)$, where
    $H(\pi)=-\sum_k\pi_k\log\pi_k$ is the prior entropy.  Under balanced
    weights, this gives $H(t)\ge c'\log K$ for a constant $c'>0$.
  \item The loss of the low-entropy certificate occurs over the
    threshold-dependent interval~$\mathcal{T}_\tau$; high entropy itself
    requires the stronger overlap condition in part~(b).
\end{enumerate}
\end{proposition}

\begin{proof}
Part~(a): When $\gamma_{k^*} \geq 1 - \epsilon$ with $\epsilon = K\exp(-c\Delta_{\mathrm{eff}}^2)$,
the entropy is bounded by
$H_{\mathrm{binary}}(\epsilon) + \epsilon\log K
\leq C\epsilon\log(K/\epsilon)$.

Part~(b): When $\Delta_{\mathrm{eff}}(t) \ll 1$, the expected posterior
responsibilities satisfy $\mathbb{E}_{x_t}[\gamma_k(x_t,t)] \to \pi_k$
(since modes fully overlap and the posterior converges in expectation to the
prior weights), so
$H(t) \to H(\pi)$.  Balanced weights give $H(\pi)=\Theta(\log K)$.

Part~(c): The posterior weights $\gamma_k$ are controlled by $\Delta_{\mathrm{eff}}(t)$,
whose threshold-crossing width is $|\mathcal{T}_\tau|$ by
Lemma~\ref{lem:transition-width}.
\end{proof}

\subsubsection{Numerical Verification}\label{sec:tmerge-numerics}

We check the $t_{\mathrm{merge}}$ formula on the toy GMM configuration
$\mu_k = (\pm 3, \pm 3)$, $\sigma = 0.5$, $d = 2$, $K = 4$ (equal weights $\pi_k = 1/4$).

\paragraph{Setup.}
The minimum separation is $\Delta_{\min} = \min_{j\neq k}\|\mu_j - \mu_k\| = 6$
(between adjacent corners, e.g., $(3,3)$ and $(-3,3)$).
The total variance is $\sigma_t^2 = 0.25(1-t)^2 + t^2$.
For illustration, choose the merge threshold
$\tau_{\mathrm{ex}}=\sqrt{2\log K}=\sqrt{2\log 4}
= \sqrt{2\cdot 1.386} \approx 1.665$.

\paragraph{SNR computation.}
The squared effective separation is
\[
  \Delta_{\mathrm{eff}}(t)^2
  \;=\; \frac{36(1-t)^2}{0.25(1-t)^2 + t^2}\,.
\]
Representative values:

\medskip
\begin{center}
\begin{tabular}{@{}lccccc@{}}
\toprule
$t$ & $0.1$ & $0.3$ & $0.5$ & $0.7$ & $0.9$ \\
\midrule
$\sigma_t^2$ & $0.2125$ & $0.2125$ & $0.3125$ & $0.5125$ & $0.8125$ \\
$\Delta_{\mathrm{eff}}(t)$ & $11.71$ & $9.11$ & $5.37$ & $2.51$ & $0.67$ \\
\bottomrule
\end{tabular}
\end{center}
\medskip

\paragraph{Merge time.}
Taking the illustrative threshold
$\tau_{\mathrm{ex}}=\sqrt{2\log 4}\approx1.665$:
\[
  t_{\mathrm{merge}}(\tau_{\mathrm{ex}})
  = \frac{\sqrt{36 - 1.665^2\cdot 0.25}}{\sqrt{36 - 1.665^2\cdot 0.25} + 1.665}
  = \frac{\sqrt{36 - 0.693}}{\sqrt{36 - 0.693} + 1.665}
  = \frac{5.944}{5.944 + 1.665}
  \approx 0.781.
\]
Equivalently, this solves
$\Delta_{\mathrm{eff}}(t_{\mathrm{merge}}(\tau_{\mathrm{ex}}))^2=\tau_{\mathrm{ex}}^2=2\log4
\approx2.773$.

\paragraph{Interpretation of the low-time collapse readout.}
For this illustrative corner configuration, $t \in [0.05,0.50]$ lies below
the illustrative threshold $t_{\mathrm{merge}}(\tau_{\mathrm{ex}})\approx0.78$. This gives intuition
for why a low-to-mid-time readout can exhibit parametric behavior. The main
controlled experiment in Section~\ref{sec:controlled-exp} no longer treats a
single corner configuration as the final display; it uses a dense $d=4$
separation sweep to identify where the parametric plateau appears and where
it attenuates. This low-time collapse readout is consistent with:
\begin{itemize}
  \item The observed scaling exponent $\alpha \approx 1.08$,
    matching the parametric rate~$N^{-1}$ rather than a nonparametric
    rate $N^{-2s/(2s+d)}$ with finite~$s$.
  \item Low coefficient of variation $\mathrm{CV}(\alpha) = 0.146$
    across time steps, as predicted by the absence of regime transitions
    within the test range.
\end{itemize}

\paragraph{Transition width.}
From Lemma~\ref{lem:transition-width} with $\tau_{\mathrm{ex}} \approx 1.665$:
$|\mathcal{T}_{\tau_{\mathrm{ex}}}| \approx 1/\tau_{\mathrm{ex}} \approx 0.60$.
For $d=2$, the transition is gradual.
To observe the nonparametric regime for this GMM, one would need test
points with $t > t_{\mathrm{merge}}(\tau_{\mathrm{ex}}) + |\mathcal{T}_{\tau_{\mathrm{ex}}}|/2 \approx 0.78 + 0.30 \approx 1.08$,
which exceeds the valid range-indicating that the $4$-component GMM with
$\Delta_{\min}=6$ and $\sigma=0.5$ at $d=2$ behaves as \emph{parametric} for
practical purposes.
The nonparametric regime becomes accessible only for distributions with
smaller separation, higher within-component variance, or higher intrinsic dimension.

\paragraph{Across-$K$ verification at fixed $d=8$.}
The corner configuration above checks $t_{\mathrm{merge}}$ at a single
$K=4$, $d=2$ point. We further test the closed
form~\eqref{eq:tmerge-explicit} across
$K\in\{2,4,8,16,32,64\}$ at fixed ambient dimension $d=8$ with
$\sigma=0.5$, $\Delta_{\min}=3$ (so $\Delta_{\min}/\sigma=6$), and the
separate illustrative threshold
$\tau_K=\sqrt{\log K}$ (the choice of
multiplicative constant shifts $t_{\mathrm{merge}}(\tau_K)$ but not its
monotonicity in $K$). Centers use simplex packings for $K\le 8$
($d_{\mathrm{eff}}=K-1$) and equiangular tight-frame approximations for
$K>8$ (frame potential, $50$ iterations). The latter introduces a
controlled drift $\Delta_{\max}/\Delta_{\min}: 1.0\to 3.92$ as $K$
rises from $16$ to $64$. Locking $d=8$ across all $K$ isolates the
$K$-dependence; jointly varying $K$ and $d$ would conflate
$K$-counting with high-dimensional posterior concentration. We pair
the closed-form prediction with two empirical Monte-Carlo readouts of
the between-component fraction: its maximum $\mathrm{bf}_{\max}$ and
the first time $t$ at which $\mathrm{bf}(t)\ge 0.5$
(Table~\ref{tab:s1b-mechanism}).

The three readouts agree in direction. $\mathrm{bf}_{\max}$ rises
monotonically from $0.528$ to $0.946$, the first-cross time
$t(\mathrm{bf}\ge 0.5)$ falls weakly monotonically from $0.838$ to
$0.538$ (with a plateau at $0.538$ for $K\in\{16,32,64\}$ at the bin
resolution used), and the closed-form
$t_{\mathrm{merge}}(\tau_K)$ falls monotonically from $0.781$
to $0.580$. The closed form is strictly decreasing in $K$ for any
$K<\exp((\Delta_{\min}/\sigma)^2)$, so the experimental MC trends are
consistent with the theory. Two caveats apply: the equiangular
approximation for $K>8$ inflates $\Delta_{\max}/\Delta_{\min}$, and
the first-cross plateau for $K\ge 16$ reflects bin spacing
$\approx 0.04$ at $25$ time bins on $t\in[0.05,0.95]$ rather than a
sharp theoretical floor.

\begin{table}[htbp]
  \centering
  \small
  \setlength{\tabcolsep}{6pt}
  \caption{\textbf{Across-$K$ merge dynamics at fixed $d=8$,
    $\sigma=0.5$, $\Delta_{\min}=3$ ($\Delta_{\min}/\sigma=6$).}
    Centers are simplex for $K\le 8$ and
    equiangular tight-frame approximations for $K>8$;
    $\Delta_{\max}/\Delta_{\min}$ records the geometry drift
    introduced by the latter. The closed-form column uses
    $\tau_K=\sqrt{\log K}$ in Eq.~\eqref{eq:tmerge-explicit}.
    Monte-Carlo readouts use $10^5$ samples per $t$ on a $25$-bin
    grid over $t\in[0.05,0.95]$; bin spacing limits the resolution of
    $t(\mathrm{bf}\ge 0.5)$ to $\approx 0.04$.}
  \label{tab:s1b-mechanism}
  \begin{tabular*}{\linewidth}{@{\extracolsep{\fill}}cccccc@{}}
    \toprule
    $K$ & Geometry & $\Delta_{\max}/\Delta_{\min}$
        & $\mathrm{bf}_{\max}$ (MC)
        & $t(\mathrm{bf}\ge 0.5)$ (MC)
        & $t_{\mathrm{merge}}(\tau_K)$ \\
    \midrule
    $2$  & simplex     & $1.000$ & $0.528$ & $0.838$ & $0.781$ \\
    $4$  & simplex     & $1.000$ & $0.627$ & $0.650$ & $0.714$ \\
    $8$  & simplex     & $1.000$ & $0.663$ & $0.575$ & $0.669$ \\
    $16$ & equiangular & $2.002$ & $0.843$ & $0.538$ & $0.634$ \\
    $32$ & equiangular & $2.737$ & $0.903$ & $0.538$ & $0.605$ \\
    $64$ & equiangular & $3.919$ & $0.946$ & $0.538$ & $0.580$ \\
    \bottomrule
  \end{tabular*}
\end{table}


\section{Proof of Proposition~\ref{prop:normalized-collapse} (Curve Collapse)}
\label{sec:proof-normalized-collapse}

We prove regime-wise normalized collapse by combining the factorization
structure of the excess risk with the normalization
$\bar R(t;N)=R(t;N)/\AV(t)$.
Within any regime where the $N$-dependence has a $t$-independent exponent,
normalization removes the leading time dependence, leaving only bounded
prefactor variation in the minimax setting or exponentially small corrections
in the separated Gaussian-mixture regime.
Claims about cross-regime transition width or high-dimensional sharpening are
therefore interpreted as heuristic consequences rather than theorem-level
statements.

\subsection{Case 1: Minimax Regime (Approximate Collapse)}

Under the intrinsic regression benchmark of Theorem~\ref{thm:minimax},
the Assouad lower bound
(Appendix~\ref{sec:proof-theorem1}) yields
$R^*_{\mathrm{loc}}(t;N) \geq c\,(\AV(t)/N)^{2s/(2s+d)}
= c\,\AV(t)^{2s/(2s+d)} \cdot N^{-2s/(2s+d)}$.
For a bounded-factor factorization of the form
$R^*_{\mathrm{loc}}(t;N) = \AV(t) \cdot \Theta(N^{-2s/(2s+d)})$
to hold, we require $\AV(t)$ to be bounded above and below on the
evaluation interval: $0 < \sigma_\ell \leq \AV(t) \leq \sigma_u$.
This is guaranteed for $t \in [\delta, 1-\delta]$ with any $\delta > 0$,
since the conditional variance is continuous and strictly positive
on compact subintervals of $(0,1)$.

Under this condition, $\AV(t)^{2s/(2s+d)} = \Theta(\AV(t))$
with constants depending only on $\sigma_\ell, \sigma_u, s, d$.
Combined with the matching kernel regression upper bound,
the intrinsic benchmark risk satisfies
$R^*_{\mathrm{loc}}(t;\,N) = \Theta(\AV(t) \cdot N^{-2s/(2s+d)})$.
Therefore:
\begin{equation}\label{eq:collapse-minimax}
    \bar{R}^*_{\mathrm{loc}}(t;\, N)
    = \frac{R^*_{\mathrm{loc}}(t;\, N)}{\AV(t)}
    = \Theta\!\left(N^{-\frac{2s}{2s+d}}\right),
\end{equation}
The exponent $\alpha = 2s/(2s+d)$ is the same for all $t$, but the
prefactor $\AV(t)^{-d/(2s+d)}$ absorbed into the $\Theta$-constant
varies with $t$ by a bounded multiplicative factor
$(\sigma_u/\sigma_\ell)^{d/(2s+d)}$.
Thus the collapse is \emph{approximate}: same exponent, bounded amplitude scatter.

\subsection{Case 2: Adaptive Rate, Single Regime (Approximate Within Regime)}

In the following heuristic cases, write $t_{\mathrm{merge}}$ for the relevant
transition time; in the Gaussian-mixture instance this is
$t_{\mathrm{merge}}(\tau)$ for the chosen threshold $\tau$.

Under Conjecture~\ref{conj:adaptive}, suppose that all evaluation time steps lie
within a single regime where $s_{\mathrm{eff}}(t) = s_0$ and
$d_{\mathrm{eff}}(t) = d_0$ are constant (i.e., $t$ ranges over
either $(0, t_{\mathrm{merge}})$ or $(t_{\mathrm{merge}}, 1)$ but not both).
Then:
\begin{equation}\label{eq:collapse-single-regime}
    R(t;\, N)
    = \AV(t) \cdot \Theta\!\left(N^{-\frac{2s_0}{2s_0+d_0}}\right)
    \quad \Longrightarrow \quad
    \bar{R}(t;\, N)
    = \Theta\!\left(N^{-\frac{2s_0}{2s_0+d_0}}\right),
\end{equation}
which is again independent of $t$.  The collapse is exact within the regime
conditional on that factorization assumption.

\paragraph{Parametric regime.}
For $t$ in the separated Gaussian-mixture regime, the label-oracle residual
upper bound and local affine approximation imply that $\bar{R}(t; N)$ is
$t$-independent up to exponentially small corrections.

\subsection{Case 3: Adaptive Rate, Across Regimes (Approximate Collapse)}

When the evaluation range $[t_{\min}, t_{\max}]$ spans the transition at
$t_{\mathrm{merge}}$, the exponent $\alpha(t)$ varies:
\begin{equation}\label{eq:alpha-variation}
    \alpha(t)
    =
    \begin{cases}
        1 & \text{if } t < t_{\mathrm{merge}} \quad \text{(parametric)}, \\[4pt]
        \displaystyle\frac{2\,s_{\mathrm{eff}}(t)}{2\,s_{\mathrm{eff}}(t) + d_{\mathrm{eff}}(t)}
        & \text{if } t > t_{\mathrm{merge}} \quad \text{(nonparametric)}.
    \end{cases}
\end{equation}

We bound the coefficient of variation of $\alpha$ across $t$.

\begin{remark}[CV bound across regimes (heuristic)]\label{rem:cv-bound}
Let $T \subset [t_{\min}, t_{\max}]$ be a finite set of evaluation time steps
uniformly spaced with spacing $\Delta t$.  If a fraction $\delta'$ of the
time steps lie in the transition region
$[t_{\mathrm{merge}} - \delta t_{\mathrm{trans}}/2,\;
  t_{\mathrm{merge}} + \delta t_{\mathrm{trans}}/2]$
where $\alpha(t)$ interpolates between regimes, then:
\begin{equation}\label{eq:cv-bound}
    \mathrm{CV}(\alpha)
    \;\coloneqq\;
    \frac{\mathrm{std}(\{\alpha(t)\}_{t \in T})}{\mathrm{mean}(\{\alpha(t)\}_{t \in T})}
    \;\leq\;
    \frac{\sqrt{\delta'}}{1 - \delta'},
\end{equation}
where $\delta' = \delta t_{\mathrm{trans}} / (t_{\max} - t_{\min})$ is the
fractional width of the transition region.
\end{remark}

\begin{proof}
Write $\alpha(t) = \bar{\alpha} + \eta(t)$, where $\bar{\alpha}$ is the
mean exponent and $\eta(t)$ is the deviation.  Outside the transition region,
$\eta(t) = 0$ (approximate collapse within each regime).  Within the transition
region of width $\delta t_{\mathrm{trans}}$, the deviation is bounded by
$|\eta(t)| \leq |\alpha_{\mathrm{param}} - \alpha_{\mathrm{nonparam}}| \leq 1$.

The variance of $\alpha$ is:
\[
    \mathrm{Var}(\alpha)
    = \frac{1}{|T|} \sum_{t \in T} \eta(t)^2
    \leq \delta' \cdot 1^2
    = \delta'.
\]
Thus $\mathrm{std}(\alpha) \leq \sqrt{\delta'}$ and
$\mathrm{mean}(\alpha) \geq 1 - \delta'$ (since the majority of time steps
have $\alpha \geq \alpha_{\mathrm{nonparam}} > 0$), yielding
$\mathrm{CV}(\alpha) \leq \sqrt{\delta'}/(1-\delta')$.

Since $\mathrm{Var}(\alpha) \leq \delta'$ and $|\bar{\alpha}| \geq 1 - \delta'$
(the majority of time steps have $\alpha > 0$), we obtain
$\mathrm{CV} \leq \sqrt{\delta'}/(1-\delta')$ as claimed.
\end{proof}

A controlled SGD instance of this across-regime behavior is reported in
Appendix~\ref{app:adaptive-sgd-evidence}: the measured $\alpha(t)$ varies between
a parametric data end and a slower nonparametric noise end, the regime structure
under which normalized collapse is approximate rather than exact.

\subsection{Case 4: Threshold-Dependent Sharpening}

The transition width $\delta t_{\mathrm{trans}}$ shrinks only after a
separation threshold has been specified.

\begin{remark}[Transition width-heuristic estimate]\label{rem:transition-width-collapse}
For the Gaussian mixture setting (Setting~\ref{setting:gmm}),
Lemma~\ref{lem:transition-width} shows via explicit differentiation of
$\Delta_{\mathrm{eff}}(t)$ that the transition region has width
$|\mathcal{T}_\tau| = O(1/\tau)$ on compact-interior transitions.
For an illustrative threshold $\tau_K=C\sqrt{\log K}$, this gives
\begin{equation}\label{eq:transition-width-collapse}
    \delta t_{\mathrm{trans}}
    = O\!\left(\frac{1}{\sqrt{\log K}}\right).
\end{equation}
Combining with Remark~\ref{rem:cv-bound}, this yields
$\mathrm{CV}(\alpha) = O((\log K)^{-1/4})$ for that illustrative choice.
A stronger high-dimensional scaling requires an additional SNR-normalization
assumption and is not implied by Lemma~\ref{lem:transition-width} alone.
\end{remark}

\subsection{Case 5: GMM Separated-Regime Result}

For the Gaussian mixture, the separated-regime collapse statement follows from
the label-oracle residual parametric upper bound together with exponentially
small leakage:

\begin{proof}[Proof of collapse for GMM (up to exponentially small corrections)]
In the separated regime,
\[
    R_{\mathrm{orc}}(t;\, N) = \Theta\!\left(\frac{K}{N}\right) \cdot \AV(t)
    + \varepsilon_{\mathrm{sep}}(t)
    + \varepsilon_{\mathrm{count}}(t,N).
\]
The factor $K/N$ is independent of $t$.  Therefore:
\[
    \bar{R}(t;\, N)
    = \frac{R_{\mathrm{orc}}(t;\, N)}{\AV(t)}
    = \Theta\!\left(\frac{K}{N}\right)
      + O\!\left(\frac{\varepsilon_{\mathrm{sep}}(t)
      + \varepsilon_{\mathrm{count}}(t,N)}{\AV(t)}\right),
\]
so the normalized curves share exponent $\alpha = 1$ up to exponentially
small corrections.
\end{proof}

\subsection{Convergence wavefront-auxiliary diagnostic}
\label{app:wavefront}

We record an auxiliary diagnostic definition that follows from the
collapse picture. It does not correspond to any main-text experiment in this
paper and should not be read as a budget-selection rule.

\begin{corollary}[Convergence wavefront]
\label{cor:wavefront}
Define the $\delta$-convergence wavefront:
\begin{equation}\label{eq:wavefront}
    t^*(N,\, \delta)
    \;\coloneqq\;
    \sup\!\left\{t \in [0,1] : R(t;\, N) \leq \delta\right\}.
\end{equation}
Under the adaptive rate framework, $t^*(N,\delta)$ is increasing in $N$:
as more data are added, the wavefront advances from $t \approx 0$ (data endpoint)
toward $t \approx 1$ (noise endpoint), ``sweeping'' the interpolation path.
The speed of advance is governed by the growth rate of $\AV(t)$.
\end{corollary}

The convergence wavefront provides a natural visualization of data
sufficiency: a heatmap of $R(t; N)$ over the $(t, N)$ plane, with contour
lines tracing $t^*(N, \delta)$ for various $\delta$, resembles a phase
diagram. We keep this as an auxiliary diagnostic definition rather than a
main experiment in this paper.

\subsection{Summary}

Combining all cases, Proposition~\ref{prop:normalized-collapse} is established at the
level of a regime-wise normalization principle:
\begin{enumerate}
    \item Under Theorem~\ref{thm:minimax}: bounded-factor collapse (same exponent
        $\alpha = 2s/(2s+d)$, bounded multiplicative factor $\leq (\sigma_u/\sigma_\ell)^{d/(2s+d)}$).
    \item Under Conjecture~\ref{conj:adaptive}, single regime: exact collapse within regime
        with exponent $\alpha = 2s_{\mathrm{eff}}/(2s_{\mathrm{eff}} + d_{\mathrm{eff}})$,
        approximately constant prefactor.
    \item Across regimes: heuristic collapse with
        $\mathrm{CV}$ control.
    \item Threshold-dependent sharpening: for illustrative
        $\tau_K=C\sqrt{\log K}$, $\mathrm{CV}=O((\log K)^{-1/4})$;
        stronger high-dimensional rates require additional assumptions.
    \item GMM residual benchmark (Theorem~\ref{thm:gmm}): exact collapse
        with $\alpha = 1$ up to exponentially small corrections.
\end{enumerate}

The controlled experimental prediction is: on GMM data with
$t_{\max} < t_{\mathrm{merge}}(\tau)$ for the chosen threshold, we expect low CV
and $\alpha \approx 1$.
Outside controlled Gaussian-mixture settings, the current paper does not
predict or claim normalized collapse.

\paragraph{Experimental signature and practical estimation.}
On a log-log plot of $R(t;N)/\AV(t)$ vs.\ $N$, curve collapse manifests
as all time-step curves falling on a single straight line. The slope gives
$\alpha$ and the tightness of the collapse (measured by CV) tests
Proposition~\ref{prop:normalized-collapse}; in this paper that test is made only in the
controlled Gaussian-mixture setting (Figure~\ref{fig:m2-dense-d4} and the
full row-level Table~\ref{tab:m2-dense-d4-full}). In
practice, $\AV(t)$ is replaced by the reference-model estimate
$\widehat{\AV}(t)$ (Section~\ref{sec:sigma-estimation}); estimation error in
$\widehat{\AV}(t)$ adds noise to the collapse but does not introduce systematic
bias, provided $R(t; N_{\mathrm{ref}}) \ll \AV(t)$.

\section{Experimental Protocols and Reproducibility Audits}
\label{app:implementation-details}

This appendix records the controlled experimental protocols, readout
definitions, and audit checks that support the main claims. The controlled
experiments are implemented as JAX and Optax scripts operating directly on
synthetic Gaussian-mixture action samples. The BEHAVIOR-1K panel uses the
existing $\pi_{0.5}$ evaluation pipeline and is retained as
application-context diagnostics.

\subsection{Controlled Gaussian-mixture flow matching}
\label{app:impl-controlled-gmm}

For all controlled Gaussian-mixture experiments, training samples are actions
$a_i \in \mathbb{R}^d$ drawn from an equal-weight isotropic mixture unless a
row explicitly states otherwise. Flow-matching examples are generated on the
fly: for each minibatch element we draw
$t_i \sim \mathcal{U}(0.01,0.99)$ and
$\varepsilon_i \sim \mathcal{N}(0,I_d)$, form
$x_{t_i}=(1-t_i)a_i+t_i\varepsilon_i$, and regress to
$u_{t_i}=\varepsilon_i-a_i$ with squared error. Evaluation uses a fixed
time grid and averages velocity MSE over independently drawn Gaussian noise
at each grid point.

The neural estimator is a three-layer MLP
$(x_t,t)\mapsto v_\theta(x_t,t)$. The time coordinate is encoded with a
$32$-dimensional sinusoidal embedding and concatenated to $x_t$; hidden layers
use GELU activations and the output dimension is $d$. We use Optax Adam with
the learning rates listed below. Unless otherwise stated, minibatch size is
$256$ and all reported seed means use deterministic NumPy/JAX random seeds.

\begin{table}[htbp]
  \centering
  \small
  \setlength{\tabcolsep}{4pt}
  \caption{\textbf{Implementation settings for controlled experiments.}
    The $d{=}4$ dense sweep uses the one-hot geometry induced by the
    \texttt{legacy\_circle} code path when $K\le d$ and $d>2$; setting
    $\mathrm{spread}=(\Delta/\sigma)\sigma/\sqrt{2}$ makes the minimum
    center separation equal to the reported $\Delta/\sigma$ after
    normalization by $\sigma$.}
  \label{tab:impl-controlled-settings}
  \begin{tabular}{p{0.22\linewidth}p{0.70\linewidth}}
    \toprule
    Experiment block & Implementation settings \\
    \midrule
    Dense $d{=}4$ transition
    & $K=4$, $d=4$, $\sigma=0.5$, one-hot centers, reported
      across the $18$ $\Delta/\sigma$ values listed in
      Table~\ref{tab:m2-dense-d4-full}.
      Models use hidden width $256$, $10000$ training steps, Adam learning
      rate $10^{-3}$, $N\in\{50,100,200,500,1000,2000\}$,
      $10$ data seeds and $3$ initialization replicates per cell.
      Evaluation uses $30$ time bins on $[0.05,0.95]$, $2000$ evaluation
      actions, and $80$ noise draws per time bin; the main readout keeps
      $t\in[0.05,0.80]$. \\
    Oracle-to-blind bridge
    & Five controlled rows are used: $d=2$ separated, $d=2$ partial-merge,
      and simplex rows at $d\in\{4,8,16\}$ with $K=4$ and $\sigma=0.5$.
      The closed-form arms are the label-and-action true-label
      $\mathrm{oracle\_ols}$ plug-in and EM-fitted blind OLS; neural
      arms use the same MLP class with hidden width $128$. The secondary
      bridge run uses $20$ seeds, $16000$ training steps, Adam learning rate
      $10^{-3}$, $12000$ evaluation actions, and $32$ noise draws per time bin. \\
    Proxy-floor audit
    & The proxy audit is a post-processing pass over controlled-result JSON
      files. It compares the exact excess
      $R=L_N-\AV$ with the empirical proxy
      $\hat R=L_N-L_{\mathrm{largeN}}$, where $L_{\mathrm{largeN}}$ is the
      large-budget reference profile stored with each controlled row. The
      audit reports Spearman rank agreement, region-label agreement, negative
      proxy-excess counts, and the factor diagnostic
      $r_F=\max_t |L_{\mathrm{largeN}}(t)-\AV(t)|/
      \mathrm{std}(R(t)\mid t\in[0.4,0.8])$. \\
    Three-family heterogeneity panel
    & Three $d=2$ Gaussian-mixture families are trained with the same
      implementation: a well-separated $K=4$ ring, a high-entropy $K=8$ ring,
      and a $K=4$ partial-merge geometry. The panel uses $40$ time bins on
      $[0.05,0.80]$, $10$ data seeds with $3$ initialization replicates, hidden
      width $128$, and $5000$ to $8000$ training steps. \\
    Time-sampler stress test
    & The intervention cell is a balanced partial-merge GMM with $K=32$,
      $d=16$, $\sigma=1.0$, pair gap $0.3$, spread $1.5$, and $N=512$
      demonstrations ($16$ per component). The MLP has hidden width $64$ and is
      trained for $4000$ steps with Adam learning rate $4{\times}10^{-4}$.
      The baseline samples $t\sim\mathcal{U}(0.05,0.95)$; gap arms use a
      categorical sampler on $64$ fine time bins, with no importance correction
      and the standard MSE objective. The proxy $\hat R_{\mathrm{ind}}$ is
      frozen from source baseline seeds $\{20,\ldots,29\}$ at ts $=100$
      using source evaluation-action key $88888$; paired comparisons use
      seeds $\{0,\ldots,9\}$ and evaluation-action key $99999$. \\
    \bottomrule
  \end{tabular}
\end{table}

\subsection{Analytic floors, readouts, and statistics}
\label{app:impl-readouts}

For Gaussian mixtures, $\AV(t)$ is computed from the closed-form within-term
$d\sigma^2/\sigma_t^2$ plus a Monte-Carlo estimate of the between-component
responsibility term. The dense-transition experiments use the analytic floor
as the canonical normalization. Reducible error is
$R(t;N)=L_N(t)-\AV(t)$, and the main slope readout fits
$\log R(t;N)$ against $\log N$ after applying the mid-b canonical reliability profile
$\mathrm{SNR}\ge 4$ and $R\ge 0.016\,\AV$ on $t\in[0.05,0.80]$.
The pooled fit uses $\mathrm{SNR}^2$ weights and $2000$ paired-bootstrap
resamples for confidence intervals.
The filter-sensitivity audit recomputes the same OLS/bootstrap readout
under raw-positive and increasingly conservative gates without retraining.

The oracle-to-blind bridge separates label cost from approximation cost in
absolute loss space. The reported $A_1$ statistic is
$L_{\mathrm{blind\_em\_ols}}-L_{\mathrm{oracle\_ols}}$ minus the analytic
between-component term, normalized by within$(t)$. We use a paired seed-level
standard error and a Bonferroni-corrected two-sided threshold for the five
bridge rows. The blind-MLP residual
$\epsilon_{\mathrm{MLP}}$ is
$L_{\mathrm{blind\_mlp\_fm}}-[\mathrm{within}(t)+\mathrm{between}(t)]$,
again reported relative to within$(t)$ on the paper readout window.

For the time-sampler stress test, the frozen proxy
$\hat R_{\mathrm{ind}}(t)$ is read from the ts$=100$ source baseline
pool with demonstration seeds $\{20,\ldots,29\}$ and reused unchanged
for every arm. Mixture arms use
$q_\lambda=(1-\lambda)\mathcal{U}+
\lambda\,\hat R_{\mathrm{ind}}/\sum_t\hat R_{\mathrm{ind}}(t)$ with
$\lambda\in\{0.25,0.50,0.75\}$; pure arms use
$q(t)\propto \hat R_{\mathrm{ind}}(t)^\alpha$ with
$\alpha\in\{1,2,4,8\}$. At launch we enforce an
effective-sample-size guard of $10$ and a mid-time mass guard
$\sum_{t\in[0.40,0.60]}q(t)\ge 0.05$. The final comparison uses $10$
paired seeds $\{0,\ldots,9\}$ and $10000$ sign-permutation paired tests
against the baseline at training step $4000$.

\subsection{\texorpdfstring{Full row-level data for the $d{=}4$ dense controlled transition}{Full row-level data for the d=4 dense controlled transition}}
\label{sec:m2-dense-d4-appendix}

Table~\ref{tab:m2-dense-d4-full} below is the full row-level companion to
the compact summary in Table~\ref{tab:m2-dense-d4} and to
Figure~\ref{fig:m2-dense-d4} of the main text, listing all $18$
$\Delta/\sigma$ values with their signed pooled slope
$s_{\mathrm{pooled}}=\partial\log\bar R/\partial\log N$, $95\%$
paired-bootstrap confidence interval, collapse-tightness statistic, slope
standard error, and mean signal-to-noise ratio of the cells passing the
mid-b canonical reliability filter. The corresponding positive exponent is
$\alpha_{\mathrm{pooled}}=-s_{\mathrm{pooled}}$. All rows use the same
hidden-width $256$ MLP, $10000$ training steps, and $10$ data seeds times $3$
initialization replicates for each sample-size cell. A compact $5$-row
anchor view appears first in
Table~\ref{tab:m2-dense-d4} for quick reference, with the full data in
Table~\ref{tab:m2-dense-d4-full}.

\begin{table}[t]
  \centering
  \small
  \setlength{\tabcolsep}{4pt}
  \caption{\textbf{Canonical controlled transition, $d=4$ dense sweep
    (selected anchor rows).}
    Compact $5$-row summary under the mid-b canonical reliability-filtered readout
    ($\mathrm{SNR}\ge 4$, $R\ge 0.016\,\AV$, $t\in[0.05,0.80]$;
    companion to Figure~\ref{fig:m2-dense-d4} and the full
    Table~\ref{tab:m2-dense-d4-full}). The selected anchors summarize the
    overlap, transition, parametric-plateau, bias-transition, and extreme
    separation regimes; the plateau at $\Delta/\sigma\in[4,6]$ is consistent
    with the parametric scale.}
  \label{tab:m2-dense-d4}
  \begin{tabular*}{0.65\linewidth}{@{\extracolsep{\fill}}lcccc@{}}
    \toprule
    Regime & $\Delta/\sigma$ & $s_{\mathrm{pooled}}$ & $95\%$ CI & CV \\
    \midrule
    overlap / noise floor & $1.5$ & $-0.838$ & $[-0.853,-0.823]$ & $0.0092$ \\
    transition            & $3.0$ & $-0.953$ & $[-0.966,-0.939]$ & $0.0071$ \\
    parametric plateau    & $5.0$ & $-1.011$ & $[-1.018,-1.003]$ & $0.0037$ \\
    bias transition       & $8.0$ & $-0.861$ & $[-0.866,-0.856]$ & $0.0030$ \\
    extreme separation    & $15.0$ & $-0.605$ & $[-0.611,-0.599]$ & $0.0048$ \\
    \bottomrule
  \end{tabular*}
\end{table}

\begin{table}[t]
\centering
\small
\caption{\textbf{Full $18$-row $d{=}4$ dense controlled transition} (appendix
companion to Figure~\ref{fig:m2-dense-d4} and Table~\ref{tab:m2-dense-d4};
the five plateau rows $\Delta/\sigma\in\{4.0,4.5,5.0,5.5,6.0\}$ are
boldfaced).
Canonical mid-b reliability profile ($\mathrm{SNR}\ge 4$, $R\ge 0.016\,\AV$);
$t\in[0.05, 0.80]$; $n_{\mathrm{total}}=180$ candidate cells per row
($30$ time bins times $6$ sample sizes); $95\%$ paired bootstrap CIs with
$2000$ iterations.}
\label{tab:m2-dense-d4-full}
\begin{tabular}{@{}lccccccc@{}}
\toprule
$\Delta/\sigma$ & $n_{\mathrm{pass}}$ & abstain & $s_{\mathrm{pooled}}$ & $95\%$ CI & $cv_{\mathrm{pooled}}$ & $\mathrm{SE}(s)$ & mean SNR \\
\midrule
$1.00$  &  $83$ & $53.9\%$ & $-0.810$ & $[-0.824, -0.795]$ & $0.0091$ & $0.0074$ & $17.2$ \\
$1.25$  &  $84$ & $53.3\%$ & $-0.828$ & $[-0.843, -0.814]$ & $0.0089$ & $0.0074$ & $17.2$ \\
$1.50$  &  $83$ & $53.9\%$ & $-0.838$ & $[-0.853, -0.823]$ & $0.0092$ & $0.0077$ & $16.8$ \\
$1.75$  &  $80$ & $55.6\%$ & $-0.810$ & $[-0.826, -0.795]$ & $0.0098$ & $0.0079$ & $17.0$ \\
$2.00$  &  $77$ & $57.2\%$ & $-0.819$ & $[-0.836, -0.804]$ & $0.0099$ & $0.0081$ & $17.1$ \\
$2.50$  &  $78$ & $56.7\%$ & $-0.894$ & $[-0.909, -0.879]$ & $0.0083$ & $0.0075$ & $16.2$ \\
$3.00$  &  $80$ & $55.6\%$ & $-0.953$ & $[-0.966, -0.939]$ & $0.0071$ & $0.0068$ & $16.5$ \\
$3.50$  &  $81$ & $55.0\%$ & $-0.964$ & $[-0.977, -0.952]$ & $0.0063$ & $0.0061$ & $17.2$ \\
$\mathbf{4.00}$ &  $87$ & $51.7\%$ & $\mathbf{-1.022}$ & $\mathbf{[-1.032, -1.012]}$ & $0.0049$ & $0.0050$ & $19.5$ \\
$\mathbf{4.50}$ &  $90$ & $50.0\%$ & $\mathbf{-1.033}$ & $\mathbf{[-1.041, -1.024]}$ & $0.0042$ & $0.0044$ & $21.6$ \\
$\mathbf{5.00}$ &  $94$ & $47.8\%$ & $\mathbf{-1.011}$ & $\mathbf{[-1.018, -1.003]}$ & $0.0037$ & $0.0037$ & $23.7$ \\
$\mathbf{5.50}$ & $103$ & $42.8\%$ & $\mathbf{-0.990}$ & $\mathbf{[-0.996, -0.984]}$ & $0.0032$ & $0.0032$ & $24.8$ \\
$\mathbf{6.00}$ & $108$ & $40.0\%$ & $\mathbf{-0.962}$ & $\mathbf{[-0.968, -0.956]}$ & $0.0031$ & $0.0030$ & $25.2$ \\
$7.00$  & $118$ & $34.4\%$ & $-0.880$ & $[-0.886, -0.875]$ & $0.0032$ & $0.0028$ & $25.0$ \\
$\mathbf{8.00}$ & $121$ & $32.8\%$ & $-0.861$ & $[-0.866, -0.856]$ & $\mathbf{0.0030}$ & $0.0026$ & $25.5$ \\
$10.00$ & $130$ & $27.8\%$ & $-0.773$ & $[-0.778, -0.768]$ & $0.0034$ & $0.0026$ & $23.7$ \\
$12.00$ & $132$ & $26.7\%$ & $-0.699$ & $[-0.705, -0.693]$ & $0.0044$ & $0.0030$ & $22.3$ \\
$15.00$ & $139$ & $22.8\%$ & $-0.605$ & $[-0.611, -0.599]$ & $0.0048$ & $0.0029$ & $22.9$ \\
\bottomrule
\end{tabular}
\end{table}

\paragraph{Sensitivity to the abstain profile.}
The mid-b canonical reliability-filtered readout used in the main text is one
member of a post-hoc readout audit computed on the same trained models,
all on $t\in[0.05,0.80]$ with the same pooled OLS and paired-bootstrap
procedure. The raw-positive readout applies no reliability gate beyond
finite positive excess risk ($R>0$); the other readouts apply increasingly
conservative SNR and $R/\AV$ gates. Table~\ref{tab:m2-dense-d4-sensitivity}
summarizes the retained count and plateau signed-slope range for the four
reviewer-facing settings used in the main text. The raw-positive fit keeps
$83.3\%$ of full-ladder candidate cells under a less guarded readout, but
its plateau range $[-1.028,-0.969]$ still tracks the signed-slope form of the
parametric scale. The canonical
mid-b readout keeps $54.6\%$ of cells and gives
$[-1.033,-0.962]$ on the same five plateau rows. The very-strict readout
keeps fewer cells and overshoots mildly on the high end of the plateau,
but it does not reverse the plateau-level conclusion.

\begin{table}[t]
\centering
\scriptsize
\setlength{\tabcolsep}{3pt}
\caption{\textbf{Filter-robust post-hoc reanalysis on the same trained
$d{=}4$ dense-sweep models.} Raw-positive uses all finite positive
excess-risk cells; the other rows apply increasingly conservative
reliability gates. Retained counts are over the full $18$-row ladder
($3240$ candidate cells). The plateau summary is computed over
$\Delta/\sigma\in\{4.0,4.5,5.0,5.5,6.0\}$. The canonical / mid-b row
uses $\mathrm{SNR}\ge4$, $R\ge0.016\,\AV$; the very-strict stress row
uses $\mathrm{SNR}\ge8$, $R\ge0.030\,\AV$.}
\label{tab:m2-dense-d4-sensitivity}
\begin{tabular*}{\linewidth}{@{\extracolsep{\fill}}lcccccc@{}}
\toprule
Readout & SNR & $R$ gate & retained & plateau $s$ & mean CV & $s_{15}$ \\
\midrule
raw-positive & $\ge0$ & $R>0$ & $2700/3240$ ($83.3\%$) & $[-1.028,-0.969]$ & $0.0033$ & $-0.604$ \\
loose & $\ge3$ & $R\ge0.010\,\AV$ & $2118/3240$ ($65.4\%$) & $[-0.986,-0.955]$ & $0.0035$ & $-0.605$ \\
canonical / mid-b & $\ge4$ & $R\ge0.016\,\AV$ & $1768/3240$ ($54.6\%$) & $[-1.033,-0.962]$ & $0.0038$ & $-0.605$ \\
very-strict & $\ge8$ & $R\ge0.030\,\AV$ & $1234/3240$ ($38.1\%$) & $[-1.145,-0.982]$ & $0.0057$ & $-0.598$ \\
\bottomrule
\end{tabular*}
\end{table}

To separate the two gates, we additionally scan a decoupled threshold grid
without retraining: $\mathrm{SNR}\in\{0,1,2,3,4,5,6,8\}$ crossed with
$R/\AV\in\{0,0.005,0.010,0.016,0.020,0.030\}$. This audit is not a
threshold-free estimator; it asks whether the plateau conclusion is continuous
over a reasonable readout neighborhood rather than isolated at the mid-b point.
Among threshold pairs retaining at least $40\%$ of full-ladder cells, $40/40$
put all five plateau signed slopes in the parametric band $[-1.10,-0.90]$
(Table~\ref{tab:m2-dense-d4-threshold-grid}). The only misses in the displayed
grid occur in the most conservative $R/\AV=0.030$ column, which retains only
between $38.1\%$ and $39.1\%$ of cells and mildly overshoots the lower edge of the band.

\begin{table}[H]
\centering
\scriptsize
\setlength{\tabcolsep}{3pt}
\caption{\textbf{Decoupled threshold-grid audit for the dense $d{=}4$
readout.} Each cell reports the plateau range
$[\min s,\max s]$ over
$\Delta/\sigma\in\{4.0,4.5,5.0,5.5,6.0\}$.}
\label{tab:m2-dense-d4-threshold-grid}
\resizebox{\linewidth}{!}{%
\begin{tabular}{@{}ccccccc@{}}
\toprule
SNR gate \textbackslash\ $R/\AV$ gate & $0.000$ & $0.005$ & $0.010$ & $0.016$ & $0.020$ & $0.030$ \\
\midrule
$0$ & $[-1.028,-0.969]$ & $[-1.004,-0.961]$ & $[-0.986,-0.955]$ & $[-1.033,-0.962]$ & $[-1.069,-0.981]$ & $[-1.120,-0.982]$ \\
$1$ & $[-1.028,-0.969]$ & $[-1.004,-0.961]$ & $[-0.986,-0.955]$ & $[-1.033,-0.962]$ & $[-1.069,-0.981]$ & $[-1.120,-0.982]$ \\
$2$ & $[-1.028,-0.969]$ & $[-1.004,-0.961]$ & $[-0.986,-0.955]$ & $[-1.033,-0.962]$ & $[-1.069,-0.981]$ & $[-1.120,-0.982]$ \\
$3$ & $[-1.028,-0.969]$ & $[-1.004,-0.961]$ & $[-0.986,-0.955]$ & $[-1.033,-0.962]$ & $[-1.069,-0.981]$ & $[-1.120,-0.982]$ \\
$4$ & $[-1.026,-0.967]$ & $[-1.004,-0.961]$ & $[-0.986,-0.955]$ & $[-1.033,-0.962]$ & $[-1.069,-0.981]$ & $[-1.120,-0.982]$ \\
$5$ & $[-1.019,-0.966]$ & $[-1.004,-0.961]$ & $[-0.986,-0.955]$ & $[-1.033,-0.962]$ & $[-1.069,-0.981]$ & $[-1.120,-0.982]$ \\
$6$ & $[-1.020,-0.966]$ & $[-1.005,-0.963]$ & $[-0.987,-0.955]$ & $[-1.033,-0.962]$ & $[-1.069,-0.981]$ & $[-1.120,-0.982]$ \\
$8$ & $[-1.024,-0.979]$ & $[-1.013,-0.977]$ & $[-0.995,-0.971]$ & $[-1.035,-0.975]$ & $[-1.080,-0.993]$ & $[-1.145,-0.982]$ \\
\bottomrule
\end{tabular}}
\end{table}

\subsection{Normalization sanity audit}
\label{app:m4-norm-sanity}

We test whether the analytic $\AV(t)$ floor is necessary by replacing
it with three alternatives while keeping the trained MLPs fixed. The
audit runs on auxiliary $d=2$ phase rows (separated, boundary, failure)
plus an overlap-heavy $d=2$ row at $\Delta/\sigma=1.41$. The full
$\AV$ floor is the canonical normalization in the main controlled
readout; this appendix audit records how the readout changes when the
between-component term is removed, when no normalization is used, or
when a large-$N$ empirical reference is substituted.

\begin{table}[t]
  \centering
  \small
  \setlength{\tabcolsep}{6pt}
  \caption{\textbf{Normalization sanity.} CV of $R(t;N)$ across time
    under four floor choices on auxiliary $d=2$ phase rows. Full $\AV$
    is canonical; within-only drops the between-component term; flat
    is no normalization; large-$N$ ref is a $50000$-sample MLP. Full
    $\AV$ is tightest except on the overlap row, where the crossing is
    a partial cancellation in $R$, not collapse.}
  \label{tab:m4-norm}
  \begin{tabular*}{0.8\linewidth}{@{\extracolsep{\fill}}lcccc@{}}
    \toprule
    Row ($\Delta/\sigma$) & Full $\AV$ & Within & Flat & Large-$N$ \\
    \midrule
    separated ($8.00$)   & $\mathbf{0.199}$ & $1.010$ & $0.676$ & $1.216$ \\
    boundary  ($3.00$)   & $\mathbf{0.105}$ & $1.285$ & $0.558$ & $0.536$ \\
    failure   ($2.00$)   & $\mathbf{0.277}$ & $1.282$ & $0.489$ & $0.771$ \\
    overlap   ($1.41$)   & $\mathbf{0.300}$ & $0.940$ & $0.808$ & $0.250$ \\
    \bottomrule
  \end{tabular*}
\end{table}

\subsection{\texorpdfstring{Proxy-floor breakdown at $d{=}2$ separated}{Proxy-floor breakdown at d=2 separated}}
\label{app:m5-d2-limitation}

The Section~\ref{sec:proxy-bridge} proxy-floor audit identifies $d{=}2$
separated as the single controlled row on which
$\hat R := L_N - L_{\mathrm{largeN}}$ departs from the exact $R$
ranking. We document the audit setup, the spread across reseed
pools, the two-factor mechanism, and the empirical calibration of
the boundary $r_F = 2$.

\begin{table}[htbp]
  \centering
  \small
  \setlength{\tabcolsep}{4pt}
  \caption{\textbf{Proxy-floor calibration.} Rank correlation
    $\rho_S=\rho_{\mathrm{Spearman}}(\hat R,R)$ and factor-product
    diagnostic $r_F$ on guard-passing calibration rows. The isolated
    high-$r_F$ $d{=}2$ separated breakdown is analyzed below.}
  \label{tab:m5-proxy}
  \begin{tabular*}{0.8\linewidth}{@{\extracolsep{\fill}}lccc@{}}
    \toprule
    Row & $\Delta/\sigma$ & $\rho_S$ & $r_F$ \\
    \midrule
    $d{=}4$ sep.             & $8.00$ & $0.997$ & $0.83$ \\
    $d{=}8$ sep.             & $6.00$ & $0.998$ & $0.71$ \\
    $d{=}8$ strict           & $8.00$ & $1.000$ & $0.47$ \\
    $K$-sweep ($d{=}8$)      & $6.00$ & $[0.959,1.000]$ & $[0.47,1.38]$ \\
    $d{=}2$ aux. ($3$)       & $[1.41,3.00]$ & $[0.997,0.998]$ & $[0.81,0.91]$ \\
    \bottomrule
  \end{tabular*}
\end{table}

\paragraph{Reseed-pool spread at $d{=}2$ separated.}
Table~\ref{tab:m5-d2-verdicts} reports the audit on five
independent instances: the canonical $5$-seed pool, three disjoint
$5$-seed pools, and a higher-power $20$-seed pool. The rank
correlation $\rho_{\mathrm{Spearman}}(\hat R, R)$ spreads across
$[0.471, 0.868]$ on the five-seed pools and lands at $0.662$ on
the twenty-seed pool. The canonical pool's apparent agreement on
the bin-level region label is a knife-edge artifact: the exact-$R$
mass difference between the second and third regions is only
$0.16\%$ of total mass on every reseed, so four of the five pools
disagree on the mid-vs-high label under independent resampling.

\begin{table}[htbp]
  \centering
  \small
  \setlength{\tabcolsep}{6pt}
  \caption{\textbf{Reseed-pool spread at $d{=}2$ separated.} Rank
    correlation $\rho_{\mathrm{Spearman}}(\hat R, R)$, the
    fraction of region labels (low/mid/high mass) reproduced by
    the proxy, and the strict-tie ranking gap (exact-$R$ mass
    difference between the second and third regions) across five independent
    reseed instances. The small strict-tie gap explains why independent
    resampling changes the mid-vs-high label on most pools.}
  \label{tab:m5-d2-verdicts}
  \begin{tabular*}{0.8\linewidth}{@{\extracolsep{\fill}}lcccc@{}}
    \toprule
    Instance & Seeds & $\rho_{\mathrm{Spearman}}$ & Region-label match & Strict-tie gap \\
    \midrule
    canonical pool   & $5$  & $0.552$ & $3/3$ & $0.16\%$ \\
    pool $0$         & $5$  & $0.471$ & $3/3$ & $0.16\%$ \\
    pool $1$         & $5$  & $0.868$ & $3/3$ & $0.16\%$ \\
    pool $2$         & $5$  & $0.704$ & $2/3$ & $0.16\%$ \\
    extended pool    & $20$ & $0.662$ & $2/3$ & $0.16\%$ \\
    \bottomrule
  \end{tabular*}
\end{table}

\paragraph{Factor-product mechanism.}
The breakdown is the product of two near-orthogonal factors. Factor
$A$ is a proxy-floor sign flip: the large-$N$ reference
$L_{\mathrm{largeN}}(t)$ drops below the analytic floor $\AV(t)$ at
three high-$t$ bins, with
$\max|L_{\mathrm{largeN}} - \AV| = 0.092$ on this row. Factor $B$
is profile flatness: the standard deviation of $R(t)$ over the
mid-to-high audit window $t \in [0.4, 0.8]$ is only $0.034$
(coefficient of variation $10\%$), with $R$ values clustered in
$[0.28, 0.38]$. The diagnostic
$r_F := \max|L_{\mathrm{largeN}} - \AV| /
\mathrm{std}(R \mid t \in [0.4, 0.8])$ equals $2.70$ on this row,
exceeding the empirical boundary $2$ calibrated on the
seventeen-cell sweep. Counterfactual manipulations confirm the
mechanism: clamping the floor at $L_{\mathrm{largeN}} \ge \AV$
recovers $\rho_{\mathrm{Spearman}}$ to $0.70$, trimming the audit
window to $t \le 0.7$ recovers it to $0.84$, and lifting the
geometry to $d \ge 4$ removes both factors simultaneously
($d{=}4$ separated gives $r_F = 0.83$ with zero sign-flip bins).

\paragraph{Empirical boundary, not a universal constant.}
The threshold $r_F = 2$ is calibrated on the present
seventeen-cell controlled dataset; rows beyond this audit may push
it. We report it as an empirical separator rather than a derived
constant. The more conservative cut $r_F = 1$ would split off
three intermediate $K$-sweep cells at $K \in \{2, 64\}$ that
retain $\rho_{\mathrm{Spearman}} \in [0.959, 0.994]$ together with
the $d{=}2$ separated breakdown, while the calibrated cut at
$r_F = 2$ isolates only the $d{=}2$ separated regime.

\paragraph{Cross-$d$ mechanism breakdown.}
The factor product is monotone in $d$
(Table~\ref{tab:m5-mechanism}). Lifting from $d{=}2$ to $d{=}4$
removes both factors at once: the proxy floor stops dipping below
$\AV$ (zero sign-flip bins from $d{=}4$ onward), and the
mid-to-high $t$ standard deviation of $R$ grows roughly as the
ambient dimension. The result is a sharp dimension transition:
$r_F$ drops from $2.70$ at $d{=}2$ to $0.83$ at $d{=}4$ and to
$0.71$ at $d{=}8$, all on otherwise matched controlled
configurations. Intermediate $d$ values are not necessary to
establish the transition; $d{=}4$ already crosses the empirical
boundary by a wide margin.

\begin{table}[htbp]
  \centering
  \small
  \setlength{\tabcolsep}{7pt}
  \begin{tabular}{@{}lccccc@{}}
    \toprule
    row & \shortstack{flip bins} & $G_F$ & \shortstack{std$(R)$ mid/high $t$} & $r_F$ & $r_F^{\mathrm{obs}}$ \\
    \midrule
    $d{=}2$ separated, $\Delta/\sigma=6$    & $3$ & $0.092$ & $0.034$ & $2.70$ & $2.79$ \\
    $d{=}4$ separated, $\Delta/\sigma=8$    & $0$ & $0.160$ & $0.193$ & $0.83$ & $0.42$ \\
    $d{=}8$ separated, $\Delta/\sigma=6$    & $0$ & $0.185$ & $0.260$ & $0.71$ & $0.20$ \\
    $d{=}8$ separated, $\Delta/\sigma=8$    & $0$ & $0.221$ & $0.467$ & $0.47$ & $0.17$ \\
    \bottomrule
  \end{tabular}
  \caption{\textbf{Cross-$d$ factor-product breakdown on separated rows.}
    Here $G_F=\max|L_{\mathrm{largeN}}-\AV|$ and mid/high $t$ denotes
    $t \in [0.4,0.8]$.
    The $d{=}2$ separated row is the only configuration with a
    non-zero sign-flip count and a small mid-to-high $t$ standard
    deviation of $R$. From $d{=}4$ onward the sign flip vanishes
    and the standard deviation grows, so $r_F$ drops well below the empirical
    boundary $2$. Both $d{=}8$ rows fall inside the rank-stable regime.
    The rightmost column reports the observable counterpart
    $r_F^{\mathrm{obs}}$ defined in the next paragraph.}
  \label{tab:m5-mechanism}
\end{table}

\paragraph{Diagnostic boundary calibration.}
Table~\ref{tab:m5-ratio-classes} bins the seventeen audited cells
by $r_F$ and reports the empirical $\rho_{\mathrm{Spearman}}$
range in each band. The bins show that the rank correlation
tracks $r_F$ monotonically: all twelve cells with $r_F < 1$ and
all three cells with $r_F \in [1, 2)$ retain
$\rho_{\mathrm{Spearman}} \ge 0.959$, while the two cells with
$r_F \ge 2$ (the canonical and twenty-seed pools of $d{=}2$
separated) drop to $\rho_{\mathrm{Spearman}} \le 0.662$. Cell
composition of each band is given in the main text and is not
repeated here.

\begin{table}[htbp]
  \centering
  \small
  \setlength{\tabcolsep}{4pt}
  \caption{\textbf{Diagnostic boundary calibration.} Distribution
    of the seventeen controlled cells across $r_F$ bands. The
    rank correlation tracks $r_F$ monotonically: cells in the
    lower two bands retain $\rho_{\mathrm{Spearman}} \ge 0.959$,
    while the two cells in the upper band drop to
    $\rho_{\mathrm{Spearman}} \le 0.662$. Cell composition of
    each band is described in the main text.}
  \label{tab:m5-ratio-classes}
  \begin{tabular*}{0.5\linewidth}{@{\extracolsep{\fill}}lcc@{}}
    \toprule
    $r_F$ band & cells & $\rho_{\mathrm{Spearman}}$ range \\
    \midrule
    $r_F < 1.0$            & $12$ & $[0.991, 1.000]$ \\
    $r_F \in [1.0, 2.0)$   & $3$  & $[0.959, 0.994]$ \\
    $r_F \ge 2.0$          & $2$  & $[0.552, 0.662]$ \\
    \bottomrule
  \end{tabular*}
\end{table}

\paragraph{Observable counterpart of the factor ratio.}
The numerator of $r_F$ uses $|L_{\mathrm{largeN}} - \AV|$, which
depends on the analytic floor. An observable counterpart that requires
only the $N$-ladder raw data is
\[
  r_F^{\mathrm{obs}} := \frac{\max_t |L_{\mathrm{ref},a}(t) - L_{\mathrm{ref},b}(t)|}
                              {\mathrm{std}_t\!\left(\hat R_{\mathrm{mean}\text{-}N} \mid t \in [0.4, 0.8]\right)},
\]
where $L_{\mathrm{ref},a}$ and $L_{\mathrm{ref},b}$ are independent
half-fold means over training seeds at the ladder-end $N$, and
$\hat R_{\mathrm{mean}\text{-}N}$ is the mean across $N$ of the
lower-$N$ excess profiles measured against
$\tfrac{1}{2}(L_{\mathrm{ref},a}+L_{\mathrm{ref},b})$. On the same
seventeen-cell label set, $r_F^{\mathrm{obs}} \le 1.234$ reproduces the
analytic verdict with accuracy $1.000$ ($15$ pass, $2$ fail; Youden's
$J = 1.000$). The observable ratio tracks the analytic ratio
monotonically (rightmost column of Table~\ref{tab:m5-mechanism}), and
the two boundaries $r_F^{\mathrm{obs}} = 1.234$ and $r_F = 2$ classify
every audited row identically.

\subsection{Three-family controlled heterogeneity panel}
\label{app:m7-heterogeneity}

The controlled VAP framework predicts that distinct mixture
geometries should produce distinct $R(t;N)$ profile shapes. To
illustrate this qualitatively, we report a three-family panel of
equal-weight Gaussian mixtures in $d{=}2$ chosen to span
geometrically extreme cases: a well-separated $K{=}4$ ring with
$\Delta/\sigma \in [8.0, 11.3]$, a high-entropy $K{=}8$ ring with
$\Delta/\sigma \in [2.6, 6.8]$, and a partial-merge $K{=}4$
configuration with cross-pair $\Delta/\sigma = 30$ and within-pair
$\Delta/\sigma = 3$. Each family is trained at the same
hyperparameters ($t \in [0.05, 0.80]$ partitioned into $40$ bins,
$10$ data seeds with $3$ initialization replicates per cell, MLP
hidden width $128$, $5000$ to $8000$ training steps).

At anchor $N = 1000$, the three families produce visibly distinct
mean $R(t;N)$ peaks at $t \approx \{0.43, 0.34, 0.78\}$
(Table~\ref{tab:m7-families}). The minimum pairwise normalized
$L^2$ distance between the three profile shapes is $0.2059$
(well-separated versus partial-merge), exceeding a descriptive
separability scale of $0.15$; the other two pairwise distances are
above $0.40$. We treat this as a qualitative separability finding
rather than a hypothesis test, and use the panel only to
illustrate that distinct controlled geometries induce distinct
$R(t;N)$ shapes.

\begin{table}[htbp]
  \centering
  \small
  \setlength{\tabcolsep}{6pt}
  \caption{\textbf{Three-family controlled heterogeneity panel.}
    Mean $R(t; N{=}1000)$ profile statistics across three
    Gaussian-mixture geometries in $d{=}2$. The minimum pairwise normalized
    $L^2$ distance between profile shapes is $0.2059$, between the
    well-separated and partial-merge families. The partial-merge family was
    selected to stabilize the visualization; this selection does not change
    the qualitative finding.}
  \label{tab:m7-families}
  \begin{tabular*}{0.8\linewidth}{@{\extracolsep{\fill}}lcccc@{}}
    \toprule
    family & geometry & $\Delta/\sigma$ range & peak $t$ & $R$ peak \\
    \midrule
    well-separated & $K{=}4$ ring   & $[8.0, 11.3]$ & $0.435$ & $0.136$ \\
    high-entropy   & $K{=}8$ ring   & $[2.6, 6.8]$  & $0.338$ & $0.052$ \\
    partial-merge  & $K{=}4$ paired & $[3.0, 30.0]$ & $0.781$ & $0.271$ \\
    \bottomrule
  \end{tabular*}
\end{table}

The partial-merge family uses paired clusters with very large
between-pair separation ($\Delta/\sigma = 30$) and modest
within-pair separation ($\Delta/\sigma = 3$); the resulting
mid-$t$ shoulder and high-$t$ peak ($t \approx 0.78$) sit one bin
from the readout boundary $t_{\max} = 0.80$, and we keep
$t_{\max} = 0.80$ for comparability with the other two families
rather than extending the window. The high-entropy family at
$K{=}8$ shifts the peak earlier ($t \approx 0.34$) because more
mixture components reach near-uniform posterior at smaller $t$,
while the well-separated family at $K{=}4$ sits between the two
($t \approx 0.43$).

\paragraph{Why we do not report a sharper readout.}
An earlier version of this panel used a single region-based
coefficient-of-variation scalar as a separator. That scalar took
similar values on the well-separated and partial-merge geometries
despite their visibly different $R(t;N)$ shapes, so we use the
pairwise-distance descriptor instead. The
descriptor is descriptive rather than a hypothesis test; its role
is to support the claim that distinct controlled geometries
induce distinct $R(t;N)$ shapes.

\subsection{CIFAR-100 DINOv2 feature-flow materials}
\label{app:cifar}

This appendix records the feature extraction, training, deployment, and
metric definitions supporting Section~\ref{sec:cifar}. The experiment uses
five paired training seeds throughout; reported confidence intervals and
seed-level signs are computed over those paired runs.

\paragraph{Representation and training protocol.}\label{app:cifar-data}
Images from the official CIFAR-100 train and test splits are passed through
frozen DINOv2 ViT-S/14. We retain the final-LayerNorm CLS representation
with dimension $384$, then apply train-split centering and a global RMS
calibration. The calibration uses all $50{,}000$ official training rows and
does not access test features. Each target seed uses a fine-balanced subset of
$25$ examples per class, or $2{,}500$ labeled examples; the five reference
models use the full $50{,}000$ labeled training rows. The same transform is
used for every arm and for the held-out reference features. There is no PCA,
whitening, learned projection, or pixel decoder. The flow-matching model is
conditioned on the fine label and uses a $432\mathbin{\to}512\mathbin{\to}512
\mathbin{\to}512\mathbin{\to}512\mathbin{\to}384$ MLP, Adam, learning rate
$3\times10^{-4}$, and batch size $512$. Uniform training samples $t$ from the
prescribed interval; the VAP-guided arm samples from the frozen proxy with
$q_4(t)\propto\widehat R(t)^4$ on $64$ categorical bins and no importance
correction. Evaluation uses a uniformly weighted grid of $25$ time points.

\paragraph{Training readouts.}\label{app:cifar-training}
The primary training comparison evaluates both arms at the common step
$11{,}500$ on the $25$-point uniform-time grid. The VAP-guided arm lowers
velocity loss by $6.78\%$ relative to the paired uniform arm, with all
$5/5$ paired seeds favoring the guided arm and a paired $95\%$ interval of
$[-7.11\%, -6.44\%]$. The seed-mean own-best comparison gives a $6.49\%$
reduction; this secondary readout is retained to distinguish a common-step
comparison from a per-method trajectory summary.

\paragraph{Deployment checkpoint selection.}\label{app:cifar-checkpoints}
For closed-loop evaluation, each method is materialized once at the mode of
its five per-seed own-best steps. This yields step $8{,}500$ for the uniform
baseline and step $10{,}500$ for the VAP-guided arm. The selected checkpoints
are then held fixed for all paired ODE rollouts. The VAP-guided arm was chosen
after inspecting the parent training trajectories, so this is a descriptive
post-selection comparison rather than an arm-independent preregistered test.
Each rollout starts from paired Gaussian noise and integrates the learned field
from $t=1$ to $t=0$ with fixed-step Heun, $100$ integration steps and $200$ network
function evaluations. The state is float32 and the integration grid is
float64; the literal primary policy calls the network on the full path,
including $t\in(0.95,1]$ and $t\in[0,0.05)$. Each method contributes $6{,}000$
generated features, $60$ per fine class, and is compared with a disjoint $6{,}000$-
feature test reference bank, also $60$ per class. The checkpoint rule is a
trajectory-selection convention; it is not an additional training run.

\paragraph{Closed-loop metrics.}\label{app:cifar-deployment}
The generated $384$-dimensional features are compared with held-out CIFAR-100
DINOv2 features. Table~\ref{tab:cifar-deployment-metrics} reports method
means over the five paired seeds. Feature FD is a global Gaussian
Fr\'echet distance in feature space. Precision, density, and coverage use
fixed $k$-nearest-neighbor radii ($k=3$ for precision and $k=5$ for density
and coverage). Fine-label accuracy and cross-entropy are computed from a
classifier evaluated on generated features. Macro conditional KID is the
frozen per-class kernel criterion. Improved recall and effective covariance
rank are reported as diversity diagnostics.

\begin{table}[!htb]
  \centering
  \small
  \setlength{\tabcolsep}{4pt}
  \caption{\textbf{Closed-loop CIFAR-100 DINOv2 metrics.} Means are over
    five paired training seeds. Feature FD is not image FID.}
  \label{tab:cifar-deployment-metrics}
  \begin{tabular}{@{}lccc@{}}
    \toprule
    Metric & Uniform & VAP-guided & Paired change \\
    \midrule
    Global feature FD $\downarrow$ & $33.333$ & $25.476$ & $-7.858$ \\
    Precision ($k=3$) $\uparrow$ & $0.151$ & $0.690$ & $+0.539$ \\
    Density ($k=5$) $\uparrow$ & $0.122$ & $1.226$ & $+1.104$ \\
    Coverage ($k=5$) $\uparrow$ & $0.234$ & $0.728$ & $+0.494$ \\
    Fine macro accuracy $\uparrow$ & $0.791$ & $0.859$ & $+0.0676$ \\
    Fine cross-entropy $\downarrow$ & $1.094$ & $0.626$ & $-0.468$ \\
    Macro conditional KID $\downarrow$ & $0.1988$ & $0.2062$ & $+0.00744$ \\
    Improved recall ($k=3$) $\uparrow$ & $0.7221$ & $0.2214$ & $-0.5007$ \\
    Effective covariance rank $\uparrow$ & $47.25$ & $40.14$ & $-7.11$ \\
    C2ST macro AUC $\uparrow$ & $0.8511$ & $0.8612$ & $+0.0101$ \\
    \bottomrule
  \end{tabular}
\end{table}

The global FD reduction is approximately $23.6\%$, while macro conditional
KID improves in only one of five paired seeds and has aggregate change in the
unfavorable direction. The C2ST macro AUC increases from $0.8511$ to $0.8612$,
making real and generated features easier to discriminate. The simultaneous
reduction in improved recall and effective covariance rank records the fidelity
and diversity trade-off summarized in the main text. A matched-size real-real
floor ($30$ features per class, $100$ repeats) has cKID mean $-0.000302$ with SD
$0.006423$ and feature FD mean $15.7088$ with SD $0.2298$. Solver-resolution
checks at $25$, $50$, $100$, and $200$ Heun steps preserve the literal cKID direction;
the supported-interior endpoint diagnostic is reported only as a sensitivity
check. The deployment table remains a feature-space result and does not
establish pixel-space image quality.

\paragraph{Reproducibility record.}
The representation and preprocessing specification is frozen before the
reported comparisons: DINOv2 ViT-S/14 final-LayerNorm CLS features use
train-only centering and global-RMS calibration, with the same transform for
all arms and reference features. The supplementary reproducibility materials
retain the paired-seed summaries, metric definitions, solver checks, and
source/reference/target role separation used by this appendix. The released
paper figures are derived from these frozen records and do not introduce an
additional representation or evaluation split.

\begin{figure}[t]
  \centering
  \includegraphics[width=0.98\linewidth]{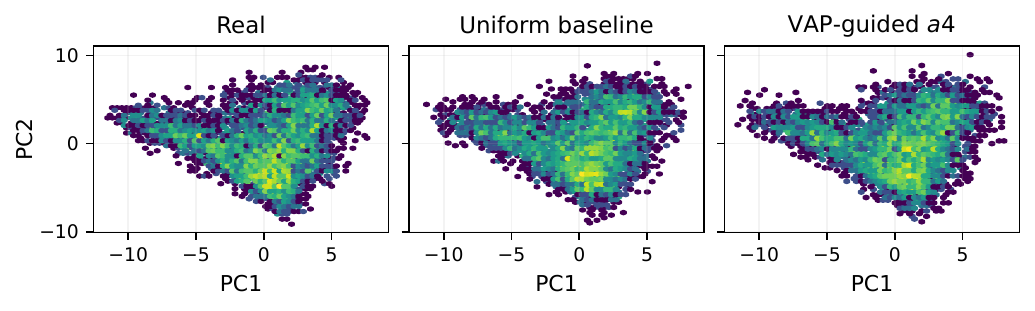}
  \caption{\textbf{PCA projection for visualization of DINOv2 features.}
    The panels compare the held-out real feature bank with features generated
    by the uniform and VAP-guided $\alpha=4$ samplers. This is a feature-space
    visualization only; PCA is not the representation used to train the flow.}
  \label{fig:cifar-dinov2-pca-density}
\end{figure}

\subsection{Cross-cell d-axis ladder of the time-sampler stress test}
\label{app:m9e-v5}

To check whether the sampler-side prescription of
Section~\ref{sec:envelope} is specific to the $(K{=}32, d{=}16)$ cell
of the main text, we replicate the seven-arm protocol on two
neighboring cells along the dimension axis: $(K{=}24, d{=}12)$ and
$(K{=}40, d{=}20)$, with $N=512$ held fixed. When $N$ is not exactly
divisible by $K$, demonstrations per component default to a
deterministic almost-balanced split (the first $N \bmod K$ components
receive one extra demonstration, so per-component counts differ by at
most one). All other configuration knobs (sigma, gap, spread, hidden
width, learning rate, batch size, training steps, evaluation grid,
fine-time bins, action budget, noise samples, source seeds
$\{20,\ldots,29\}$, paired test seeds $\{0,\ldots,9\}$, source action
key, paired-test action key, ESS and mid-time guard rails) match
Section~\ref{sec:envelope} exactly. Plateau readings of
$L_N(\mathrm{ts}{=}4000)/\AV$ are $1.0276$ at $K{=}24$ (closer to
the analytic floor than the canonical cell) and $1.0360$ at $K{=}40$
(essentially matching the canonical $1.0370$).

\begin{table}[htbp]
  \centering
  \small
  \setlength{\tabcolsep}{4pt}
  \caption{\textbf{Controlled time-sampler stress test.} Plateau paired
    statistics over $10$ seeds at training step $4000$. The metric
    $\Delta_{\mathrm{red}}$ is the percent change in t-averaged reducible
    loss relative to the baseline; every gap arm has negative paired
    seed-level deltas for all seeds.}
  \label{tab:m9e-v4}
  \begin{tabular}{@{}llccc@{}}
    \toprule
    Arm & Sampler & Mean $\Delta_{\mathrm{red}}\,(\%)$ & SE\,(\%) & $p_{\mathrm{perm}}$ \\
    \midrule
    \texttt{gap\_mix\_025} & mixture ($\lambda=0.25$) & $-9.32$ & $0.98$ & $0.0015$ \\
    \texttt{gap\_mix\_050} & mixture ($\lambda=0.50$) & $-13.95$ & $1.19$ & $0.0028$ \\
    \texttt{gap\_mix\_075} & mixture ($\lambda=0.75$) & $-17.33$ & $1.33$ & $0.0018$ \\
    \texttt{gap\_pure\_a1} & pure ($\alpha=1$) & $-18.84$ & $1.37$ & $0.0019$ \\
    \texttt{gap\_pure\_a2} & pure ($\alpha=2$) & $\mathbf{-20.21}$ & $1.69$ & $0.0019$ \\
    \texttt{gap\_pure\_a4} & pure ($\alpha=4$) & $-19.15$ & $2.10$ & $0.0017$ \\
    \texttt{gap\_pure\_a8} & pure ($\alpha=8$) & $-17.92$ & $2.63$ & $0.0023$ \\
    \bottomrule
  \end{tabular}
\end{table}

\begin{table}[htbp]
  \centering
  \small
  \setlength{\tabcolsep}{4pt}
  \caption{\textbf{Cross-cell d-axis ladder of the time-sampler
    stress test: paired $\Delta_{\mathrm{red}}\,(\%)$ at training
    step $4000$ over $10$ paired seeds.} The headline metric is the
    reducible-loss percent change as defined in
    Table~\ref{tab:m9e-v4}. The middle column is the canonical
    $(K{=}32, d{=}16)$ cell of Section~\ref{sec:envelope}. Mixture
    arms and pure exponents $\alpha\in\{1,2\}$ reproduce the
    canonical direction across all three cells with sign distribution
    $10/10$ and $p_{\mathrm{perm}}<0.005$. Pure $\alpha=4$ remains
    negative but weakens at the largest cell
    ($9/10$, $p_{\mathrm{perm}}\approx0.008$), while the most
    aggressive pure sampler ($\alpha=8$) reverses sign in the late
    plateau only there.}
  \label{tab:m9e-v5-arms}
  \begin{tabular*}{0.95\linewidth}{@{\extracolsep{\fill}}lccc@{}}
    \toprule
    Arm & $K{=}24, d{=}12$ & $K{=}32, d{=}16$ (canonical) & $K{=}40, d{=}20$ \\
    \midrule
    \texttt{gap\_mix\_025} & $-7.76$  & $-9.32$  & $-10.22$ \\
    \texttt{gap\_mix\_050} & $-13.06$ & $-13.95$ & $-16.17$ \\
    \texttt{gap\_mix\_075} & $-16.38$ & $-17.33$ & $-19.95$ \\
    \texttt{gap\_pure\_a1} & $-17.84$ & $-18.84$ & $-21.39$ \\
    \texttt{gap\_pure\_a2} & $-20.56$ & $-20.21$ & $-21.56$ \\
    \texttt{gap\_pure\_a4} & $-19.76$ & $-19.15$ & $-11.45$ \\
    \texttt{gap\_pure\_a8} & $-19.25$ & $-17.92$ & $\mathbf{+11.93}$ \\
    \bottomrule
  \end{tabular*}
\end{table}

\begin{figure}[htbp]
  \centering
  \safeincludegraphics[width=\linewidth]{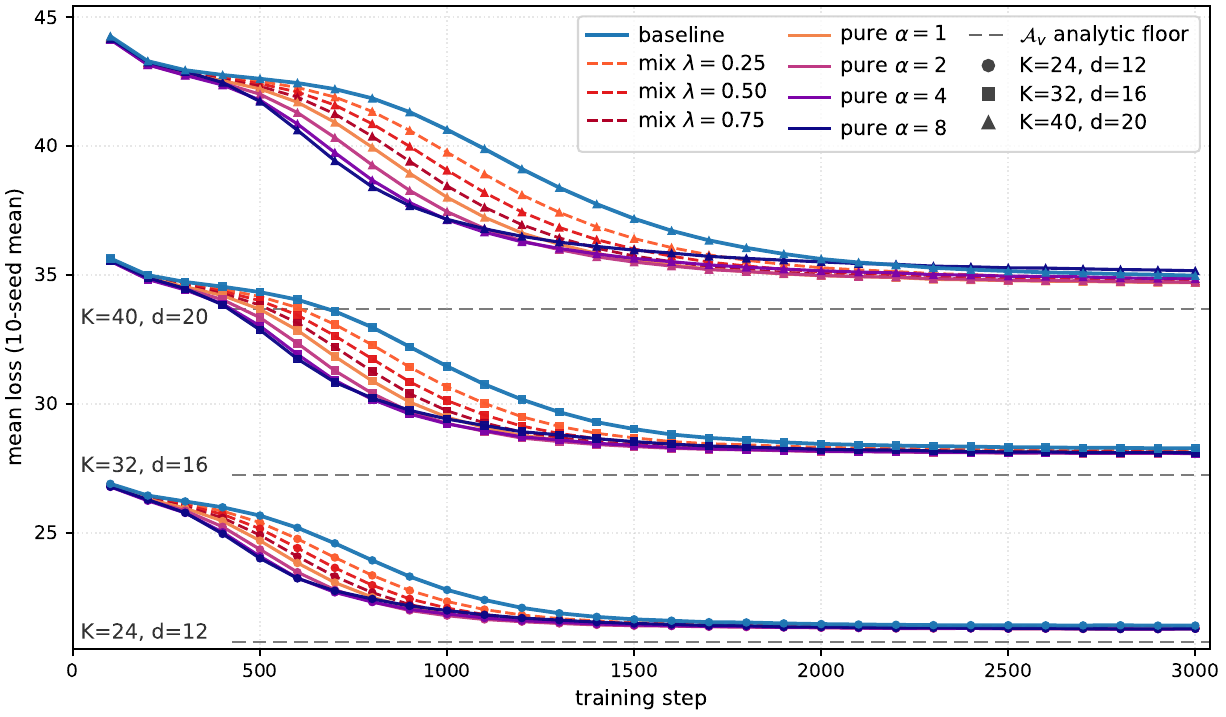}
  \caption{\textbf{Cross-cell d-axis ladder of the time-sampler stress
    test.} Three panels share a common y-range. The middle panel
    reproduces Figure~\ref{fig:m9e-v4} for reference. The
    panel ($K{=}24, d{=}12$) keeps every tested arm strictly below
    baseline at every checkpoint, including $\alpha=8$. The
    panel ($K{=}40, d{=}20$) shows a late-plateau reversal of
    $\alpha=8$: the curve crosses the baseline near training step
    $2500$ and ends slightly above it. Lower exponents and mixture
    arms remain stable across all three cells.}
  \label{fig:m9e-v5-3cell}
\end{figure}

\paragraph{Cross-cell synthesis.}
Mixture arms and the lower pure exponents
($\alpha\in\{1,2\}$) reproduce the canonical direction across all
three cells with paired sign distributions of $10/10$ and
$p_{\mathrm{perm}}<0.005$. The intermediate $\alpha=4$ stays negative
across all three cells but weakens at $K{=}40$ ($-11.45\%$ versus
$-19.15\%$ canonical). The most aggressive sampler ($\alpha=8$) is
the single failure mode: stable on the smaller and canonical cells,
it plateau-reverses to $+11.93\%$ at $K{=}40$ with sign distribution
$3/10$ and $p_{\mathrm{perm}}=0.095$. We read this as the upper bound
of a usable $\alpha$ envelope shrinking with the mode count $K$,
rather than a global failure of the sampler-side prescription.

\paragraph{Caveats.}
This panel is a single-axis sensitivity analysis on a single task
family (partial-merge GMM); cross-task, cross-architecture, and
cross-learning-rate stability remain open. The $K{=}24$ cell sits
slightly closer to the analytic floor ($L/\AV=1.0276$) than the
canonical and $K{=}40$ cells, so its absolute reducible budget is
smaller; the signal direction is nonetheless preserved across all
seven arms. The $K{=}40$ $\alpha=8$ sign reversal happens in the
late plateau only; the early-training peak is comparable to the
canonical cell. The $\alpha$ envelope upper bound stated here is
descriptive rather than a hypothesis test.

\section{Oracle-to-Blind Bridge Auxiliary Details}
\label{app:bridge-aux}

\subsection{Label-and-action Cram\'er-Rao sanity}
\label{app:m6a-cr-sanity}

\paragraph{Bridge estimator definitions.}
The true-label plug-in $\mathrm{oracle\_ols}$ estimates each component mean
by ordinary least squares (OLS), which here reduces to its labeled sample
mean, and uses the known within-component variance and the true component
label at prediction time. The blind EM-OLS arm fits a spherical Gaussian
mixture with the known component count $K$ to unlabeled training actions
using EM. It plugs the fitted means and mixture weights, together with the
arithmetic mean of the fitted component variances, into the marginal
Gaussian-mixture velocity formula and averages over posterior component
probabilities given $(x_t,t)$. Thus, neither fitting nor velocity prediction
uses true labels. The OLS suffix is an experimental arm name inherited from
the oracle comparison; the blind arm does not perform a separate OLS
velocity regression. Its prediction is closed form conditional on the
iteratively fitted mixture parameters.

The bridge claim of Section~\ref{sec:proxy-bridge} presupposes that
the closed-form label-and-action arm $L_{\mathrm{oracle\_ols}}$ achieves
the analytic floor within$(t) = d \sigma^2 / \sigma_t^2$
asymptotically. We verify this precondition on the same five
controlled rows used for the bridge audit. At $N = 2000$ in the
readout window $t \in [0.10, 0.28]$, the empirical
$L_{\mathrm{oracle\_ols}}$ matches the analytic floor to within
$0.014\%$ to $0.044\%$ (Table~\ref{tab:m6a-floor}); over a wider
$N$ grid, the ratio of the empirical excess
$R_{\mathrm{oracle\_ols}}(t, N) := L_{\mathrm{oracle\_ols}}(t, N) -
\mathrm{within}(t)$ to the theoretical Cram\'er-Rao within-cluster
term $\mathrm{CR}_{\mathrm{within}}(t, N)$ lies in $[0.93, 1.15]$
across all five rows, with cross-$t$ coefficient of variation in
$[0.04, 0.06]$ (Table~\ref{tab:m6a-ratio}). These are
implementation-sanity readouts: they confirm that the closed-form
estimator hits the analytic floor at finite sample size and
follows the Cram\'er-Rao rate, but they are not standalone novel
claims.

The within$(t)$ formula depends only on $(d, \sigma)$ and not on
the cluster centers, so the two $d{=}2$ rows in
Table~\ref{tab:m6a-floor} (a separated configuration and a
partial-merge configuration with the same $(d, \sigma)$) produce
identical $L_{\mathrm{oracle\_ols}}$ profiles by mean-invariance;
this is theory-correct and not data duplication.

\begin{table}[htbp]
  \centering
  \small
  \setlength{\tabcolsep}{6pt}
  \caption{\textbf{\texorpdfstring{$\mathrm{oracle\_ols}$}{oracle_ols} arm versus analytic floor.} Empirical
    $L_{\mathrm{oracle\_ols}}$ at $N = 2000$ versus the analytic
    floor within$(t) = d \sigma^2 / \sigma_t^2$, computed on the
    readout window $t \in [0.10, 0.28]$ with $20$ data seeds and
    eval $12000 \times 32$ Monte-Carlo samples per cell. The two
    $d{=}2$ rows share $(d, \sigma)$ and therefore share
    within$(t)$ by mean-invariance.}
  \label{tab:m6a-floor}
  \begin{tabular*}{0.8\linewidth}{@{\extracolsep{\fill}}lcc@{}}
    \toprule
    row & within$(t)$ at $N{=}2000$ & relative gap \\
    \midrule
    $d{=}2$ separated     & $2.4043$  & $-0.014\%$ \\
    $d{=}2$ partial-merge & $2.4043$  & $-0.014\%$ \\
    $d{=}4$ simplex       & $4.8086$  & $+0.008\%$ \\
    $d{=}8$ simplex       & $9.6171$  & $+0.024\%$ \\
    $d{=}16$ simplex      & $19.2343$ & $+0.044\%$ \\
    \bottomrule
  \end{tabular*}
\end{table}

\begin{table}[htbp]
  \centering
  \small
  \setlength{\tabcolsep}{6pt}
  \caption{\textbf{Cram\'er-Rao rate audit.} Ratio of empirical
    $R_{\mathrm{oracle\_ols}}(t, N)$ to the theoretical Cram\'er-Rao
    within-cluster term $\mathrm{CR}_{\mathrm{within}}(t, N)$,
    averaged over the kept-bin filter $t \ge 0.10$, on the same
    five rows as Table~\ref{tab:m6a-floor}. The cross-$t$
    coefficient of variation reports the within-row variability
    of the rate.}
  \label{tab:m6a-ratio}
  \begin{tabular*}{0.8\linewidth}{@{\extracolsep{\fill}}lccc@{}}
    \toprule
    row & $\mathrm{ratio}_{\mathrm{mean}}$ & CV across $t$ & $N$ grid \\
    \midrule
    $d{=}2$ separated     & $0.962$ & $0.048$ & $\{50, 100, 200, 500\}$ \\
    $d{=}2$ partial-merge & $0.962$ & $0.048$ & $\{50, 100, 200, 500\}$ \\
    $d{=}4$ simplex       & $0.930$ & $0.041$ & $\{50, 100, 200, 500\}$ \\
    $d{=}8$ simplex       & $1.133$ & $0.061$ & $\{50, 100, 200, 500, 1000\}$ \\
    $d{=}16$ simplex      & $1.153$ & $0.053$ & $\{50, 100, 200, 500, 1000, 2000\}$ \\
    \bottomrule
  \end{tabular*}
\end{table}

\subsection{Blind-Attainability Panel}
\label{app:blind-attain-panel}

Theorem~\ref{thm:gmm} states a label-oracle residual-regression benchmark of
order $K\AV(t)/N$. In the separated regime used by this audit,
$\AV(t)=\AVw(t)+\AVb(t)$ with $\AVb(t)$ exponentially small, so the
within-cluster Cram\'er-Rao specialization is $K\AVw(t)/N$. We complement
Section~\ref{sec:proxy-bridge} with a panel
that tests whether the closed-form blind EM-OLS arm tracks this rate,
and whether the neural blind MLP-FM arm reaches the marginal Bayes
floor up to a bounded residual, on the same five controlled rows. The
setup reuses the same five controlled bridge cells:
five rows spanning $d \in \{2, 4, 8, 16\}$ with $K = 4$ and
$\sigma = 0.5$; an $N$ ladder $\{50, 100, 200, 500, 1000, 2000\}$;
twenty seeds per cell; window $t \in [0.10, 0.28]$ pre-registered before
the readout.

For each arm we form the excess
$R_{\mathrm{arm}}(t; N) = L_{\mathrm{arm}}(t; N) - \mathrm{floor}_{\mathrm{arm}}(t)$
with $\mathrm{floor}_{\mathrm{oracle\_ols}} = $ within$(t)$ and
$\mathrm{floor}_{\mathrm{blind}} = $ within$(t) + $ between$(t)$ (the
analytic marginal Bayes asymptote, equal to $\AV(t)$ in
Definition~\ref{def:sigma}). We fit
$\alpha_{\mathrm{pooled}} = -\partial \log \overline{R} / \partial \log N$
on the seed-mean profile, restricted to $N$ where
$\overline{R} > 1.5 \times $ SE (noise-floor guard); $95\%$ confidence
intervals come from $1000$-resample paired-seed bootstrap.

Table~\ref{tab:blind-attain} reports the per-row outcome. The
$\mathrm{oracle\_ols}$ arm matches the band $[0.85, 1.15]$ on the
high-$d$ rows ($d \in \{8, 16\}$) and drifts upward at low $d$ where its
excess approaches the
Monte Carlo floor; under a noise-floor-aware threshold all five rows
fall in band. Blind EM matches $[0.75, 1.25]$ on four of five rows
(the partial-merge boundary lands at $1.36$ with bootstrap CI
$[1.18, 1.67]$ overlapping the band). The neural blind residual
$\epsilon_{\mathrm{MLP}}^B$ at $N = 2000$ is below $1.8\%$ of
within$(t)$ on every row, including the partial-merge boundary.
We report blind MLP under a floor-magnitude criterion rather than a
rate-band criterion because its excess combines a constant
function-approximation floor, a small-$N$ approximation transient, and
a finite-$N$ parameter decay; the OLS log-log slope on this composite
signal is $N$-range dependent and is not a clean single rate.

\begin{table}[H]
  \centering
  \small
  \setlength{\tabcolsep}{6pt}
  \caption{\textbf{Blind-attainability panel.} Per-row
    $\alpha_{\mathrm{pooled}}$ with $95\%$ bootstrap CI for
    $\mathrm{oracle\_ols}$ and closed-form blind EM
    (low-$t$ window $t \in [0.10, 0.28]$, five
    controlled bridge cells, $20$ seeds), and $\epsilon_{\mathrm{MLP}}^B$ as a fraction
    of within$(t)$ at $N = 2000$ for the neural blind arm.
    PASS bands: $[0.85, 1.15]$ for $\mathrm{oracle\_ols}$, $[0.75, 1.25]$ for blind EM,
    $\epsilon_{\mathrm{MLP}}^B < 5\%$ for blind MLP.}
  \label{tab:blind-attain}
  \begin{tabular*}{0.9\linewidth}{@{\extracolsep{\fill}}lccc@{}}
    \toprule
    row & $\mathrm{oracle\_ols}$ $\alpha$ [CI] & EM $\alpha$ [CI] & MLP $\epsilon_{\mathrm{MLP}}^B$ \\
    \midrule
    $d{=}2$ sep      & $1.24$ [$0.97$, $2.03$] & $1.23$ [$0.99$, $1.55$] & $1.56\%$ \\
    $d{=}2$ p.m.     & $1.24$ [$0.97$, $2.03$] & $1.36$ [$1.18$, $1.67$] & $0.62\%$ \\
    $d{=}4$ simplex  & $1.17$ [$0.91$, $1.71$] & $1.22$ [$1.01$, $1.58$] & $1.54\%$ \\
    $d{=}8$ simplex  & $0.99$ [$0.86$, $1.26$] & $1.02$ [$0.83$, $1.50$] & $1.52\%$ \\
    $d{=}16$ simplex & $0.93$ [$0.82$, $1.08$] & $0.97$ [$0.87$, $1.13$] & $1.76\%$ \\
    \bottomrule
  \end{tabular*}
\end{table}

\subsection{Empirical Sweep of the Two-Stage Estimator}
\label{app:two-stage-sweep}

We instantiate the heuristic two-stage construction of
Remark~\ref{rem:oracle} (Stage~1: cluster recovery via the EM algorithm;
Stage~2: residual OLS within each estimated cell) on a controlled GMM
with $d{=}2$, $K{=}4$, $\sigma{=}0.5$, and four cluster centers on a
circle of radius equal to the swept spread, so the minimum pairwise
separation $\Delta_{\min}$ scales linearly with spread. We sweep
$\Delta_{\min}/\sigma \in \{1.41, 3.00, 4.24, 5.66, 8.00, 11.31\}$ to
span merge through separated regimes. The $N$ ladder is
$\{50, 100, 200, 500, 1000, 2000\}$ with $20$ training seeds and an
eval set of $10000$ paired test samples per training seed. The setup
is closed form (no neural network) so the comparison isolates Stage~1
misclassification cost.

Stage~1 is \texttt{sklearn.mixture.GaussianMixture} with $K{=}4$
spherical components, trained on the training actions $a_i$; hard
labels at training and eval are obtained as the argmax of the
time-$t$ posterior, computed under the analytic likelihood
$x_t \mid k \sim \mathcal{N}((1-t)\mu_k, \sigma_t^2 I)$ using the
EM-fit means with the analytic $\sigma_t^2$. Stage~2 fits the residual
intercept per estimated cluster as
$\hat b_{\hat k}(t) = \mathrm{mean}_{i: \hat k_i = \hat k}
(u_{t,i} - \beta_1(t) x_{t,i})$. We compare against the label-oracle
residual estimator, the label-and-action oracle (matching the
$\mathrm{oracle\_ols}$ arm of Section~\ref{sec:proxy-bridge}), and a
soft-EM marginal-Bayes mixture (matching $\mathrm{blind\_em\_ols}$).
The rate readout is
$\alpha_{\mathrm{pooled}} = -\partial \log \overline{R} / \partial \log N$
on the low-$t$ window $t \in [0.10, 0.28]$ with the same arm-aware
floor convention used in Appendix~\ref{app:blind-attain-panel}; $95\%$
confidence intervals come from $500$-resample paired-seed bootstrap.

Table~\ref{tab:two-stage-sweep} reports the per-spread outcome. At
$\Delta_{\min}/\sigma \geq 8$ the two-stage estimator attains the
label-oracle residual rate within band $[0.75, 1.25]$ ($\alpha = 0.82$ at
$\Delta_{\min}/\sigma{=}8$; $\alpha = 0.98$ at
$\Delta_{\min}/\sigma{=}11.31$, matching the label-oracle residual
estimator to two decimal places). Below $\Delta_{\min}/\sigma \leq 6$ the
rate collapses to $\alpha \in [0.17, 0.34]$ as Stage~1 misclassification
introduces an $N$-independent residual that asymptotes to $4$ to $13$
percent of within$(t)$ at $N{=}2000$. The misclassification floor
closely tracks the analytic between/within ratio across the sweep
(within $0.5$ percentage points), identifying the
$O(K \exp(-c''\Delta_{\mathrm{eff}}^2)\Delta_{\max}^2/\sigma_t^2)$
term in Remark~\ref{rem:oracle} as the dominant non-oracle
attainability cost. The empirical separated-to-merge transition is at
$\Delta_{\min}/\sigma \approx 6$ to $8$ in this setup; we do not
extrapolate this threshold beyond the swept range.

\begin{table}[H]
  \centering
  \small
  \setlength{\tabcolsep}{4pt}
  \caption{\textbf{Two-stage attainability sweep.} Per-spread
    $\alpha_{\mathrm{pooled}}$ with $95\%$ bootstrap CI for
    label-oracle residual estimator, two-stage EM (estimated label via
    Remark~\ref{rem:oracle}), and soft-EM marginal Bayes; the
    two-stage residual at $N{=}2000$ as percent of within$(t)$
    is reported alongside the analytic between/within ratio for
    reference. Window $t \in [0.10, 0.28]$, $20$ seeds,
    $M_{\mathrm{eval}}{=}10000$.}
  \label{tab:two-stage-sweep}
  \scriptsize
  \setlength{\tabcolsep}{2pt}
  \begin{tabular*}{\linewidth}{@{\extracolsep{\fill}}cccccc@{}}
    \toprule
    $\Delta_{\min}/\sigma$ & label-resid. $\alpha$ [CI] & two-stage $\alpha$ [CI] & soft-EM $\alpha$ [CI] & two-stage $R$ (\%w) & between/within (\%) \\
    \midrule
    $1.41$  & $1.05$ [$0.88$, $1.23$] & $0.21$ [$0.20$, $0.23$] & $0.59$ [$0.44$, $0.86$] & $7.13$  & $12.1$  \\
    $3.00$  & $1.13$ [$0.92$, $1.33$] & $0.17$ [$0.15$, $0.18$] & $1.30$ [$0.72$, $1.87$] & $13.43$ & $13.5$  \\
    $4.24$  & $1.08$ [$0.92$, $1.31$] & $0.20$ [$0.18$, $0.21$] & $1.39$ [$0.88$, $2.18$] & $9.60$  & $9.9$   \\
    $5.66$  & $1.06$ [$0.92$, $1.19$] & $0.34$ [$0.31$, $0.40$] & $1.40$ [$0.72$, $2.52$] & $3.96$  & $4.0$   \\
    $8.00$  & $1.09$ [$0.93$, $1.18$] & $0.82$ [$0.75$, $0.92$] & $1.02$ [$0.67$, $1.36$] & $0.45$  & $0.5$   \\
    $11.31$ & $0.98$ [$0.85$, $1.13$] & $0.98$ [$0.85$, $1.13$] & $0.94$ [$0.53$, $1.33$] & $0.23$  & $< 0.1$ \\
    \bottomrule
  \end{tabular*}
\end{table}

\section{BEHAVIOR-1K Materials}\label{app:b1k-tasks}

This appendix records BEHAVIOR-1K application-context diagnostics from earlier
diagnostic work: a leakage-free $\pi_{0.5}$ fine-tuning profile panel on
fifteen tasks over multiple demonstration seeds (Section~\ref{app:b1k-diag-panel}),
the diagnostic implementation and BC hyperparameters, and short descriptions
of five representative simulation tasks, collected together in
Appendix~\ref{app:b1k-task-descriptions}. These materials are not used as
primary evidence for the main controlled claims. For every task we summarize
the required behavior, the principal coordination challenge, the randomization
protocol that controls each rollout, and the maximum inference horizon.

\subsection{BEHAVIOR-1K diagnostic implementation}
\label{app:impl-b1k}

We record the BEHAVIOR-1K evaluation implementation for auditability. We
fine-tune and evaluate $\pi_{0.5}$ checkpoints through the
existing BEHAVIOR-1K/OpenPI pipeline. Offline evaluation loads parameter-only
checkpoints on a single GPU, uses local BEHAVIOR-1K data, and evaluates
time-resolved velocity MSE without environment rollout. The BC hyperparameters
used for these simulation fine-tunes are listed in
Table~\ref{tab:b1k-bc-hyperparams}.

\begin{figure}[H]
  \centering
  \safeincludegraphics[width=\linewidth]{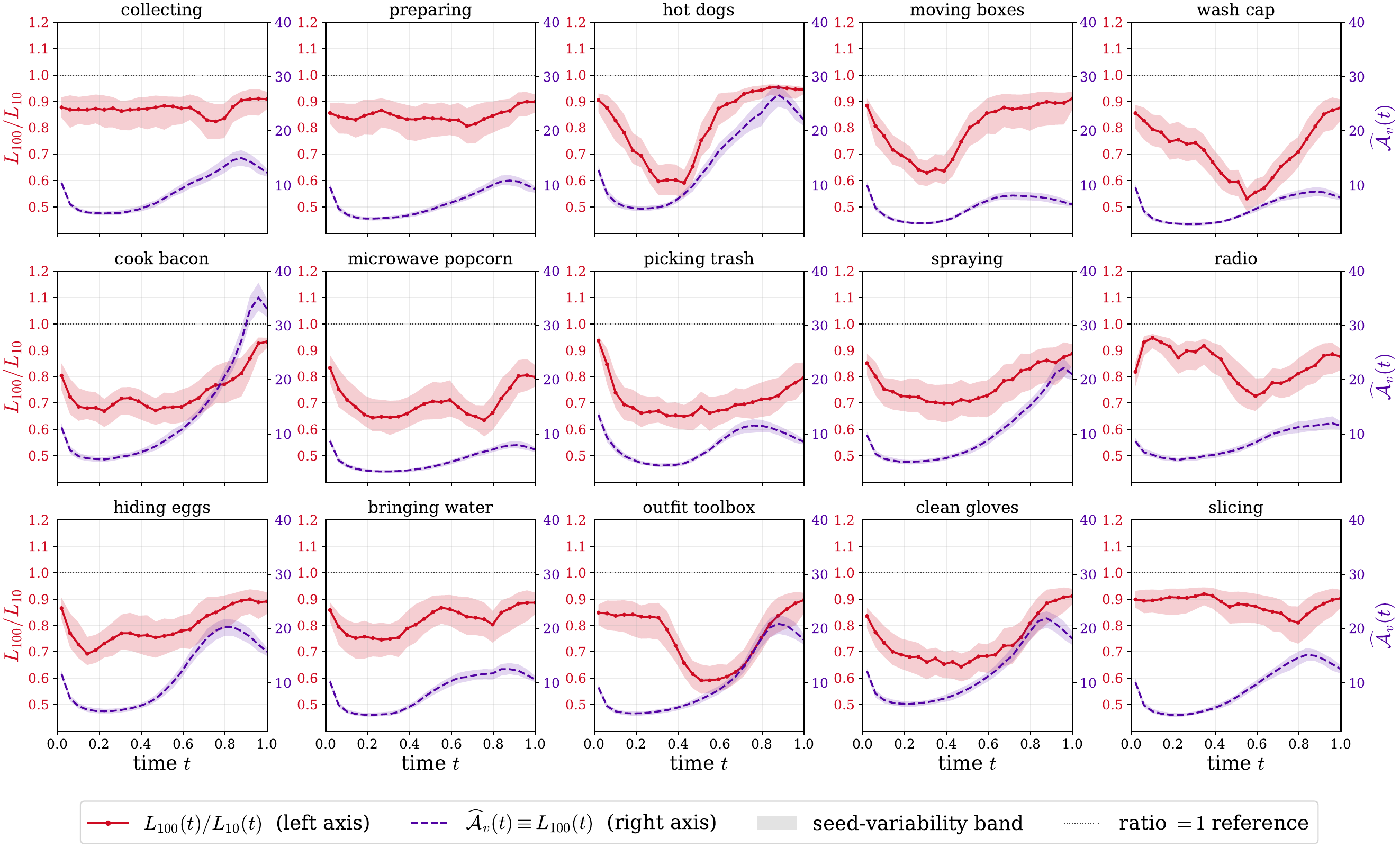}
  \caption{\textbf{Per-task ratio profiles and large-budget reference profile on BEHAVIOR-1K.}
    Each panel shows one leakage-free trio from the fifteen-task
    BEHAVIOR-1K diagnostic panel over multiple demonstration seeds. The left axis plots the per-bin ratio
    $L_{100}(t)/L_{10}(t)$ with the $\text{ratio}=1$ reference line; the right
    axis plots the large-budget reference profile $L_{100}(t)$, which is an
    upper-biased proxy for $\AV(t)$ rather than an absolute Bayes-floor
    estimate. All fifteen trios improve in mean loss from $N=10$ to $N=100$, but
    the temporal location of improvement is task-dependent.}
  \label{fig:b1k-per-task-grid}
\end{figure}

Each task in the leakage-free panel uses $N\in\{10,25,100\}$ demonstrations
over multiple demonstration seeds. Evaluation uses the
\texttt{random\_seed42} validation split, explicitly forbidding the
\texttt{N=10}, \texttt{N=25}, and \texttt{N=100} training pools for the
corresponding demonstration seed and task. The profile settings are
$25$ time bins, $20$ noise samples per bin, $50$ validation batches, batch size $32$,
and evaluation seed $42$. The output JSONs store the validation-split hash,
forbidden-pool disclosure, loss profile, and time grid used to construct
Table~\ref{tab:b1k-clean-panel} and Figure~\ref{fig:b1k-per-task-grid}.

\begin{table}[H]
  \centering
  \small
  \setlength{\tabcolsep}{18pt}
  \caption{\textbf{Hyperparameters of BC on simulation (based on $\pi_{0.5}$).}}
  \label{tab:b1k-bc-hyperparams}
  \begin{tabular}{cc}
    \toprule
    \textbf{Hyperparameter} & \textbf{Value} \\
    \midrule
    Action Expert Width & $1024$ \\
    Action Expert Depth & $18$ \\
    Action Expert MLP Dim & $4096$ \\
    Batch Size & $128$ \\
    Chunk Size & $30$ \\
    Learning Rate & $10^{-5}$ \\
    LR Scheduler & Cosine \\
    Optimizer & AdamW \\
    AdamW Betas & $[0.9, 0.95]$ \\
    AdamW Epsilon & $10^{-8}$ \\
    Weight Decay & $10^{-10}$ \\
    Max Training Steps & $30{,}000$ \\
    Fine-tune Method & Full parameter \\
    \bottomrule
  \end{tabular}
\end{table}

\subsection{Leakage-free diagnostic panel}
\label{app:b1k-diag-panel}

\paragraph{Protocol.}
We fine-tune $\pi_{0.5}$~\citep{pi05_2025} on BEHAVIOR-1K
tasks~\citep{li2024behavior}. Each task uses the Beta baseline time-sampling
protocol and $N\in\{10,25,100\}$ demonstrations over multiple demonstration seeds,
and the same evaluation settings: $25$ time bins, $20$ noise samples per bin,
$50$ validation batches, and evaluation seed~$42$. Main-panel evaluations use
the \texttt{random\_seed42} validation split with an explicit forbidden training
pool containing the \texttt{N=10}, \texttt{N=25}, and \texttt{N=100} entries
for the corresponding demonstration seed. All fifteen tasks in Table~\ref{tab:b1k-clean-panel}
use \texttt{forbidden\_train\_sets\_source=cli\_override}, with one
leakage-free validation hash per task. This audit is important:
the BEHAVIOR-1K results are interpreted as a leakage-free diagnostic panel,
not as a benchmark-wide law.

\paragraph{Profile readout.}
All fifteen tasks pass the endpoint gate $\bar L_{100}<\bar L_{10}$, but the
temporal location of the improvement varies substantially. We split the
$25$-bin grid into low/mid/high regions ($8/9/8$ bins:
$t<0.32$, $0.32\le t\le0.68$, and $t>0.68$) and report regional ratios
$\langle L_{100}\rangle/\langle L_{10}\rangle$. Three descriptive profile
families appear: near-uniform improvement, high-time reversal, and smooth
no-reversal bottlenecks. These are observed profile shapes in a clean
multi-demonstration-seed panel, not claimed as benchmark-wide task-family labels.

\begin{table}[H]
  \centering
  \scriptsize
  \setlength{\tabcolsep}{4pt}
  \caption{\textbf{Leakage-free BEHAVIOR-1K diagnostic panel.}
    All fifteen tasks use the \texttt{random\_seed42} protocol with the
    explicit $N=10/25/100$ forbidden training pool, reported over
    multiple demonstration seeds. Region ratios are low/mid/high
    $L_{100}/L_{10}$; lower is better. Reg.~CV is the coefficient of
    variation across the three region ratios. Peak~$t$ is the time bin
    maximizing the improvement gap $L_{10}(t)-L_{100}(t)$.}
  \label{tab:b1k-clean-panel}
  \begin{tabular*}{\linewidth}{@{\extracolsep{\fill}}lcccccc@{}}
    \toprule
    Task & $\bar L_{10}/\bar L_{25}/\bar L_{100}$ &
    $L_{100}/L_{10}$ & $L_{25}/L_{10}$ &
    Region ratios & Reg. CV & Peak $t$ \\
    \midrule
    \texttt{collecting} & $10.148$ / $10.435$ / $8.942$ & $0.881$ & $1.028$ &
    $0.876$ / $0.878$ / $0.885$ & $0.0045$ & $0.755$ \\
    \texttt{preparing} & $8.003$ / $7.978$ / $6.821$ & $0.852$ & $0.997$ &
    $0.850$ / $0.830$ / $0.869$ & $0.0189$ & $0.714$ \\
    \texttt{cook\_hot\_dogs} & $16.177$ / $16.737$ / $14.293$ & $0.883$ & $1.035$ &
    $0.767$ / $0.769$ / $1.019$ & $0.1391$ & $0.428$ \\
    \texttt{moving\_boxes} & $7.121$ / $6.825$ / $5.847$ & $0.821$ & $0.958$ &
    $0.756$ / $0.789$ / $0.898$ & $0.0743$ & $0.388$ \\
    \texttt{wash\_cap} & $7.856$ / $7.231$ / $5.606$ & $0.713$ & $0.920$ &
    $0.797$ / $0.594$ / $0.768$ & $0.1249$ & $0.633$ \\
    \texttt{cook\_bacon} & $17.284$ / $16.040$ / $13.577$ & $0.786$ & $0.928$ &
    $0.715$ / $0.696$ / $0.850$ & $0.0913$ & $0.878$ \\
    \texttt{make\_microwave\_popcorn} & $7.323$ / $6.132$ / $5.202$ & $0.710$ & $0.837$ &
    $0.716$ / $0.688$ / $0.719$ & $0.0255$ & $0.020$ \\
    \texttt{picking\_up\_trash} & $10.998$ / $10.238$ / $7.937$ & $0.722$ & $0.931$ &
    $0.772$ / $0.665$ / $0.727$ & $0.0578$ & $0.020$ \\
    \texttt{spraying\_fruit\_trees} & $12.744$ / $11.413$ / $10.191$ & $0.800$ & $0.896$ &
    $0.762$ / $0.717$ / $0.848$ & $0.0684$ & $1.000$ \\
    \texttt{turning\_on\_radio} & $9.862$ / $8.590$ / $8.285$ & $0.840$ & $0.871$ &
    $0.919$ / $0.794$ / $0.838$ & $0.0608$ & $0.633$ \\
    \texttt{hiding\_Easter\_eggs} & $13.711$ / $13.080$ / $11.338$ & $0.827$ & $0.954$ &
    $0.764$ / $0.778$ / $0.882$ & $0.0651$ & $0.633$ \\
    \texttt{bringing\_water} & $10.045$ / $9.785$ / $8.365$ & $0.833$ & $0.974$ &
    $0.783$ / $0.834$ / $0.855$ & $0.0367$ & $0.796$ \\
    \texttt{outfit\_a\_basic\_toolbox} & $13.493$ / $12.013$ / $10.159$ & $0.753$ & $0.890$ &
    $0.841$ / $0.628$ / $0.802$ & $0.1221$ & $0.714$ \\
    \texttt{clean\_boxing\_gloves} & $15.464$ / $12.714$ / $11.879$ & $0.768$ & $0.822$ &
    $0.728$ / $0.678$ / $0.849$ & $0.0950$ & $0.714$ \\
    \texttt{slicing\_vegetables} & $10.196$ / $9.602$ / $8.951$ & $0.878$ & $0.942$ &
    $0.917$ / $0.883$ / $0.861$ & $0.0257$ & $0.796$ \\
    \bottomrule
  \end{tabular*}
\end{table}

\FloatBarrier

\subsection{Representative task descriptions}
\label{app:b1k-task-descriptions}

\phantomsection\label{app:b1k-radio}
\paragraph{Turning on a radio.}
The robot first uses its onboard cameras to identify the radio on the
tabletop and approach it. The right gripper is then used to take hold of
the device, after which the left gripper depresses the power button; once
the radio has been switched on, it is returned to the table surface. What
makes this scenario non-trivial is the joint requirement of locomotion
plus dual-arm operation: the right hand has to keep a stable grip so that
the reaction force produced by the button press does not dislodge the
radio. At the start of every rollout the radio is placed at a different
pose on the table, although it is always within the robot's field of
view and reach. Inference is allowed at most $6{,}000$ steps.

\begin{figure}[htbp]
  \centering
  \safeincludegraphics[width=1\linewidth]{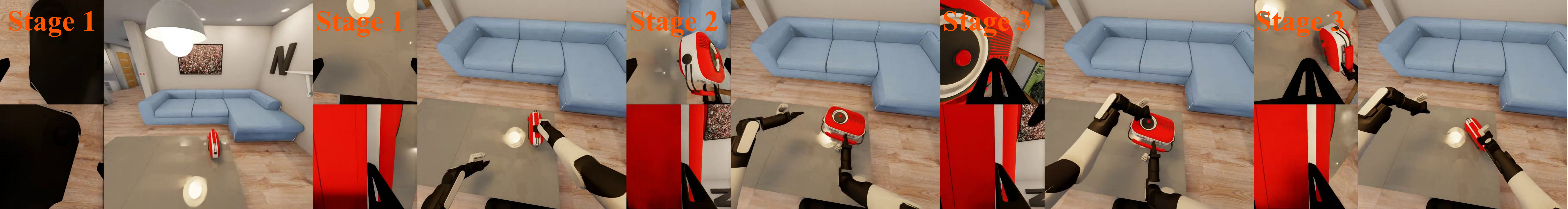}
  \caption{\textbf{Turning on a radio.} BEHAVIOR-1K task visualization
    used for application-context reporting.}
  \label{fig:b1k-radio}
\end{figure}

\phantomsection\label{app:b1k-trash}
\paragraph{Picking up trash.}
In this scenario the right gripper holds a refuse container throughout
the entire rollout while the robot navigates in turn to three soda cans
distributed across the scene. At each can, the left gripper is used to
lift the can and drop it into the container kept by the right hand;
after all three cans have been collected, the filled container is set
back onto the floor. The difficulty arises from a long-horizon mobile
manipulation pattern that imposes asymmetric bimanual roles: the right
arm acts as a passive but stable carrier, while the left arm is in
charge of repeated grasp-and-release motions. Per rollout, both the
container and the cans are spawned at randomized positions so that
nontrivial navigation is required, and the robot's initial base pose is
also drawn from a bounded distribution to probe generalization. The
inference budget for each rollout is $13{,}000$ steps.

\begin{figure}[htbp]
  \centering
  \safeincludegraphics[width=1\linewidth]{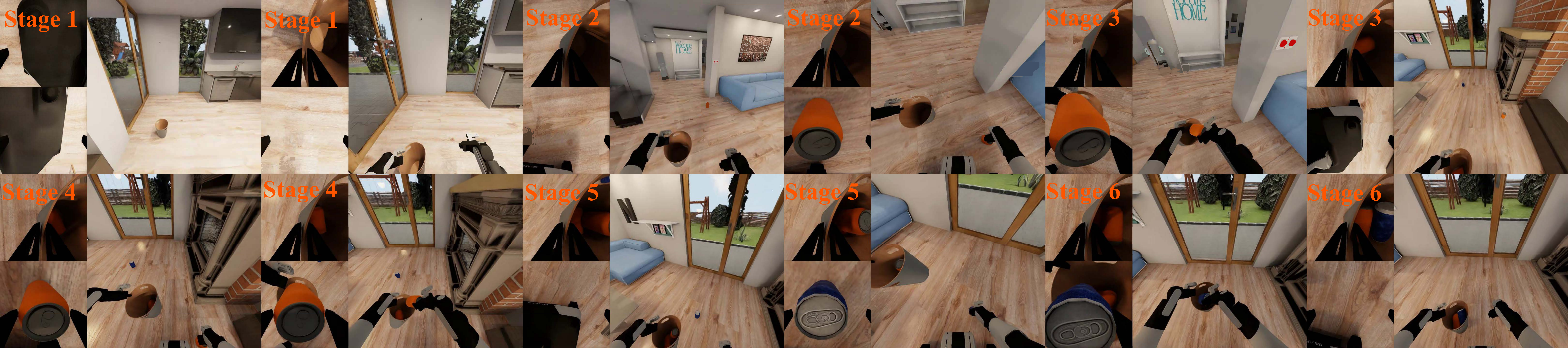}
  \caption{\textbf{Picking up trash.} BEHAVIOR-1K task visualization
    used for application-context reporting.}
  \label{fig:b1k-trash}
\end{figure}

\phantomsection\label{app:b1k-spray}
\paragraph{Spraying fruit trees.}
Here the robot uses vision to find a sprayer, picks it up with the left
gripper, and then visits two fruit trees one after the other. Upon
arrival at each tree, the right gripper toggles the sprayer's switch to
the on position, performs the watering action, and toggles the switch
back off before the platform moves on to the next tree. The chief
difficulty is bimanual tool use combined with mobile manipulation: the
left arm must hold the fluid container steady so that the spray remains
aimed at the target, while the right arm produces the contact force on
the switch without disturbing this alignment. The sprayer and both trees
are spawned at random workspace positions, forcing the robot to actively
search the scene and replan its path; the starting base pose of the
robot is also varied across rollouts. The inference horizon is capped at
$10{,}000$ steps.

\begin{figure}[htbp]
  \centering
  \safeincludegraphics[width=1\linewidth]{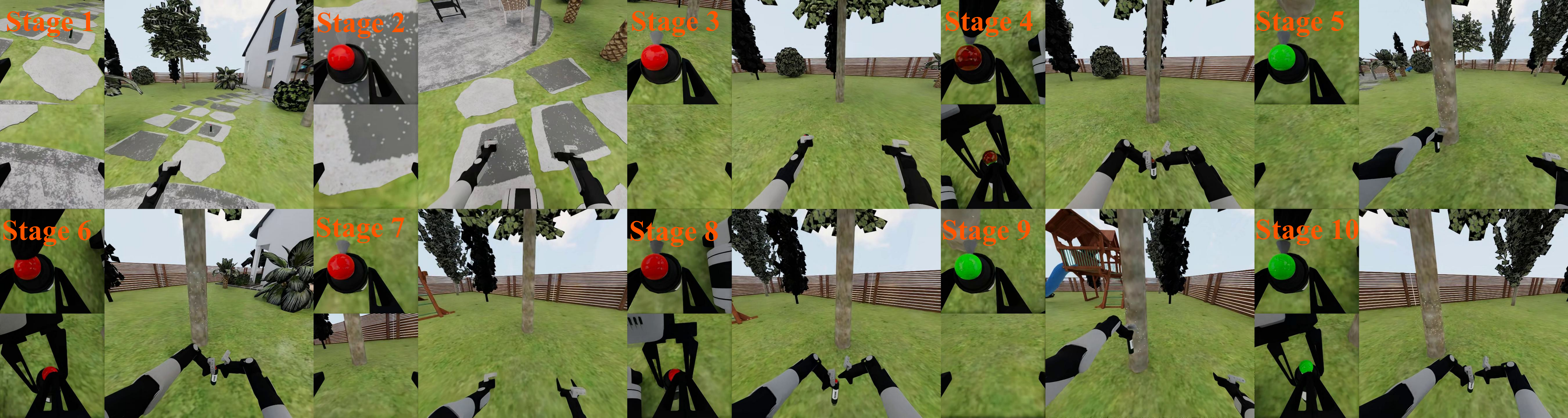}
  \caption{\textbf{Spraying fruit trees.} BEHAVIOR-1K task visualization
    used for application-context reporting.}
  \label{fig:b1k-spray}
\end{figure}

\phantomsection\label{app:b1k-hotdogs}
\paragraph{Cooking hot dogs.}
This task strings together articulated-object handling with intermediate
buffering of cargo. The robot first opens the refrigerator, takes out
two hot dogs, and shuts the door; one of the items is set aside on the
tabletop while the other is loaded into the microwave; the buffered hot
dog is then transferred from the table into the microwave, the
microwave door is closed, and the start switch is engaged. The
fundamental challenge is long-horizon planning across multiple
workstations (the storage compartment, the staging area, and the
cooking appliance), with the kinematics of the appliance doors having
to be coordinated with pick-and-place operations. Initial object poses
and the robot's starting base pose are varied between rollouts to
evaluate generalization, and the inference horizon is set to $15{,}000$
steps.

\begin{figure}[htbp]
  \centering
  \safeincludegraphics[width=1\linewidth]{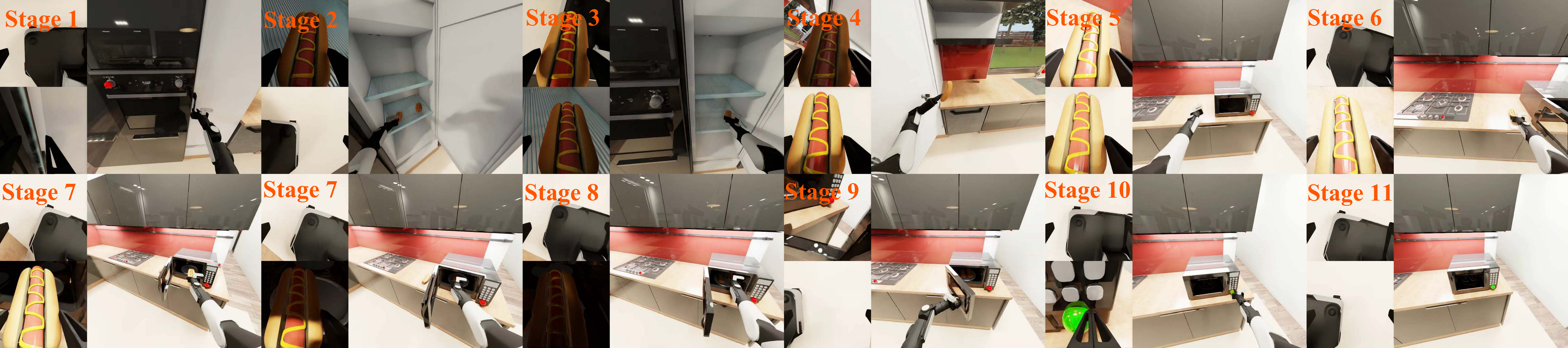}
  \caption{\textbf{Cooking hot dogs.} BEHAVIOR-1K task visualization
    used for application-context reporting.}
  \label{fig:b1k-hotdogs}
\end{figure}

\phantomsection\label{app:b1k-cap}
\paragraph{Washing a baseball cap.}
The robot moves to a washing machine, opens its door, and finds the two
baseball caps that need to be cleaned. With one cap held in each
gripper, it walks back to the washer and inserts both caps into the
narrow drum cavity one after the other; the door is then shut and the
power switch is pressed. The hard part of this task is a long-horizon
mobile manipulation pipeline that combines articulated-object operation
with carrying objects in both hands at the same time: the robot has to
navigate while its grippers are occupied, and then perform
collision-free placements inside a tightly bounded workspace. Both the
poses of the two caps and the robot's initial base pose are randomized
across trials, and inference is limited to $10{,}000$ steps.

\begin{figure}[htbp]
  \centering
  \safeincludegraphics[width=1\linewidth]{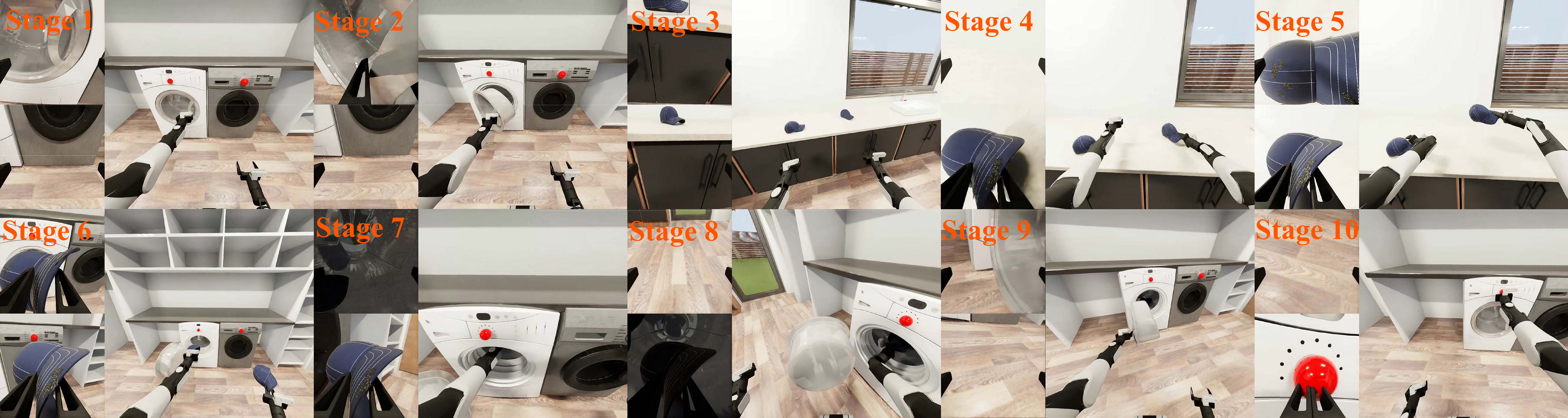}
  \caption{\textbf{Washing a baseball cap.} BEHAVIOR-1K task visualization
    used for application-context reporting.}
  \label{fig:b1k-cap}
\end{figure}

\end{document}